**Identifying acute illness phenotypes via deep temporal interpolation and clustering network on physiologic signatures**

Yuanfang Ren, PhD[1,2$], Yanjun Li[3,4$], Tyler J. Loftus, MD[1,5$], Jeremy Balch, MD[5], Kenneth L. Abbott, MD[5], Shounak Datta, PhD[1,2], Matthew M. Ruppert, BS[1,2], Ziyuan Guan, MS[1,2], Benjamin Shickel, PhD[1,2], Parisa Rashidi, PhD[1,6], Tezcan Ozrazgat-Baslanti, PhD[1,2*], Azra Bihorac, MD, MS[1,2*]

[1]Intelligent Critical Care Center, University of Florida, Gainesville, FL

[2]Department of Medicine, Division of Nephrology, Hypertension, and Renal Transplantation, University of Florida, Gainesville, FL

[3]Department of Medicinal Chemistry, College of Pharmacy, University of Florida, Gainesville, FL

[4]Center for Natural Products, Drug Discovery and Development, University of Florida, Gainesville, FL

[5]Department of Surgery, University of Florida, Gainesville, FL

[6]J. Crayton Pruitt Family Department of Biomedical Engineering, University of Florida, Gainesville, FL

[$]These authors contributed equally.

*These authors contributed equally as senior authors.

**Correspondence to:**

Azra Bihorac MD, MS, Department of Medicine, Division of Nephrology, Hypertension, and Renal Transplantation, PO Box 100224, Gainesville, FL 32610-0254. Phone: (352) 273-9009; Fax: (352) 392-5465; Email: abihorac@ufl.edu

**Abstract**

**Background:** While the initial few hours of a hospital admission can significantly impact a patient's clinical trajectory, early clinical decisions often suffer due to data paucity. By using clustering analysis for patient vital signs that were recorded in the first six hours after hospital admission, unique patient phenotypes with distinct pathophysiological signatures and clinical outcomes may be revealed and support early clinical decision-making. Historically, phenotyping based on these early vital signs has proven challenging, as vital signs are typically sampled sporadically.

**Methods:** We created a single-center, longitudinal dataset of electronic health record data for 75,762 adult patients admitted to a tertiary care center for at least six hours. We proposed a novel, deep temporal interpolation and clustering network to simultaneously extract latent representations from sparse and irregularly sampled vital sign data and derived distinct patient phenotypes within a training cohort (n=41,502). Model and hyper-parameters were selected based on a validation cohort (n=17,415). A test cohort (n=16,845) was used to analyze reproducibility and correlation with clinical biomarkers.

**Results:** The three cohorts—training, validation, and testing—had comparable distributions of age (54-55 years), sex (55% female), race, comorbidities, and illness severity. Four distinct clusters were identified. Phenotype A (18%) had the greatest prevalence of comorbid disease with increased prevalence of prolonged respiratory insufficiency, acute kidney injury, sepsis, and long-term (three-year) mortality. Phenotypes B (33%) and C (31%) had a diffuse pattern of mild organ dysfunction. Phenotype B's favorable short-term clinical outcomes were tempered by the second highest rate of long-term mortality. Phenotype C had favorable clinical outcomes. Phenotype D (17%) exhibited early and persistent hypotension, high incidence of early surgery, and substantial biomarker incidence of inflammation. Despite early and severe illness, phenotype D had the second lowest long-term mortality. After comparing the various phenotypes' sequential organ failure assessment scores, the results of the clustering did not simply provide a recapitulation of previous acuity assessments.

**Conclusions:** Within a heterogeneous cohort of patients in hospitals, four phenotypes with distinct categories of disease and clinical outcomes were identified by using a deep temporal interpolation and clustering network. This tool may impact triage decisions and have important implications for clinical decision-support under time constraints and uncertainty.

## Introduction

Every year in the United States, more than 36 million hospital admissions occur, with approximately seven hundred thousand in-hospital deaths, nearly one-fourth of which are potentially preventable.[1-3] A significant source of preventable harm during the early stages of hospital admission is the misdiagnosis and under-triage of high-risk patients to general hospital wards.[4,5] In this crucial period, clinicians are required to make a series of decisions involving monitoring, testing, and treatment that can significantly influence the patient's clinical trajectory.[1,6,7] This series of decisions entails analysis of a variety of data representing essential physiologic processes.[6-8] For example, values and trends in vital signs may indicate whether a patient requires intensive monitoring in an intensive care unit (ICU) or if they can be safely transferred to a hospital's general ward. The trajectories of early vital signs may be useful for identifying distinct physiological signatures that are linked to specific patient phenotypes and clinical outcomes.

Unsupervised clustering analyses of vital signs and other clinical variables have shown promise for helping clinicians identify novel clinical phenotypes for sepsis and acute respiratory distress syndrome.[8-11] However, these phenotypes have not been evaluated with large, heterogeneous cohorts that include all hospitalized patients. In addition, broader phenotyping based on vital signs has proven more challenging, in part because sampling occurs at irregular intervals, thus complicating the application of conventional time series analyses and machine learning clustering techniques. In the past decade, however, deep learning has garnered significant achievements in the healthcare domain to facilitate the clinical decision-making process with its superior capability to detect the intricate patterns inherent in raw clinical data and to approximate highly complex functions.[12-14] Although several advanced deep learning algorithms have been developed to manage the irregularly-sampled time series data,[15-17] there remains a dearth of work specifically focused on the clinical phenotype identification, particularly using the early stages vital sign data.

To fill this gap and address the patient stratification challenge, our study presents a novel deep temporal interpolation and clustering (dTIC) network. This innovative tool is designed to extract latent representations from sparse and irregularly sampled time series vital sign data, and concurrently stratify patients into distinct phenotypes. The dTIC network exhibits considerable potential to effectively facilitate clinical decision-making, offering a promising solution to existing limitations in patient phenotype identification during the critical initial hours of hospital admission.

## Methods

### Data Source and Participants

By using electronic health records (EHR) of 75,762 hospital admissions of 43,598 unique patients that represent adults (18 years or older) of all demographics, we created a longitudinal dataset of adult patients in the University of Florida Health's 1,000-bed academic hospital who remained admitted for six hours or longer (including emergency department admission) between June 1, 2014, and April 1, 2016. We excluded patients without sufficient vital sign measurements within six hours of hospital admission—that is, when two or more of the six vital sign measurements (systolic and diastolic blood pressure, heart rate, respiratory rate, temperature, and oxygen saturation) were completely missing (eFigure 1). This study was approved by the University of Florida institutional review board as exempt with waiver of informed consent (IRB201901123). All methods were performed in accordance with relevant guidelines and regulations.

### Study design

To mitigate any consequences of dataset drift because of adjustments in clinical practice or patient population, we adhered to the guidelines of the Type 2b analysis category[18] under the Transparent Reporting of a multivariate prediction model for Individual Prognosis or Diagnosis (TRIPOD) in order to split the dataset chronologically into three categories—training (patients admitted from June 1, 2014, to May 31, 2015, n=41,502), validation (patients admitted from June 1, 2015, to October 31, 2015, n=17,415), and testing (patients admitted from November 1, 2015, to April 1, 2016, n=16,845)—which followed a previous paper setting.[19] Using the training cohort, we identified acute illness phenotypes by applying unsupervised machine learning clustering to chronologically ordered measurements of patient vital signs from the first six hours of hospital admission. We utilized the validation cohort to select the hyperparameters of our dTIC model. Within the testing cohort, we assessed phenotype reproducibility by predicting phenotypes and analyzing phenotype frequency distributions and clinical outcomes.

**Identifying acute illness phenotypes via early physiologic signatures**

We removed outliers from raw time series vital signs and explored distributions, missingness, and correlation (eTable 1 and eFigure 2). In instances where a time-series variable was entirely absent from a patient's record, we imputed the mean value with a timestamp at hospital admission derived from the training cohort.

The dTIC network, designed to extract representations from sparse and irregularly sampled time-series data such as vital signs and subsequently derive acute illness phenotypes, incorporates four components: an interpolation model, a sequence transformation model, a re-interpolation model, and a clustering model (Figure 1 and eFigure 3). Initially, the interpolation model[20] converts the raw time-series data into a meta-representation of data sampled at pre-defined, regularly spaced reference time points. Following this, a sequence-to-sequence (seq2seq) model, equipped with the gated recurrent unit layers,[21] learns the features of the interpolated data and encodes them into low-dimensional vectors to form a unifying contextual representation. The decoder within the seq2seq model learns from this context vector and outputs regular time-series data of identical length as its input. Subsequently, a radial basis function network[22] re-interpolates the regular time series output back to raw irregular time points. This process generates estimates that can be compared to observations, providing a measure of network performance.

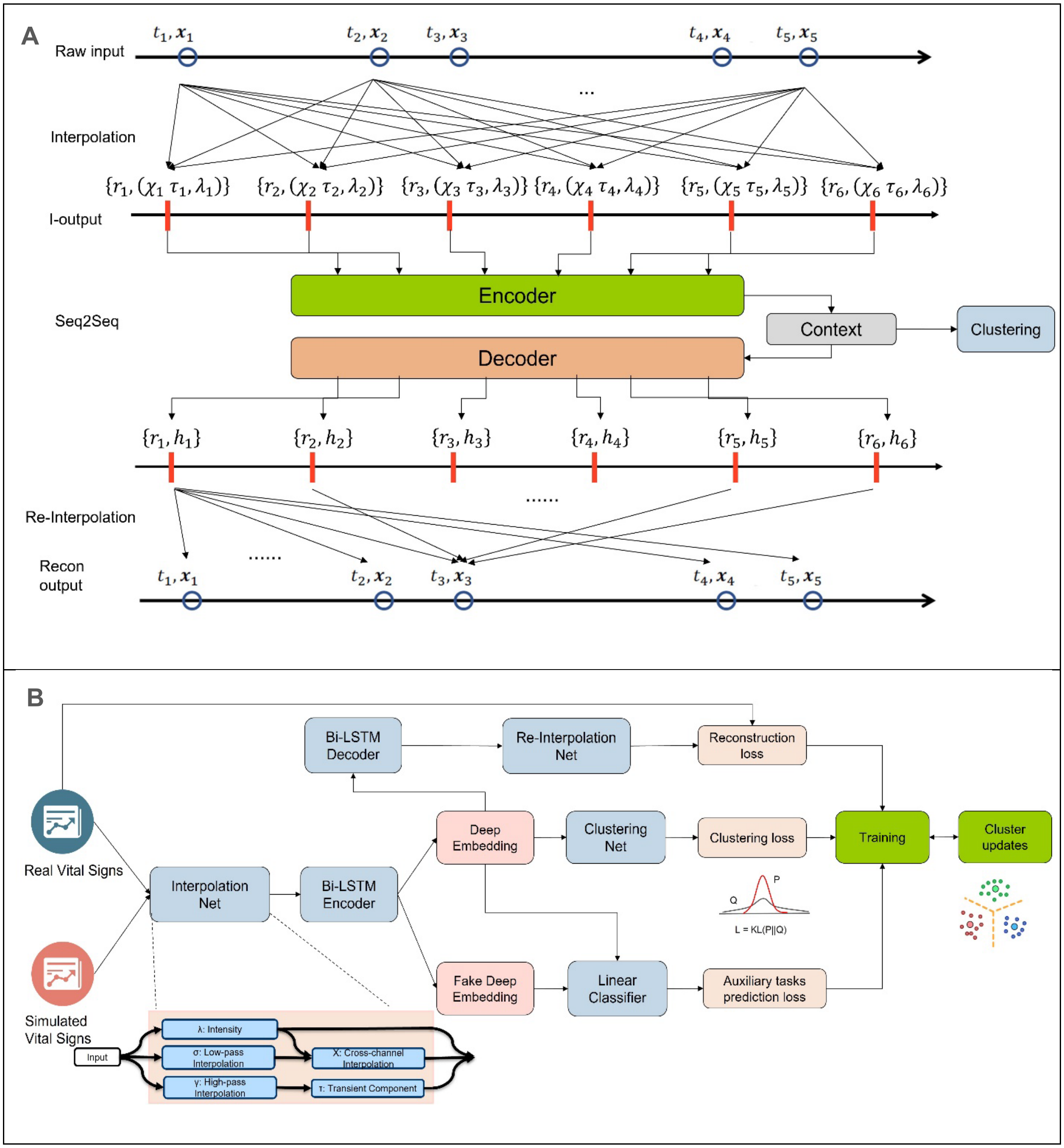

**Figure 1. Schematic representation of deep temporal interpolation and clustering network architecture.** (A) The detailed architecture of deep interpolation network, specifically tailored for handling sparse and irregularly sampled time series data. (B) The full architecture of deep

temporal interpolation and clustering network, designed to concurrently extract the feature representation and determine cluster assignments.

To foster a more comprehensive representation of time-series data, we employ two auxiliary prediction tasks: 1) predicting the minimum values of systolic and diastolic blood pressure as well as oxygen saturation, along with the maximum values of heart rate, respiratory rate, and temperature within the seventh hour; and 2) predicting whether the learned representation originated from actual time-series data, by feeding both real and synthetic time series data into the model (eMethods). Prior to clustering process, the dTIC network undergoes pre-training, which is achieved by minimizing both the mean square error of these reconstructed estimates and the prediction error of auxiliary tasks.

In addition, we stack a clustering network on top of the aforementioned feature extraction model to perform concurrent representation learning and clustering. The goal is to enhance the alignment of feature representations and cluster assignments.[23] This integrated approach demonstrates significant potential in learning clustering-friendly representations by which objects can be effectively grouped. The initial cluster assignments are obtained using the contextual representation derived from the pre-trained dTIC network and k-means clustering.[24] For a detailed description of our proposed network, readers are referred to eMethods.

**Clinical outcomes**

With every hospital admission, we extracted information for demographics, 19 clinical biomarkers that are assessed upon admission (eTable 2), acuity scores for both Sequential Organ Dysfunction Assessment (SOFA) and Modified Early Warning Score (MEWS), and patient outcomes.[25,26] Data processing details are explained in the eMethods section. The primary outcomes were 30-day mortality and 3-year mortality, and the median duration until follow-up was 4.3 years, according to calculations using the reverse Kaplan-Meier method. The secondary outcomes consisted of admission to an intensive care unit (ICU) or intermediate care unit (IMC), mechanical ventilation (MV), acute kidney injury (AKI), sepsis, venous thromboembolism, and renal replacement therapy (RRT).

**Statistical methods**

To ascertain the optimal number of phenotypes with the dTIC approach, we evaluated a combination of phenotype size, Davies-Bouldin index,[27] silhouette score,[28] elbow method,[29] and gap statistic method.[30] Once the optimal phenotype number was ascertained, patterns of vital signs were visualized by using t-distribution stochastic neighbor embedding (t-SNE) plots, ranked plots that show phenotype pairwise mean standardized differences, line plots with 95% confidence intervals, alluvial plots, and chord diagrams (see eMethods for a comprehensive description).

We assessed the reproducibility of derived phenotypes by assessing their frequency distributions and associated clinical characteristics in the testing cohort. The testing cohort's phenotype assignments were determined through the clinical characteristics of the specific cohort cluster. Predictions were based on the minimum Euclidean distance between individual patients to the phenotype's centroid (eMethods).

To compare phenotypes, we used the $\chi^2$ test for categorical variables and analysis of variance as well as the Kruskal-Wallis test for continuous variables. We used Kaplan–Meier curves to illustrate overall survival and the log-rank test to compare overall survival. Comparisons of adjusted hazard ratios (HR) were made for all phenotypes by using Cox proportional-hazards regression while controlling for demographics, comorbidities, and acuity score when admitted. Using the Bonferroni correction, all *p* values were adjusted for multiple comparisons. To ensure that the phenotypes did not simply recapitulate existing acuity scores, we used alluvial plots and chord diagrams to compare the phenotypes to patients' SOFA scores within 24 hours of hospital admission. Python version 3.7 and R version 3.5.1 were used to perform analyses.

## Results

### Patients

All three cohorts were comparable in clinical characteristics, biomarker distributions, and outcomes (eTables 3 and 4). [24] Across the cohorts, sex was equally distributed and the patients' average age was 54 years old. Nearly two-thirds of the admissions were urgent admission, 18% of the patients were transfers from other hospitals, 27% were admitted to the hospital's ICU or IMC, and 28% underwent surgery while admitted. Of the 27% of patients who were admitted to the ICU or IMC, 22%–27% had high SOFA (>6) or MEWS (>4) scores when admitted. Of the 73% of patients who were admitted to hospital wards, only 2%–3% had high acuity scores. For all cohorts, the 30-day mortality rate was 4% and the 3-year mortality rate was 19%.

### Derivation and characteristics of phenotypes

In the training cohort, the dTIC model determined an optimal fit with a four-class model, optimizing a combination of metrics including phenotype size, Davies-Bouldin index, silhouette score, and elbow and gap statistic methods (eFigure 4 and eTable 5). These four phenotypes were associated with distinct pathophysiological signatures and clinical outcomes (Tables 1 and 2, eTables 6 and 7, Figure 2). The phenotypes were categorized as phenotype A (18% of the cohort), B (33% of the cohort), C (31% of the cohort), and D (17% of the cohort) in relation to the descending systolic blood pressure value (Figure 2A).

**Phenotype A.** Phenotype A had the greatest burden of comorbid disease, such as hypertension (54%) and cardiovascular disease (32%), the highest proportion of African American race (27% vs. 17%–25% in other phenotypes) and emergent admissions (95%). Phenotype A had the highest rate of prolonged respiratory insufficiency (9% received MV, 58% of whom received ventilator support for more than two days), AKI (21%), sepsis (14%), and three-year mortality (25%). Patients in phenotype A had the second greatest incidence of admission to ICU/IMC (35%), hospital mortality (4%), and 30-day mortality (6%).

**Phenotype B.** Phenotype B exhibited physiological signatures similar to those of phenotype A, but displayed a diffuse pattern of mild organ dysfunction with persistent, uncorrected blood pressure abnormalities during the first six hours. Phenotype B had favorable short-term clinical outcomes, manifested by the second lowest rate of ICU/IMC admission (19%), AKI (17%), sepsis (8%), hospital mortality (2%), and 30-day mortality (3%). Phenotype B corresponded to the second highest rate of three-year mortality.

**Phenotype C.** Phenotype C exhibited low early physiological derangement with a diffuse pattern of mild organ dysfunction. Phenotype C exhibited favorable clinical outcomes, which manifested as the lowest rate of ICU/IMC admission (19%), AKI (13%), sepsis (6%), hospital mortality (1%), 30-day mortality (2%), and three-year mortality (15%). They had the second highest rate of surgery within 24 hours of admission (27%) but similar rates of admission to wards as phenotype B.

**Phenotype D.** Phenotype D was characterized by early and persistent hypotension, a high incidence of vasopressor support (53%), and the highest proportion requiring early surgery (57%). Phenotype D had significant biomarker incidence of inflammation, evidenced by the highest median white blood cell count ($11 \times 10^9$/L compared with $9 \times 10^9$/L in other phenotypes), premature neutrophils (15% vs. 7%–12% in other phenotypes), and C-reactive protein (39 mg/L vs. 15–20 mg/L in other phenotypes); and the lowest median lymphocytes (10%). Phenotype D had the highest rate of ICU/IMC admission (46%), MV (19%), hospital mortality (5%), and 30-day mortality (6%). They had the second highest incidence of AKI (19%) and sepsis (12%). Despite early and severe illness, phenotype D had favorable long-term outcomes with the second lowest three-year mortality (18%).

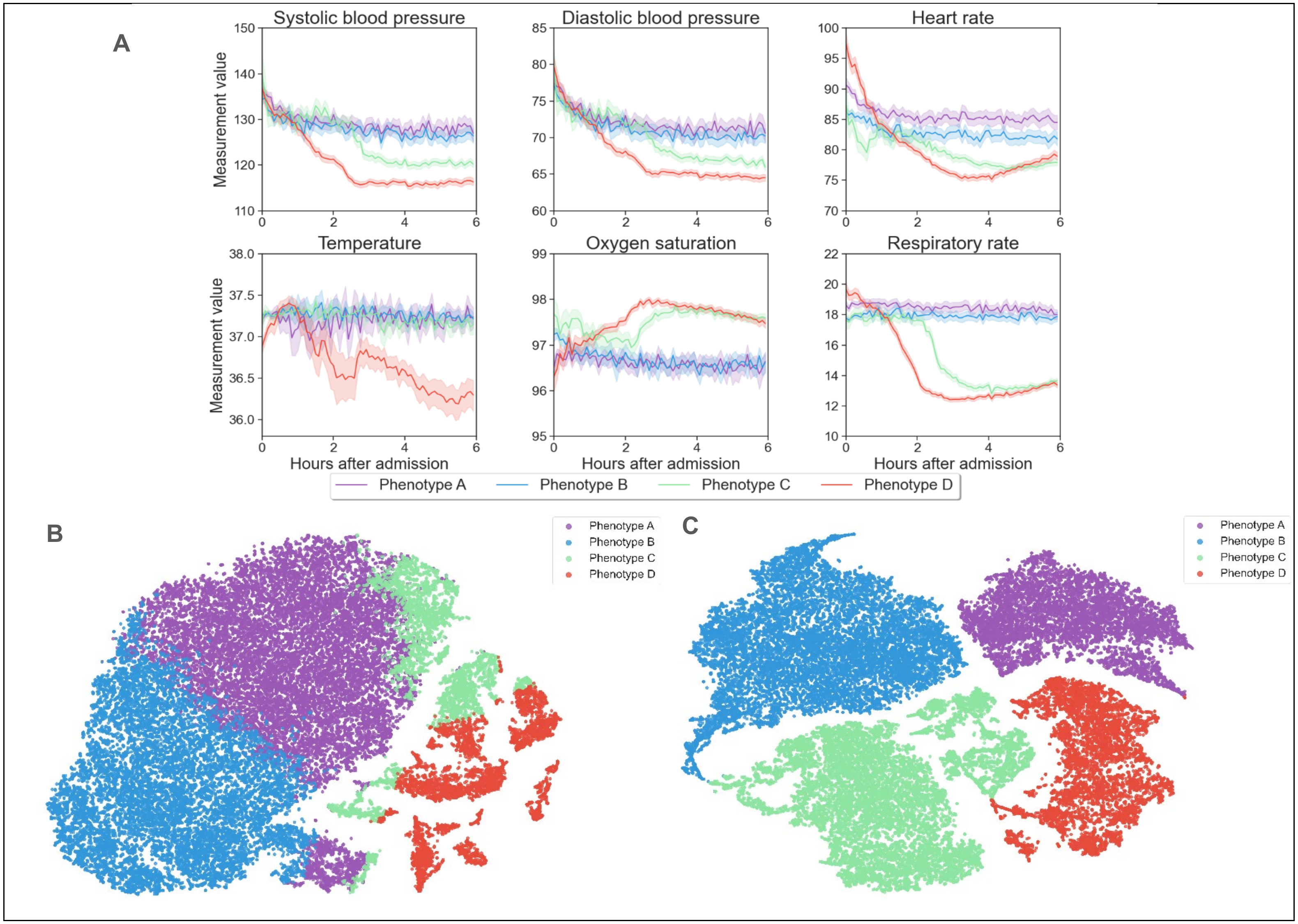

**Figure 2. Vital sign representations across identified phenotypes.** (A) Distribution of vital signs recorded within the initial six hours following hospital admission. (B) Visualization of initial phenotypes, as assigned by the pre-trained deep temporal interpolation network, without the integration of the clustering network. The t-distributed stochastic neighbor embedding (t-SNE)

technique was utilized to reduce the original 128-dimensional vital sign representations to two dimensions. Each dot signifies an individual patient, with separate colors indicating different phenotypes. (C) Visualization of final phenotypes, as assigned by the deep temporal interpolation and clustering network utilizing the t-SNE technique. The network simultaneously learns feature representation and cluster assignments, thus facilitating clustering-friendly representation learning.

**Table 1. Phenotype biomarkers and clinical characteristics.**

| **Variables** | **Total** | **Acute Illness Phenotypes** | | | |
|---|---|---|---|---|---|
| | | Phenotype A | Phenotype B | Phenotype C | Phenotype D |
| Number of encounters (%) | 41,502 | 7,647 (18) | 13,710 (33) | 12,901 (31) | 7,244 (17) |
| **Clinical characteristics—Preadmission** | | | | | |
| Age, mean (SD), years | 54 (19) | 57 (19)[a] | 53 (19)[a] | 51 (19) | 57 (17)[a] |
| Female sex, n (%) | 22,745 (55) | 3,963 (52)[a] | 7,595 (55)[a] | 7,391 (57) | 3,796 (52)[a] |
| Race, n (%) | | | | | |
| White | 29,076 (70) | 5,203 (68)[a] | 9,421 (69) | 9,021 (70) | 5,431 (75)[a] |
| African American | 9,634 (23) | 2,036 (27)[a] | 3,411 (25)[a] | 2,930 (23) | 1,257 (17)[a] |
| Primary insurance, n (%) | | | | | |
| Private | 9,591 (23) | 1,323 (17)[a] | 2,917 (21)[a] | 3,314 (26) | 2,037 (28)[a] |
| Medicare | 18,499 (45) | 3,839 (50)[a] | 6,120 (45)[a] | 5,158 (40) | 3,382 (47)[a] |
| Medicaid | 9,231 (22) | 1,641 (21)[a] | 3,213 (23) | 3,104 (24) | 1,273 (18)[a] |
| Uninsured | 4,181 (10) | 844 (11) | 1,460 (11) | 1,325 (10) | 552 (8)[a] |
| Residing neighborhood characteristics | | | | | |
| Proportion of African Americans (%), mean (SD) | 18.7 (17.5) | 19.6 (17.8)[a] | 19.3 (17.8)[a] | 18.6 (17.5) | 17.2 (16.1)[a] |
| Proportion below poverty (%), mean (SD) | 22.7 (10.1) | 23.8 (10.1)[a] | 23.1 (10.1)[a] | 22.5 (10.0) | 21.2 (9.8)[a] |
| Distance from hospital (mile), median (IQR) | 18 (3, 34) | 14 (3, 27)[a] | 14 (3, 32)[a] | 18 (3, 36) | 23 (9, 40)[a] |
| **Comorbidities** | | | | | |
| Hypertension, n (%) | 21,639 (52) | 4,129 (54)[a] | 7,205 (53) | 6,704 (52) | 3,601 (50)[a] |
| Cardiovascular disease, n (%)[b] | 12,058 (29) | 2,413 (32)[a] | 3,991 (29) | 3,682 (29) | 1,972 (27) |
| Diabetes mellitus, n (%) | 10,111 (24) | 1,972 (26)[a] | 3,370 (25) | 3,100 (24) | 1,669 (23) |
| Chronic kidney disease, n (%) | 6,518 (16) | 1,450 (19)[a] | 2,467 (18)[a] | 1,802 (14) | 799 (11)[a] |
| **Admission characteristics of patients** | | | | | |

| Variables | Total | Acute Illness Phenotypes | | | |
|---|---|---|---|---|---|
| | | Phenotype A | Phenotype B | Phenotype C | Phenotype D |
| Emergent admission, n (%) | 30,177 (73) | 7,257 (95)[a] | 11,764 (86)[a] | 8,064 (63) | 3,092 (43)[a] |
| Transfer from different hospital, n (%) | 7,115 (17) | 1,986 (26)[a] | 3,014 (22)[a] | 1,087 (8) | 1,028 (14)[a] |
| **Diagnostic groups of primary admissions** | | | | | |
| Circulatory system diseases | 7,719 (19) | 1,934 (25)[a] | 2,425 (18)[a] | 1,834 (14) | 1,526 (21)[a] |
| Infectious and respiratory diseases | 3,306 (8) | 961 (13)[a] | 1,161 (8)[a] | 705 (5) | 479 (7)[a] |
| Childbirth and pregnancy complications | 3,148 (8) | 391 (5)[a] | 1,147 (8)[a] | 1,248 (10) | 362 (5)[a] |
| Digestive &genitourinary system diseases | 5,184 (12) | 789 (10)[a] | 1686 (12)[a] | 1,876 (15) | 833 (11)[a] |
| Musculoskeletal, connective tissue, and skin diseases | 3,651 (9) | 317 (4)[a] | 1042 (8)[a] | 1,222 (9) | 1,070 (15)[a] |
| Neoplasms | 2,743 (7) | 93 (1)[a] | 665 (5)[a] | 1,138 (9) | 847 (12)[a] |
| **Clinical biomarkers and interventions within 24 hours of admission** | | | | | |
| Surgery on day admitted, n (%) | 8,644 (21) | 272 (4)[a] | 813 (6)[a] | 3,466 (27) | 4,093 (57)[a] |
| ICU or IMC admission within initial 24 hours, n (%) | 9,426 (23) | 2,372 (31)[a] | 1,979 (14) | 1,921 (15) | 3,154 (44)[a] |
| **Cardiovascular system** | | | | | |
| Hypotension (MAP < 60 mmHg) at any point, n (%) | 14,470 (35) | 2,234 (29)[a] | 2,903 (21)[a] | 4,445 (34) | 4,888 (67)[a] |
| Duration, median (IQR), # of minutes | 57 (15, 168) | 86 (30, 224)[a] | 92 (30, 233)[a] | 33 (10, 120) | 37 (10, 129) |
| Vasopressors used, n (%) | 7,531 (18) | 421 (6)[a] | 641 (5)[a] | 2633 (20) | 3836 (53)[a] |
| Outside of the operating room | 1,403 (3) | 242 (3)[a] | 146 (1)[a] | 229 (2) | 786 (11)[a] |
| Hypertension (SBP > 160 mmHg) at any point, n (%) | 14,838 (36) | 2,923 (38)[a] | 3,684 (27)[a] | 4,272 (33) | 3,959 (55)[a] |
| Duration, median (IQR), # of minutes | 120 (27, 356) | 174 (52, 445)[a] | 214 (73, 477)[a] | 114 (19, 336) | 44 (9, 165)[a] |
| Troponin, tested, n (%) | 14,616 (35) | 4,502 (59)[a] | 4,862 (35)[a] | 3,055 (24) | 2,197 (30)[a] |
| Abnormal result in those tested, n (%) | 3,398 (23) | 1,109 (25)[a] | 884 (18) | 585 (19) | 820 (37)[a] |
| **Respiratory system** | | | | | |
| Maximum administered FiO2, median (IQR), % | 0.21 (0.21, 0.40) | 0.21 (0.21, 0.29)[a] | 0.21 (0.21, 0.28)[a] | 0.21 (0.21, 0.40) | 0.40 (0.29, 0.40)[a] |
| Room air only, n (%) | 23,963 (58) | 4,615 (60) | 10,130 (74)[a] | 7,874 (61) | 1,344 (19)[a] |
| 0.22–0.40, n (%) | 14,790 (36) | 2,496 (33)[a] | 3,252 (24)[a] | 4,484 (35) | 4,558 (63)[a] |
| > 0.40, n (%) | 2,749 (7) | 536 (7)[a] | 328 (2)[a] | 543 (4) | 1,342 (19)[a] |

| Variables | Total | Acute Illness Phenotypes | | | |
|---|---|---|---|---|---|
| | | Phenotype A | Phenotype B | Phenotype C | Phenotype D |
| $P_aO2/FiO_2$, tested with arterial blood gas, n (%) | 6,113 (15) | 1,352 (18)[a] | 1,033 (8)[a] | 1,273 (10) | 2,455 (34)[a] |
| <200 in those tested, n (%) | 2,265 (37) | 496 (37) | 301 (29) | 430 (34) | 1,038 (42)[a] |
| Mechanical ventilation, n (%) | 2,123 (5) | 434 (6)[a] | 191 (1)[a] | 314 (2) | 1,184 (16)[a] |
| **Acid-base and kidney status** | | | | | |
| Preadmission estimated glomerular filtration rate[c] (mL/min per 1.73 $m^2$), median (IQR) | 95 (78, 111) | 93 (74, 109)[a] | 96 (77, 111)[a] | 97 (80, 113) | 93 (79, 106)[a] |
| Maximum/reference creatinine[c] ratio, mean (SD) | 1.24 (0.66) | 1.30 (0.70)[a] | 1.22 (0.59)[a] | 1.20 (0.67) | 1.28 (0.74)[a] |
| Renal replacement therapy, n (%) | 641 (2) | 160 (2)[a] | 168 (1) | 175 (1) | 138 (2)[a] |
| Maximum anion gap, median (IQR), mmol/L | 14 (12, 17) | 15 (12, 18)[a] | 14 (12, 16) | 14 (11, 16) | 15 (12, 18)[a] |
| Arterial blood gas tested, n (%) | 6,115 (15) | 1,353 (18)[a] | 1,033 (8)[a] | 1,274 (10) | 2,455 (34)[a] |
| pH < 7.3 among tested, n (%) | 1437 (23) | 298 (22) | 160 (15) | 242 (19) | 737 (30)[a] |
| Maximum base deficit among tested, mean (SD), mmol/L | 4.8 (4.7) | 5.4 (5.0)[a] | 4.5 (4.6) | 4.4 (4.3) | 4.8 (4.6) |
| Lactate, tested, n (%) | 15,447 (37) | 3,935 (51)[a] | 4,308 (31)[a] | 3,706 (29) | 3,498 (48)[a] |
| 2 – 4 mmol/L among tested, n (%) | 3,739 (24) | 978 (25) | 956 (22) | 870 (23) | 935 (27)[a] |
| > 4 mmol/L among tested, n (%) | 1,374 (9) | 331 (8)[a] | 207 (5) | 216 (6) | 620 (18)[a] |
| **Inflammation** | | | | | |
| Maximum white blood cell count, median (IQR), $x10^9/L$ | 9 (7, 13) | 9 (7, 13) | 9 (6, 12)[a] | 9 (7, 12) | 11 (8, 15)[a] |
| Maximum premature neutrophils (bands), median (IQR), % | 10 (4, 20) | 12 (4, 22)[a] | 7 (3, 15) | 9 (3, 18) | 15 (7, 26)[a] |
| Minimum lymphocytes, median (IQR), % | 16 (9, 24) | 15 (8, 24)[a] | 16 (10, 25) | 17 (10, 26) | 10 (6, 18)[a] |
| C-reactive protein, tested, n (%) | 5,862 (14) | 1,246 (16)[a] | 2,396 (17)[a] | 1,759 (14) | 461 (6)[a] |
| Maximum C-reactive protein, median (IQR), mg/L | 18 (5, 77) | 20 (5, 89)[a] | 18 (5, 73)[a] | 15 (4, 70) | 39 (7, 112)[a] |
| Erythrocyte sedimentation rate, tested, n (%) | 3,903 (9) | 775 (10) | 1,591 (12)[a] | 1,253 (10) | 284 (4)[a] |
| Maximum erythrocyte sedimentation rate, median (IQR), mm/h | 40 (19, 73) | 42 (18, 72) | 41 (20, 75) | 39 (19, 71) | 32 (15, 66) |
| Maximum temperature, mean (SD), Celsius | 37.7 (0.6) | 37.7 (0.6) | 37.6 (0.6)[a] | 37.7 (0.6) | 37.9 (0.6)[a] |
| 38 - 39, n (%) | 8,633 (21) | 1,519 (20) | 2,238 (16)[a] | 2,486 (19) | 2,390 (33)[a] |
| > 39, n (%) | 1,548 (4) | 354 (5)[a] | 453 (3) | 368 (3) | 373 (5)[a] |

| Variables | Total | Acute Illness Phenotypes | | | |
|---|---|---|---|---|---|
| | | Phenotype A | Phenotype B | Phenotype C | Phenotype D |
| Minimum temperature, mean (SD), Celsius | 36.7 (1.0) | 36.7 (0.8) | 36.8 (0.7)[a] | 36.7 (0.7) | 36.3 (1.7)[a] |
| **Hematologic** | | | | | |
| Minimum hemoglobin, mean (SD), g/dL | 11.5 (2.3) | 11.6 (2.4) | 11.7 (2.3) | 11.6 (2.2) | 10.9 (2.3)[a] |
| Maximum RDW, mean (SD), % | 15.5 (2.1) | 15.6 (2.1)[a] | 15.6 (2.3)[a] | 15.4 (2.1) | 15.3 (1.9) |
| Minimum platelets, median (IQR), x10$^9$/L | 210 (161, 269) | 209 (160, 271)[a] | 216 (165, 277) | 214 (165, 270) | 195 (150, 247)[a] |
| Platelets < 200, n (%), x10$^9$/L | 16,707 (40) | 3,289 (43)[a] | 5,347 (39)[a] | 4,769 (37) | 3,302 (46)[a] |
| < 100 | 2,643 (16) | 526 (7) | 910 (7)[a] | 715 (6) | 492 (7) |
| 100 - 200 | 14,064 (84) | 2,763 (84) | 4,437 (83)[a] | 4,054 (85) | 2,810 (85) |
| International normalized ratio, tested, n (%) | 20,357 (49) | 4,830 (63)[a] | 6,942 (51)[a] | 5,201 (40) | 3,384 (47)[a] |
| >= 2 | 1,836 (9) | 432 (9) | 666 (10) | 433 (8) | 305 (9) |
| **Neurologic** | | | | | |
| Glasgow Coma Scale score, n (%) | | | | | |
| Moderate neurologic dysfunction (9 - 12) | 1,708 (4) | 385 (5)[a] | 340 (2) | 284 (2) | 699 (10)[a] |
| Severe neurologic dysfunction (<= 8) | 1,482 (4) | 359 (5)[a] | 144 (1)[a] | 207 (2) | 772 (11)[a] |
| **Liver and metabolic** | | | | | |
| Bilirubin tested, n (%), mg/dL | 21,183 (51) | 4,759 (62)[a] | 8,018 (58)[a] | 5,894 (46) | 2,512 (35)[a] |
| ≥ 2 | 1,427 (7) | 271 (6) | 542 (7) | 395 (7) | 219 (9)[a] |
| Maximum glucose, median (IQR), mg/dL | 126 (104, 170) | 127 (105, 175)[a] | 120 (101, 161) | 123 (101, 164) | 144 (116, 187)[a] |
| Albumin, tested, n (%) | 21,368 (51) | 4,780 (63)[a] | 8,070 (59)[a] | 5,951 (46) | 2,567 (35)[a] |
| < 2.5 | 1,243 (6) | 304 (6)[a] | 419 (5) | 260 (4) | 260 (10)[a] |
| 2.5 - 3.5 | 6,904 (32) | 1,678 (35)[a] | 2,515 (31)[a] | 1,719 (29) | 992 (39)[a] |

Abbreviations: ICU: intensive care unit; IMC: intermediate care unit; MAP: mean arterial pressure; SBP: systolic blood pressure; RDW: red cell distribution width; SD: standard deviation; IQR: interquartile range.

[a] The p values represent significant differences (p< 0.05) compared to phenotype C, adjusted for multiple comparisons using the Bonferroni method. The supplementaty tables provide p values for every within-group comparison.

[b] For the consideration of cardiovascular disease, the following were taking into consideration: history of congestive heart failure, peripheral vascular disease, and coronary artery disease.

[c] Race correction was not applied to derive the reference glomerular filtration rate and reference creatinine (refer to eMethods for further information).

**Table 2. Phenotype illness severity, resource use, and clinical outcomes.**

| **Variables** | **Total** | **Acute Illness Phenotypes** | | | |
|---|---|---|---|---|---|
| | | Phenotype A | Phenotype B | Phenotype C | Phenotype D |
| Number of encounters (%) | 41,502 | 7,647 (18) | 13,710 (33) | 12,901 (31) | 7,244 (17) |
| **Acuity scores, first 24 hours of admission** | | | | | |
| SOFA score > 6, n (%) | 3,506 (8) | 656 (9)[a] | 508 (4)[a] | 768 (6) | 1,574 (22)[a] |
| ICU/IMC patients, SOFA score ≤ 6, n (%) | 6,882 (17) | 1,822 (24)[a] | 1,690 (12) | 1,514 (12) | 1,856 (26)[a] |
| ICU/IMC w/ SOFA score > 6, n (%) | 2,544 (6) | 550 (7)[a] | 289 (2)[a] | 407 (3) | 1,298 (18)[a] |
| Ward w/ SOFA score ≤ 6, n (%) | 31,114 (75) | 5,169 (68)[a] | 11,512 (84)[a] | 10,619 (82) | 3,814 (53)[a] |
| Ward w/ SOFA score > 6, n (%) | 962 (2) | 106 (1)[a] | 219 (2)[a] | 361 (3) | 276 (4)[a] |
| MEWS score ≥ 5, n (%) | 2,828 (7) | 873 (11)[a] | 575 (4)[a] | 387 (3) | 993 (14)[a] |
| ICU/IMC w/ MEWS score ≤ 4, n (%) | 7,235 (17) | 1,703 (22)[a] | 1,618 (12) | 1,643 (13) | 2,271 (31)[a] |
| ICU/IMC w/ MEWS score > 4, n (%) | 2,191 (5) | 669 (9)[a] | 361 (3) | 278 (2) | 883 (12)[a] |
| Ward w/ MEWS score ≤ 4, n (%) | 31,439 (76) | 5,071 (66)[a] | 11,517 (84) | 10,871 (84) | 3,980 (55)[a] |
| Ward w/ MEWS score > 4, n (%) | 637 (2) | 204 (3)[a] | 214 (2)[a] | 109 (1) | 110 (2)[a] |
| **Resource use throughout hospitalization** | | | | | |
| Days in hospital, median (IQR) | 4 (2, 7) | 4 (2, 7)[a] | 4 (2, 7)[a] | 3 (2, 6) | 4 (3, 7)[a] |
| Surgery during hospital stay, n (%) | 11,634 (28) | 860 (11)[a] | 2,006 (15)[a] | 4,452 (35) | 4,316 (60)[a] |
| ICU/IMC[b] admission, n (%) | 11,121 (27) | 2,700 (35)[a] | 2,673 (19) | 2,446 (19) | 3,302 (46)[a] |
| Days in ICU/IMC[c], median (IQR) | 4 (2, 7) | 4 (3, 7)[a] | 4 (3, 7)[a] | 4 (2, 6) | 4 (3, 8)[a] |
| More than 48 hrs in ICU/IMC, n (%) | 8,332 (75) | 2,068 (77)[a] | 2,008 (75)[a] | 1,751 (72) | 2,505 (76)[a] |
| Mechanical ventilation, n (%) | 3,218 (8) | 695 (9)[a] | 554 (4)[a] | 628 (5) | 1341 (19)[a] |
| Hours on mechanical ventilation hours, median (IQR)[d] | 35 (14, 113) | 44 (17, 127)[a] | 35 (13, 116) | 25 (11, 101) | 33 (14, 113) |
| More than 48 hrs on mechanical ventilation, n (%) | 1,661 (52) | 403 (58)[a] | 284 (51) | 302 (48) | 672 (50) |
| Renal replacement therapy, n (%) | 1,262 (3) | 314 (4)[a] | 396 (3) | 322 (2) | 230 (3)[a] |
| **Complications** | | | | | |
| Acute kidney injury, n (%) | 6905 (17) | 1,598 (21)[a] | 2,279 (17)[a] | 1,680 (13) | 1,348 (19)[a] |
| Community-acquired AKI, n (%) | 3839 (56) | 924 (58) | 1,200 (53) | 897 (53) | 818 (61)[a] |

| Variables | Total | Acute Illness Phenotypes | | | |
|---|---|---|---|---|---|
| | | Phenotype A | Phenotype B | Phenotype C | Phenotype D |
| Hospital-acquired AKI, n (%) | 3066 (44) | 674 (42) | 1,079 (47) | 783 (47) | 530 (39)[a] |
| Worst AKI staging, n (%) | | | | | |
| Stage 1 | 4360 (63) | 961 (60)[a] | 1,479 (65) | 1,112 (66) | 808 (60)[a] |
| Stage 2 | 1362 (20) | 346 (22)[a] | 425 (19) | 300 (18) | 291 (22) |
| Stage 3 | 848 (12) | 206 (13) | 270 (12) | 202 (12) | 170 (13) |
| Stage 3 with RRT | 335 (5) | 85 (5) | 105 (5) | 66 (4) | 79 (6) |
| Venous thromboembolism, n (%) | 1257 (3) | 261 (3)[a] | 481 (4)[a] | 334 (3) | 181 (2) |
| Sepsis, n (%) | 3750 (9) | 1,049 (14)[a] | 1,102 (8)[a] | 754 (6) | 845 (12)[a] |
| Hospital disposition, n (%) | | | | | |
| Hospital mortality | 1141 (3) | 294 (4)[a] | 278 (2)[a] | 184 (1) | 385 (5)[a] |
| Different, LTAC, SNF, or Hospice | 4475 (11) | 1,140 (15)[a] | 1,591 (12)[a] | 932 (7) | 812 (11)[a] |
| Short-term rehabilitation or home | 35886 (86) | 6,213 (81)[a] | 11,841 (86)[a] | 11,785 (91) | 6,047 (83)[a] |
| 30-day mortality, n (%) | 1633 (3.9) | 439 (6)[a] | 458 (3)[a] | 278 (2) | 458 (6)[a] |
| Three-year mortality, n (%) | 8013 (19) | 1,892 (25)[a] | 2,861 (21)[a] | 1,975 (15) | 1,285 (18)[a] |

Abbreviation: ICU: intensive care unit; IMC: intermediate care unit; IQR: interquartile range; MEWS: modified early warning score; SOFA: sequential organ failure assessment.

[a] The p values represent significant differences ($p < 0.05$) compared to phenotype C, adjusted for multiple comparisons using the Bonferroni method. The supplemental tables provide p values for all within-group comparisons.

[b] These calculated values were derived from data collected at any point during the hospitalization period.

[c] These calculated values were derived for patients who were admitted to the ICU or IMC.

[d] These calculated values were derived for patients who required mechanical ventilation.

**Patterns of vital signs**

In order to determine which vital signs had the most notable effect on cluster designations, we compared the standardized mean differences between pairs of phenotypes (Figure 3). The smallest contributors to the differences in phenotypes were temperature and oxygen saturation. Respiratory rate and heart rate differed significantly across phenotypes except for C and D, which manifested as differences in temperature.

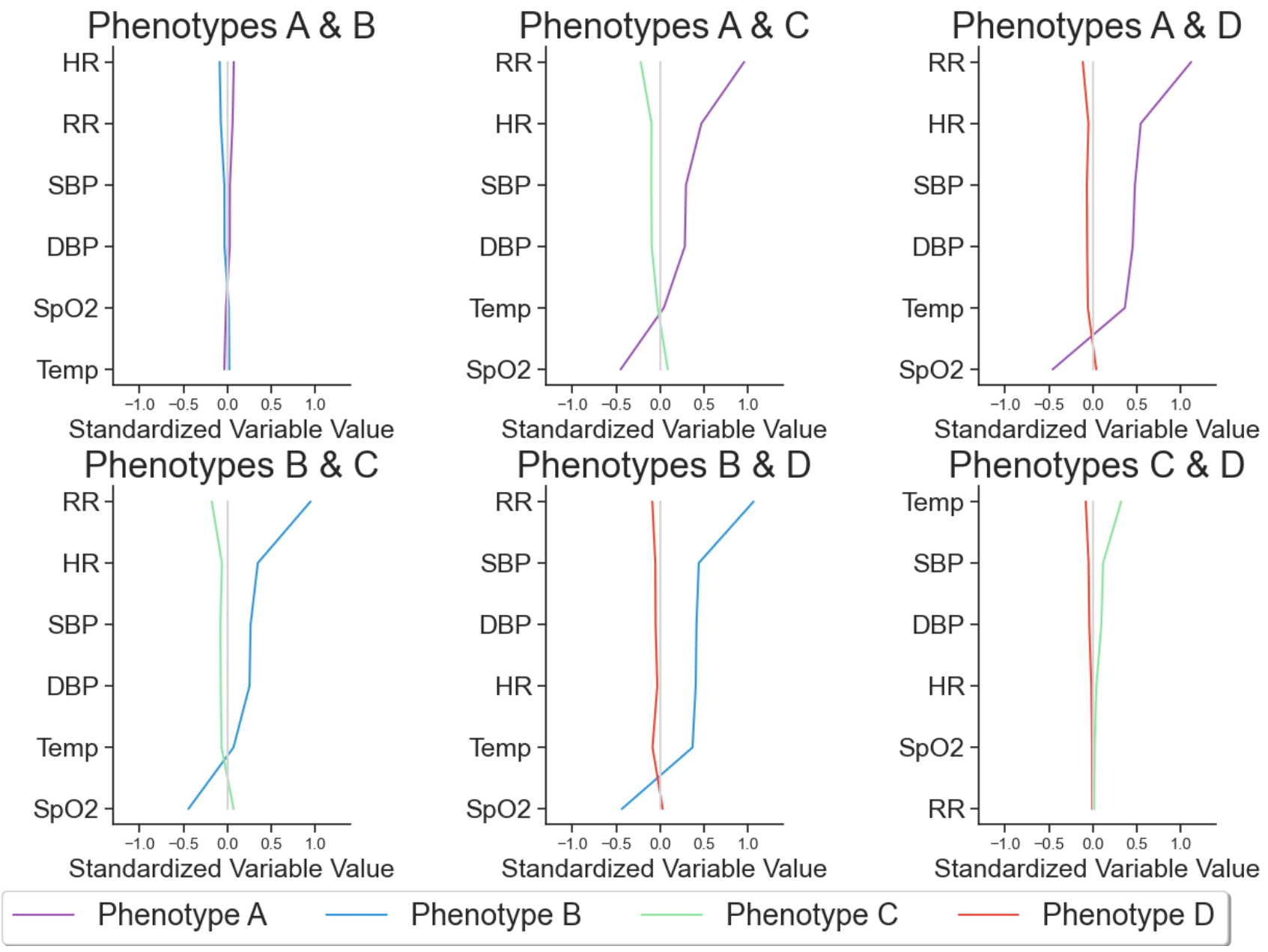

**Figure 3. Contributions of vital signs to cluster assignments.** The pairwise phenotype comparisons of vital sign values, which have been standardized to a mean of 0 and standard deviation of 1. The comparison reveals that temperature and oxygen saturation contribute the least to differences between phenotypes. Conversely, respiratory rate and heart rate exhibit considerable variation across all phenotypes, with the exception of phenotypes C and D, where these vital signs appear more consistent. Temp: temperature; SpO2: peripheral capillary oxygen saturation; DBP: diastolic blood pressure; SBP: systolic blood pressure; RR: respiratory rate; HR: heart rate.

**Relationship with organ support**

The association between phenotypes and the highest SOFA score recorded within 24 hours after admission is depicted in eFigure 5; the SOFA components for every phenotype are depicted with chord diagrams in eFigure 6. The highest percentages of patients with cardiovascular and respiratory dysfunction were found in Phenotypes C and D; however, every phenotype had significant percentages of patients from the entire spectrum of SOFA scores and component subscores; the clustering into phenotypes did not only restate prior SOFA acuity assessments.

**Relationship with survival probabilities**

Three-year survival adjusted for demographics and comorbidities (Figure 4, eFigures 7A and 7B), was significantly lower for males (HR 1.4, 95% CI 1.4–1.5) and for patients 65 years or older (HR 2.8, 95% CI 2.6–2.9). With phenotype C as reference, survival probability was lower for phenotype A (HR 1.8, 95% CI 1.6–1.9), B (HR 1.5, 95% CI 1.4–1.6), and D (H 1.2, 95% CI 1.1–1.3, all p<0.001). Similar three-year survival probability adjusted for additional SOFA score was modeled (eFigures 7C and 7D), demonstrating strong associations between higher SOFA score and lower survival probability (SOFA 2–4: HR 1.8, 95% CI 1.6-1.9; SOFA 5 or greater: HR 3.1, 95% CI 2.8–3.3, all p<0.001). After being adjusted for SOFA, phenotypes A and B had much stronger associations with lower survival probability (HR 1.9, 95% CI 1.8–2.1; HR 1.7, 95% CI 1.6–1.9, all p<0.001), but phenotype D had a stronger association with improved survival probability (HR 0.9, 95% CI 0.8–1.0, p=0.035).

Three-year survival probability for clinical entities of sepsis, AKI, and surgical patients were modeled after adjusting for demographics and comorbidities (eFigures 8–10). Using phenotype C as a reference, the survival probability for patients with sepsis was found to be lower for phenotype A (HR 1.4, 95% CI 1.2–1.7, p<0.001), B (HR 1.1, 95% CI 0.9–1.3, p=0.27), and D (HR 1.7, 95% CI 1.4–2.0, p<0.001); for patients with AKI, probability of survival was lower for phenotype A (HR 1.3, 95% CI 1.1–1.4, p<0.001), B (HR 1.1, 95% CI 1.0–1.3, p=0.052), and D (HR 1.4, 95% CI 1.2–1.6, p<0.001); for patients undergoing surgery, probability of survival was lower for phenotype A (HR 2.1, 95% CI 1.7–2.6) and B (HR 1.8, 95% CI 1.6–2.2, all p<0.001), but higher for phenotype D (HR 0.9, 95% CI 0.8–1.1, p=0.349).

**Reproducibility**

In the training and testing cohorts, the percentage of patients in each phenotype was stable (phenotype A: 18% and 18%; phenotype B: 33% and 35%; phenotype C: 31% and 31%; phenotype D: 17% and 16%). Phenotypes were reproducible in the testing cohort. Within the testing cohort, the phenotypes were similar to the training cohort in terms of clinical characteristics, biomarkers, and patient outcomes (eFigures 11 and 12, eTables 8 and 9). Across the training and testing cohorts, there were similar distributions of SOFA scores, survival scores, and diagnosis groups (eFigures 7 and 13–17).

**Evaluation of representation learning of deep interpolation network**

The efficacy of the dTIC network in learning cluster-friendly feature representations from sparse, irregularly sampled time series data was clearly depicted through a visual representation of acuity illness phenotypes using t-SNE (Figures 2B and 2C).

The reconstruction error of physiologic signatures, presented in eFigures 18 and 19 demonstrated that the dTIC network can accurately regenerate the input physiologic signatures across all vital signs, where the disparity between the observed and reconstructed data was negligible.

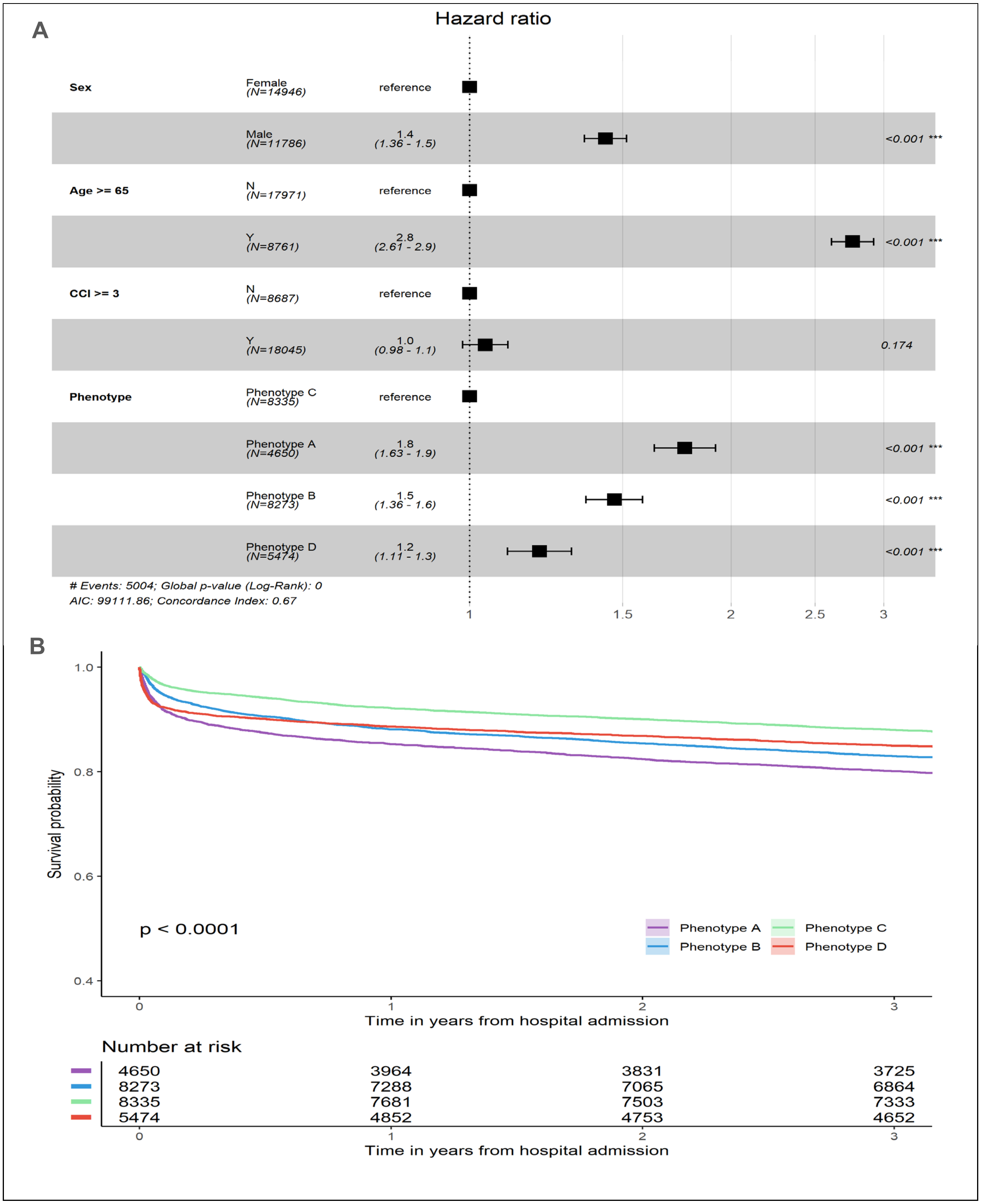

**Figure 4. Survival curves and adjusted Cox proportional hazards modeling.** (A) The survival curves for each phenotype, considering adjustments for both demographic information and comorbidities. (B) The adjusted Cox proportional hazards models, incorporating demographic information and comorbidities into the analysis. CCI: Charlson Comorbidity Index.

**Discussion**

To the best of our knowledge, this is the first instance of using deep interpolation network clustering to phenotype a diverse cohort of hospitalized patients based on early vital sign measurements. Although previous studies have demonstrated the efficacy of clustering in identifying patient subgroups within larger cohorts with similar clinical presentations, such as sepsis and diastolic heart failure,[10,31] this novel approach of employing a deep interpolation network is unique.

Our group has also previously employed consensus clustering to early vital sign measurements and distinguished four phenotypes, albeit without the use of a deep interpolation network[19]. In that study, each vital sign sequence was resampled on an hourly basis by averaging the multiple measurements within a one-hour window and imputing the missing vital signs through forward and backward propagation. Resampling techniques for processing irregularly sampled vital signs could introduce bias, however, since the frequency of measurements and dynamic patterns of vitals within each one-hour window is ignored. Unlike the previous approach, this present study fully capitalizes on irregularly sampled time series circumventing the error-prone resampling. Both studies identified a phenotype with highest incidence of prolonged respiratory insufficiency, sepsis, AKI, and three-year mortality. Moreover, both identified a phenotype exhibiting early and persistent hypotension along with a need for early surgery. Several differences emerged, however, when comparing the identified phenotypes, especially in our study. Phenotype D exhibited worse short-term clinical outcomes. These comparative findings suggest that deep representation of irregularly sampled early vital signs can potentially unveil diverse patient subgroups, differing from what conventional machine learning approaches might reveal.

Other researchers have also investigated patient phenotyping. Seymour et al.[10] performed clustering analyses on septic patients, hypothesizing that the pathophysiology of sepsis is inherently heterogeneous and recognizing distinct sepsis phenotypes may facilitate the provision of targeted therapies. This rationale is supported by the inconclusiveness of most sepsis drug trials. Their clustering was performed on both clinical variables and immune response biomarkers, resulting in the identification of four distinct clusters. They conducted simulations in which varying proportions of each cluster were introduced to previously reported randomized controlled trials, indicating unique treatment responses across different clusters. Similarly, Shah et al.[32] executed clustering analyses for patients with heart failure and preserved ejection fraction. Shah et al. used a combination of echocardiogram and electrocardiogram data in addition to clinical variables for clustering, identifying three distinct phenotypes with unique clinical outcomes. This was true even after accounting for traditional risk factors. Such findings suggest that clustering methods have the capacity to identify phenotypic subgroups of patients that traditional clinical parameters might not be able to identify, and these subgroups could potentially exhibit different treatment responses and clinical outcomes.

The present study employs a novel method to optimally cluster hospitalized patients into four distinct phenotypes using a limited set of early vital sign data gathered shortly after hospital admission. The identified generalized patterns relating to patient acuity and trajectory—even before an established diagnosis was made—may have significant clinical implications for patient triage and targeted care strategies. Among the four identified phenotypes, phenotype D appeared to represent patients who were demonstrably ill upon admission and received appropriate level of care with surgical source control or correction of their underlying pathology, consequently exhibited improved 30-day and three-year survival. Phenotype A also appeared to have a high degree of illness severity, though this was likely related to acute on chronic exacerbation of underlying comorbidities. These patients typically present a clinical challenge to their healthcare providers. Phenotypes B and C had less acuity of illness and were most often admitted to general wards. Phenotype C underwent surgery more often, possibly representing patients with urgent, but not emergent, correctable surgical and medical pathology. Phenotype B, much like phenotype A, may include patients with chronic comorbidities though of less severity.

Our study has several limitations. First, the potential to generalize our findings may be somewhat limited due to our study's reliance on data from a single institution. While it is recognized that patient populations can vary across different institutions, we argue that vital signs, being a direct expression of physiological status, would maintain consistency across diverse healthcare contexts. Second, for this study, we confined our input features to vital signs from the first six hours following hospital admission. It is important to note that critical lab results and imaging findings, which can substantially impact patient clustering, are frequently available within this timeframe. Future iterations of our model will explore the potential advantages of incorporating such data for a more comprehensive understanding of patient physiology. Finally, the ability of early clustering to augment clinical prognostication and decision-making, though promising, remains largely theoretical until it is evaluated in a prospective clinical trial setting.

## Conclusions

In this study, we developed and evaluated a novel deep temporal interpolation and clustering network for extracting latent representations from sparse and irregularly sampled time-series data—specifically, vital sign measurements obtained during the first six hours of hospital admission—and identified four distinct patient phenotypes. Each phenotype exhibited unique pathophysiological signatures and associated clinical outcomes, and did not simply recapitulate known, recognized clinical phenotypes, such as SOFA score. Our algorithm has the potential to significantly enhance early clinical decision-making, such as triage decisions, especially in situations where data availability is limited. Future efforts will focus on incorporating this model with historical patient data and additional elements from the EHR, along with external validation of these findings in clinical trials.

**Author contributions**

AB conceived the original idea for the study, and sought and obtained funding. TOB and AB had full access to all the data in the study and take responsibility for the integrity of the data and the accuracy of the data analysis. Analyses: YR and YL. Interpretation of data: All authors. The article was written by YR, YL, TL, TOB, and AB with input from all coauthors. All authors participated in critically revising the manuscript for important intellectual content and gave final approval of the version to be published. AB and TOB served as senior authors. AB is guarantor for this article.

**Competing interests**

The authors declare no competing interests.

**Data Availability**

The vitals data used in this analysis include both date and time stamps. In order to prevent patient privacy compromises because of identifiers in the data, our data cannot be publicly shared in a repository. Upon reasonable request, data can be shared by the University of Florida Intelligent Critical Care Center at ic3-center@ufl.edu and the University of Florida Integrated Data Repository at IRBDataRequest@ahc.ufl.edu.

# Supplementary Online Content

## eMethods

### A. Data source and participants and study design

#### Data source and participants

This project was approved by the University of Florida institutional review board under a waiver of informed consent and with authorization under the Health Insurance Portability and Accountability Act. Transparent Reporting of a multivariable prediction model for Individual Prognosis Or Diagnosis (TRIPOD) recommendations were followed under the Type 2b analysis category (nonrandom split-sample development and validation). Using the University of Florida Health (UFH) Integrated Data Repository as Honest Broker, we created a longitudinal dataset from electronic health records of all adults (age ≥18 years) admitted to the 1000-bed academic hospital at UFH between June 1, 2014 and April 1, 2016. The dataset includes structured and unstructured clinical data, demographic information, vital signs, laboratory values, medications, diagnoses, and procedures. Patients completely missing at least two of the six vital sign measurements used for clustering (systolic and diastolic blood pressure, heart rate, respiratory rate, temperature, and peripheral capillary oxygen saturation) in the first six hours of admission were excluded from the analysis (eFigure 1). The final cohort consisted of 75,762 hospital admissions for 43,598 patients.

#### Study design

We non-randomly split the dataset by admission dates into three cohorts: training (admissions between June 1, 2014 and May 31, 2015, n = 41,502, 55% of all admissions), validation (admissions between June 1, 2015 and October 31, 2015, n = 17,415, 22% of all admissions), and testing (admissions between November 1, 2015 and April 1, 2016, n = 16,845, 23% of all admissions). To determine acute illness phenotypes using early physiologic signatures, we derived the clinical phenotypes using unsupervised clustering methods that were applied to the repeated measurements of six vital signs available within the first six hours of hospital presentation in the training cohort. We selected hyper-parameters of clustering model using validation cohort. We assessed phenotype reproducibility by predicting phenotypes in the testing cohort and assessing phenotype frequency distributions and clinical outcomes.

### B. Approach to preprocess electronics health records (EHR) data

EHR vital data elements in our cohort studies were irregularly sampled time series. Prior to clustering algorithms, we excluded outliers based on the expert-defined ranges (eTable 1). For the time series missing entirely, which is due to having no measurements during the hospitalization in the plausible range for a variable, we assigned the starting point (time t=0) value of the time series to the mean value of corresponding variables in the training cohort, as listed in eTable 1. We standardized values using Min-Max scaler (eTable 1). We directly input these irregular sampled time series to our clustering algorithm without any time interval imputation.

### C. Deep Interpolation Network

In this section, we describe our proposed Deep Temporal Interpolation and Clustering Network (dTIC) for clustering the patients based on their vital sign data during the early stages of hospital admission. Using the raw sparse and irregularly sampled time series vital sign as the input, dTIC can automatically extract a unified and abstract representation of the entire time-series data of an encounter and cluster the patients via an end-to-end unsupervised manner. The overall network architecture consists of four main compounds: Interpolation model, Seq2Seq model, Re-interpolation model and Clustering model.

dTIC learns the feature representations and cluster assignments via a two-pass manner. In the first pass, dTIC learns the feature representations, determines the optimal number of clusters and thereby generates the initial cluster assignments. In the second pass, dTIC simultaneously learns feature representations and cluster assignments to refine the results.

Figure 1a provides the schematic representation of the dTIC architecture for feature learning process in the first learning pass. We first interpolate the raw time-series vital sign data to a regularly sampled meta-representation with pre-defined reference time points via an interpolation model[1]. Then we feed the interpolated time-series data into a Seq2Seq model with Gated Recurrent Unit (GRU)[2] layers for feature embedding and extracting a unified context vector lying in the low-dimensional feature space by the encoder. The context vector contains the global time-series information and is further used by other downstream tasks (e.g., clustering, classification). The decoder in the Seq2Seq model learns from the context vector and outputs the time-series data with the same length of the Seq2Seq model's input. Then, we deploy a radial basis function network-based model[3] to re-interpolate the fixed-length output to the raw irregular time points for reconstructing the raw vital signs data at corresponding time points. In order to enhance the feature representation, we also make an auxiliary prediction task via a linear neural network in which stacked real and synthetic time series data are input into the interpolation network and the classifier predicts if the learned context vector is from real data. The full feature extraction model is end-to-end trained by minimizing the reconstruction loss measured with mean square error (real data only) and classification loss measured with binary cross-entropy. The extracted feature representation (context vector) will be used to determine the optimal number of clusters and obtain the initial cluster assignment.

Figure 1b provides the overall schematic representation of the dTIC architecture for feature representation learning and cluster assignment process. In addition to the feature extraction model, a clustering network is added in the second learning pass. The clustering network computes a soft assignment between the embedded points and the cluster centroids and matches the soft assignment to the target distribution in order to simultaneously improve clustering assignment and feature representation.[4] The full dTIC model is end-to-end trained by minimizing the reconstruction loss measured with mean square error (real data only), classification loss measured with binary cross-entropy, and clustering loss measured with Kullback-Leibler (KL) divergence between an embedded distribution and a target data distribution. We describe the components of the dTIC in detail in the following subsections.

**Interpolation Model**

It is common that the time series vital sign data in electronic health records to be both sparse and irregularly sampled, which means large and irregular intervals widely exist between the data observation time points. Such sparsity and irregularity pose a significant challenge for machine / deep learning techniques to analyze the crucial vital sign data for improving the human health outcome. To deal with this problem, we adopt the network proposed by Shukla and Marlin [1] first to interpolate the raw time-series data to a regularly sampled meta-representation with pre-defined reference time points.

In our study, we utilize six vital signs multivariate time series data, e.g., two kinds of blood pressure (systolic and diastolic), heart rate, temperature, Spo2, and respiratory rate. Take one variable out of six as an example. For one patient, the raw time-series data is denoted as $e = \{(t_i, x_i)|i = 1, \ldots, I\}$, where $I$ represents the total number of observations, $t_i$ is the time point, and $x_i$ is the corresponding observed value. The time intervals between adjacent observation time points vary a lot. The interpolation model can map irregular $e$ value to the regular time series data which is defined at the $T$ reference time points $r = [r_1; \ldots; r_T]$ with evenly spaced interval.

The interpolation model consists of two layers, where the first layer separately performs the interpolation for each variable, and the second layer aggregates the information across all the studied variables. The model generates three different channel groups at each reference time point, which respectively represents smooth trends $\chi$, short time-scale transients $\tau$, and local observation frequencies $\lambda$. The interpolation model enables the single observation data point to be considered by all the reference time points and allows for the information to be shared across multiple variables. For more detailed interpolation mathematic denotation, the reader is referred to [1].

**Seq2Seq Model**

With the interpolated time-series data as the input, we develop a Seq2Seq model to learn its low-dimensional representation, which can embed the contextual information over the full timeline. Seq2Seq model is a method of the encoder-decoder framework that maps an input of sequence to an output of sequence, and it is broadly used in machine translation, text summarization, conversational modeling, and some other tasks. With a single layer GRU network[2] as the encoder, the input sequence is encoded to a fixed-length contextual vector $h_T$, which is the hidden state of the last time step. The hidden state of GRU updating mechanism, illustrated in the following equation, ensures that every internal hidden node state will be calculated by the previous state $h_{i-1}$ and current time step input $(\chi_i\ \tau_i,\ \lambda_i)$.

$$h_i = GRU_{Enc}((\chi_i\ \tau_i,\ \lambda_i),\ h_{i-1})$$

A single-layer GRU network is also used for a decoder. At each time step, the decoder updates its current hidden state $s_t$ with the concatenated features incorporating the previous decoded output $o_{t-1}$ and global context vector $h_T$ as the input:

$$s_t = GRU_{Dec}([o_{t-1};\ h_T], s_{t-1})$$

**Re-Interpolation Model**

To unsupervised learn the useful representation, a common strategy is to build an autoencoder learning framework by reconstructing the input itself from the extracted bottleneck representation. Therefore, on top of the Seq2Seq model, we develop a re-interpolation network to map the output with the evenly spaced intervals to the raw irregular time points. Similar to the interpolation model, the transformation is also based on a radial basis function network. Our re-interpolation model allows the embedded values at every reference time point to make a continuous contribution to reconstructed values at all the raw time points, but the contribution weight is exponentially decayed in terms of the distance between the referenced time point $r_i$ and target time point $t_j$:

$$w(r_i, t_j, \theta) = exp(-\theta(r_i - t_j)^2)$$

where $\theta$ is learnable network parameters.

After the re-interpolation, we can easily calculate the mean square error at every input time point and minimizing this reconstruction loss is served as the learning objective of the DIN model. It is worth noting that the interpolation, Seq2Seq, and re-interpolation models in the DIN are jointly optimized. Compared with the work [2], it effectively improves the model learning capacity and allows the clustering representation to contain more global information across the full timeline. After the model training, we also visualize the reconstruction performance of the test cohort to verify our model learning capacity.

**Clustering Model**

Taking the low-dimensional feature generated in the first pass, we determine the optimal number of clusters using the combination of phenotype size, Davies-Bouldin index (DBI)[5], silhouette score[6], elbow method[7] and gap statistic method[8]. Once the number of cluster is determined, any standard clustering algorithm can be used to derive initial cluster centroids. In our case, we apply the centroid-based classical k-means clustering.

We improve the feature representations and cluster assignments by a clustering network[4]. The clustering network alternates between 1) computing a soft assignment between the embedded points *i* and the cluster centroids by measuring the similarity between them; 2) minimizing the KL divergence between the soft assignment and the auxiliary distribution. For more detailed description of clustering model, the reader is referred to [4].

**Classifier**

To enhance the feature representation learning, we make two auxiliary prediction tasks: 1) predicting the maximum or minimum vital sign value within the next hour; and 2) predicting if the learned representation is from real time series data after feeding both real and synthetic time series data into the model. We predict minimum systolic and diastolic blood pressure, and peripheral capillary oxygen saturation; maximum heart rate, respiratory rate and temperature within the next hour. We feed both real and synthetic time series data into the model. We generate the synthetic time series data by randomly replacing values at 50% time points. It worth noting that synthetic time series data is only used for classification and is not counted in the optimization of reconstruction loss and clustering loss. We use a linear neural network to do the prediction task.

### D. Data visualization

- Chord plots

Chord diagram is widely used to represent connection and relationship between several entities. We generated two sets of chord diagrams to visualize the patients' distribution regarding different studied variables.

One set of chord diagrams were created to visualize the distribution of phenotypes across worst SOFA scores of six organ systems within first 24 hours of admission. These six organ systems include:

- Cardiovascular
- Respiratory
- Coagulation
- Liver
- Neurologic
- Renal

For each organ system, percent of patients with organ dysfunction, that is with SOFA score of 2 or more were calculated. For each phenotype, the larger percent of patients with higher score of that organ system, the border the ribbon. Phenotypes are shown in separate colors.

The other set of chord diagrams were created to visualize the distribution of nine most common admission diagnosis groups by phenotypes. These most common admission diagnosis groups vary from cohorts, including:

- Nonspecific chest pain
- Abdominal pain
- Complication of device; implant or graft
- Other and unspecific lower respiratory disease
- Septicemia (except in labor)
- Acute cerebrovascular disease
- Cardiac dysrhythmias.
- Congestive heart failure; nonhypertensive
- Malaise and fatigue
- Osteoarthritis
- Other complications of pregnancy

For each phenotype, the larger percentage of patients with that admission diagnosis group, the border the ribbon. Phenotypes are shown in separate colors. Diagrams were generated with Circlize R package[9].

- Alluvial plots

Alluvial plots were generated to visualize distribution of phenotypes across worst Sequential Organ Failure Assessment Score (SOFA) scores of patients within first 24 hours of admission. Phenotypes were grouped in the left column and the total SOFA scores were categorized into 3 levels (0-1, 2-4 and 5+) listed in the right column. Ribbons connect the phenotypes and SOFA categories, which indicates a percentage of patients in a phenotype fall into a particular SOFA category and vice versa. The larger percentage of patients, the boarder the ribbon. Phenotypes are shown in separate colors. Plots were generated with Alluvial R package[10].

- t-SNE plots

t-Distributed Stochastic Neighbor Embedding (t-SNE) is a nonlinear dimensionality reduction technique well-suited for embedding high-dimensional data for visualization in a low-dimensional space. In our work, the t-SNE plots depicted the 2 dimensional feature space of the patients vital signs after reducing their original dimension from 36 to 2 by t-SNE algorithm. Each dot represents a patient, and patients in different phenotypes are colored differently. Plots were generated by scikit-learn t-SNE Python package[11].

- Line plots

Line plots were generated to visualize the time-series vital sign data as it is well-suited for analyzing trends of different variables along time. We created a line plot for each vital sign studied in our work. To better observe the trends of different vital signs, for each encounter, we resampled the raw time series data to 5 minute frequency by averaging multiple measurements every 5 minutes. Then line plots of phenotypes for each vital sign were created by plotting the mean value and 95% confidence interval around the mean. The six vital signs

include:

- o Systolic blood pressure
- o Diastolic blood pressure
- o Heart rate
- o Temperature
- o Blood oxygen saturation
- o Respiratory rate

Phenotypes are shown in separate colors. Plots were created by Seaborn lineplot Python package[12].

### E. Predicting cluster members in new datasets

In the testing cohort, we used a prospective approach to assign phenotype membership to subject based upon clinical characteristics of typical cluster members in the training cohort.

To accomplish this, we first preprocessed the data using the procedure above (B). We then predicted phenotype assignments by calculating the Euclidean distance from each testing cohort admission to the centroid of each phenotype from training cohort. Consider the $i$th subject with $p$ features. We represent it as $X_i = [x_{i1},\ x_{i2},\ \cdots,\ x_{ip}]$. We denote the mean of the kth phenotype with $\mu_k = [\mu_{k1}, \mu_{k2}, \cdots, \mu_{kp}]$ and represent it as the center of the phenotype. Thus, we calculate the Euclidean distance of the ith admission to the center of the kth phenotype, $d_{i,k}$ as:

$$d_{i,k} = \sqrt{\sum_{j=1}^{p} (x_{ij} - \mu_{kj})^2}$$

We calculate distances of all admissions to all phenotype centroids and assigned each admission to its nearest phenotype.

### F. Definition of clinical characteristics

Chronic disease burden was characterized by Charlson-Deyo comorbidity index scores.[13] Chronic kidney disease was determined from medical histories obtained prospectively at the time of enrollment and from a validated combination of International Classification of Diseases codes from electronic health records[14]. Severity of illness was characterized by SOFA and Modified Early Warning Score (MEWS) based on worst values within first 24 hours of hospital admission[15]. Missing SOFA and MEWS scores were imputed with 0.

Measurements for clinical biomarkers that fell outside of expert-defined ranges were considered outliers and were removed from the data. All measurements within 24 hours of hospital admission were used to detect highest or lowest value. Ranges of outliers and directionality of worst values are listed in eTable 2. Only results among patients with measurements were reported. We presented continuous variables as mean (SD) and median values with interquartile ranges and as frequencies and percentages for categorical variables.

For blood pressure, invasive measurements were used, and in absence of invasive measurements at a specific date and timestamp, noninvasive measurements were used. Duration of blood pressure below certain cutoff was determined in minutes after forward-propagating previous values. We identified number of pressors and need for inotrope in the first 24 hours of admission based on medications file where dopamine, droxidopa, midodrine, ephpedrine, epinephrine, norepinephrine, phenylephrine, and vasopressin were considered for vasopressors and dobutamine and milrinone for inotrope. Troponin measurements includes Troponin T and Troponin I. In order to determine FiO2 value at each date and time stamp, formulas were used to imputed FiO2 from oxygen delivery device and corresponding oxygen flow rate.[15] If no oxygen flow rate is given, default FiO2 was imputed based on respiratory device. If oxygen flow rate is outside specified range, minimum and maximum flow rate were used for imputing FiO2. If formula result is greater than maximum per-device FiO2, the maximum FiO2 was imputed. In absence of PaO2 to calculate PaO2/FiO2 ratio, SpO2/FiO2 to PaO2/FiO2 conversion was used[15, 16].

To determine reference creatinine, we used previously validated modification of the NHS England alert algorithm.[17] For patients with available preadmission measurements, reference value was defined as either the lowest in the last 7 days or a median of values from the preceding 8 to 365 days depending on availability of previous results. For patients with no available preadmission measurements and no history of chronic kidney disease (CKD) we used the lowest of admission creatinine and estimated baseline creatinine using the Modification of Diet in Renal Disease Study equation assuming that baseline estimated glomerular filtration rate (eGFR) is 75 ml/min per 1.73 m2. For patients with known history of CKD and no available preadmission measurements we used lowest creatinine value on admission day. After first seven days of hospitalization, minimum serum creatinine measurements in preceding 7 days was used as the reference creatinine. Reference creatinine was used to estimate preadmission reference glomerular filtration rate using Chronic Kidney Disease Epidemiology Collaboration equation.[18] Chronic kidney disease was determined from medical histories obtained prospectively at the time of enrollment and from a validated combination of International Classification of Diseases codes from electronic health records.[18] Chronic kidney disease stages were determined based on reference eGFR according to guidelines[19, 20].

**Diagnosis codes**

We determined category of admission diagnosis codes, which are assigned either as International Classification of Diseases, 9th Revision, Clinical Modification (ICD-9-CM) or International Classification of Diseases, 10th Revision, Clinical Modification (ICD-10-CM) code. We used general equivalence mappings to assist with the conversion ICD-10-CM codes to ICD-9-CM codes[21]. The Clinical Classification Software (CCS)[19] consists of two related classification systems, single-level and multi-level, which are designed to meet different needs. We used multi-level CCS which expands the single-level CCS into a hierarchical system and enables evaluating larger aggregations of conditions and procedures or exploring them in greater detail. The multi-level system has four levels for diagnoses and three levels for procedures, which provide the opportunity to examine general groupings or to assess very specific conditions and procedures. We showed distribution of most common Level 1 and Level 2 codes for each cluster as well as distribution of all admission diagnosis codes that are present in at least 1% proportion of patients.

## G. Definition of clinical outcomes

We determined complications occurring anytime during hospitalization, including infectious and mechanical wound complications (wound complications), acute kidney injury (AKI), mechanical ventilation (MV) and intensive care unit (ICU) admission for greater than 48 hours, cardiovascular (CV) complications, neurological complications and/or delirium, sepsis, and venous thromboembolism (VTE). We used the exact dates and times to calculate the duration of MV, ICU, and hospital stay. In order to determine the duration of invasive mechanical ventilation, we developed an algorithm to identify the start and stop times for ventilation based on flowsheet data. Patient was determined to be on mechanical ventilation at a time point if the respiratory device is recorded as ventilator or endotracheal tube (ETT) or there is a recorded measurement value for tidal volume, end-tidal carbondioxide (etCO2), positive end-expiratory pressure (PEEP), mechanical respiratory rate, or ventilator mode. We identified need for pressors or inotropes (dobutamine, dopamine, droxidopa, midodrine, milrinone, ephpedrine, epinephrine, norepinephrine, phenylephrine, or vasopressin) during hospitalization based on detailed medication records data as binary variable. Acute kidney injury (AKI) was determined using available clinical information according to Kidney Disease: Improving Global Outcomes criteria (0.3 mg/dl increase in serum creatinine within 48 hours or 50% increase from baseline within seven days or decrease in urine output to less than 0.5 ml/kg/hr for six hours).[19] Community-acquired AKI was defined as development of AKI within 24 hours of hospital admission. Delirium was defined as at least one positive Confusion Assessment Method (CAM) score or having ICD-9 or ICD-10 codes for delirium. The International Classification of Diseases, Ninth and Tenth Revision, Clinical Modification (ICD-9-CM, ICD-10-CM) were used to the remaining complications[22-26]. Date of death was determined using hospital records and the Social Security Death Index database was used to confirm death dates and obtain death dates for subjects who were not in hospital records. Thirty-day and three-year mortality were defined if the death date is thirty days or three year from hospital admission.

## eFigure 1. Cohort selection and exclusion criteria

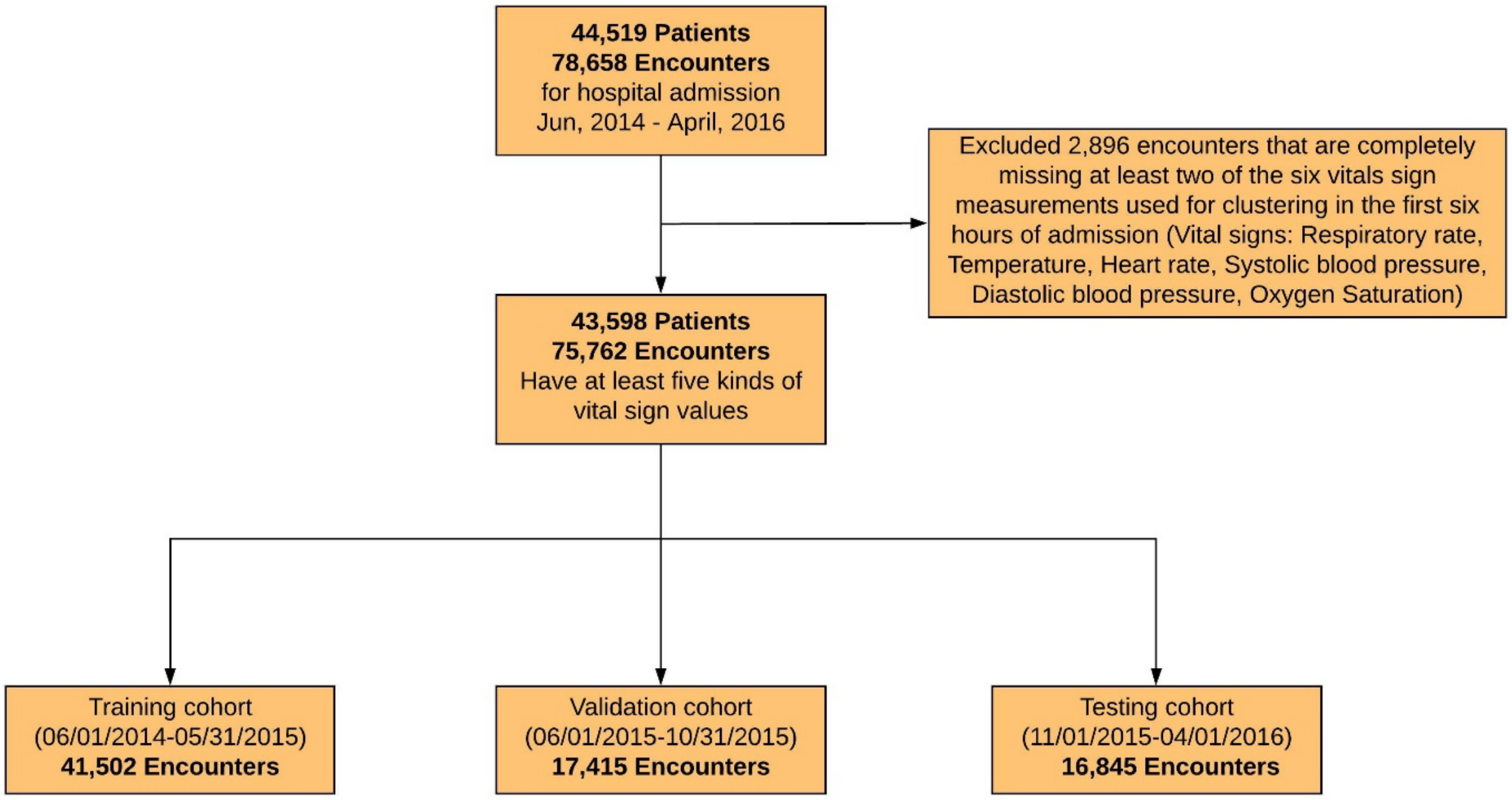

**eFigure 2. Spearman correlation heat map for the training cohort (N=41,502)**

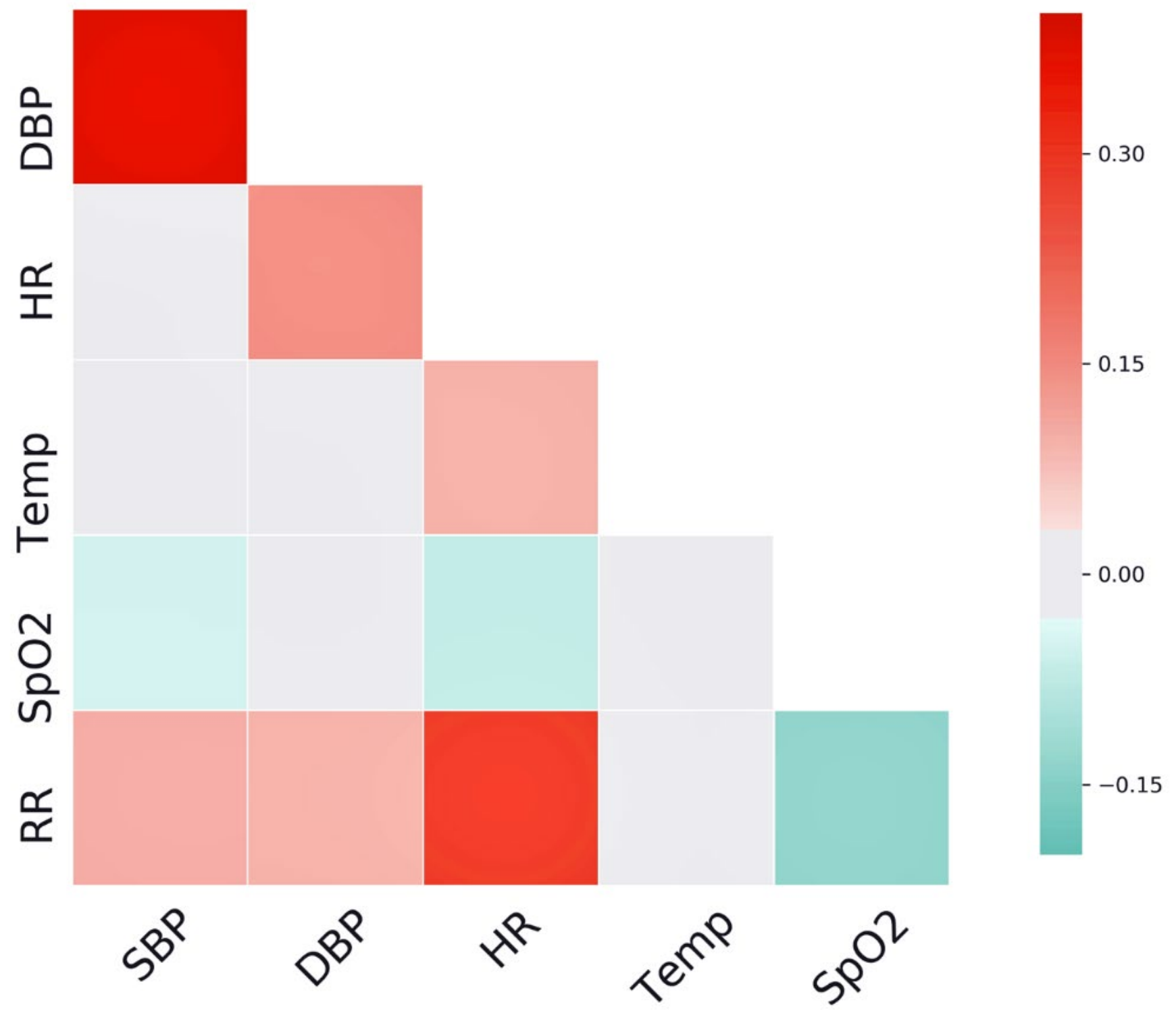

Spearman correlation heat map shows the pairwise spearman rank order correlation coefficient among the 6 vital signs studied in our paper. The darker red color, the higher correlation in positive direction.

Abbreviations: RR: respiratory rate; SpO2: peripheral capillary oxygen saturation; Temp: temperature; HR: heart rate; SBP: systolic blood pressure; DBP: diastolic blood pressure.

**eFigure 3. Deep temporal interpolation and clustering network algorithm**

---

**Algorithm 1:** Deep Temporal Interpolation and Clustering Network

---

**Input:** Input data: $X$; Target distribution update interval: $T$; Stopping threshold: $\delta$; Maximum iterations: MaxIter

**Output:** Cluster centers $\mu$ and labels $s$

Initialize the weights $W_1$ of Interpolation Net, Bi-LSTM Encoder and Decoder, Re-interpolation Net and Linear Classifier (MLPs)

**for** $iter \in \{0, 1, \ldots, MaxIter\}$ **do**

  Randomly choose a batch of samples $B \in X$

  Generate an associated batch of fake sample $\bar{B}$

  Compute the training loss with $\mathcal{L} = \mathcal{L}_{\text{reconstruction}} + \mathcal{L}_{\text{auxiliary}}$

  Update $W_1$ and deep embedding representation $h$ using Adam

**end**

Incorporate the Cluster Net into the full model, and initialize its weight $W_2$

Initialize the cluster centers $\mu$ by applying K-means on $h$

Initialize the label assignment $s$

**for** $iter \in \{0, 1, \ldots, MaxIter\}$ **do**

  Randomly choose a batch of samples $B \in X$

  Generate an associated batch of fake sample $\bar{B}$

  Compute the training loss with $\mathcal{L} = \mathcal{L}_{\text{reconstruction}} + \mathcal{L}_{\text{auxiliary}} + \lambda\mathcal{L}_{\text{clustering}}$

  Update $W_1$, $W_2$, $h$ and $\mu$ using Adam

  **if** $iter \,\%\, T == 0$ **then**

    Compute new label assignments $s_{new}$ for all $n$ samples

    **if** $sum(s \neq s_{new})/n < \delta$ **then**

      Stop training

    **end**

    Update $s$ with $s_{new}$

  **end**

**end**

Apply K-means to the deep embedded representations $h$ and generate the final cluster result $s$.

---

**eFigure 4. Gap statistic and elbow approaches showing the optimal number of clusters**

(**A**).

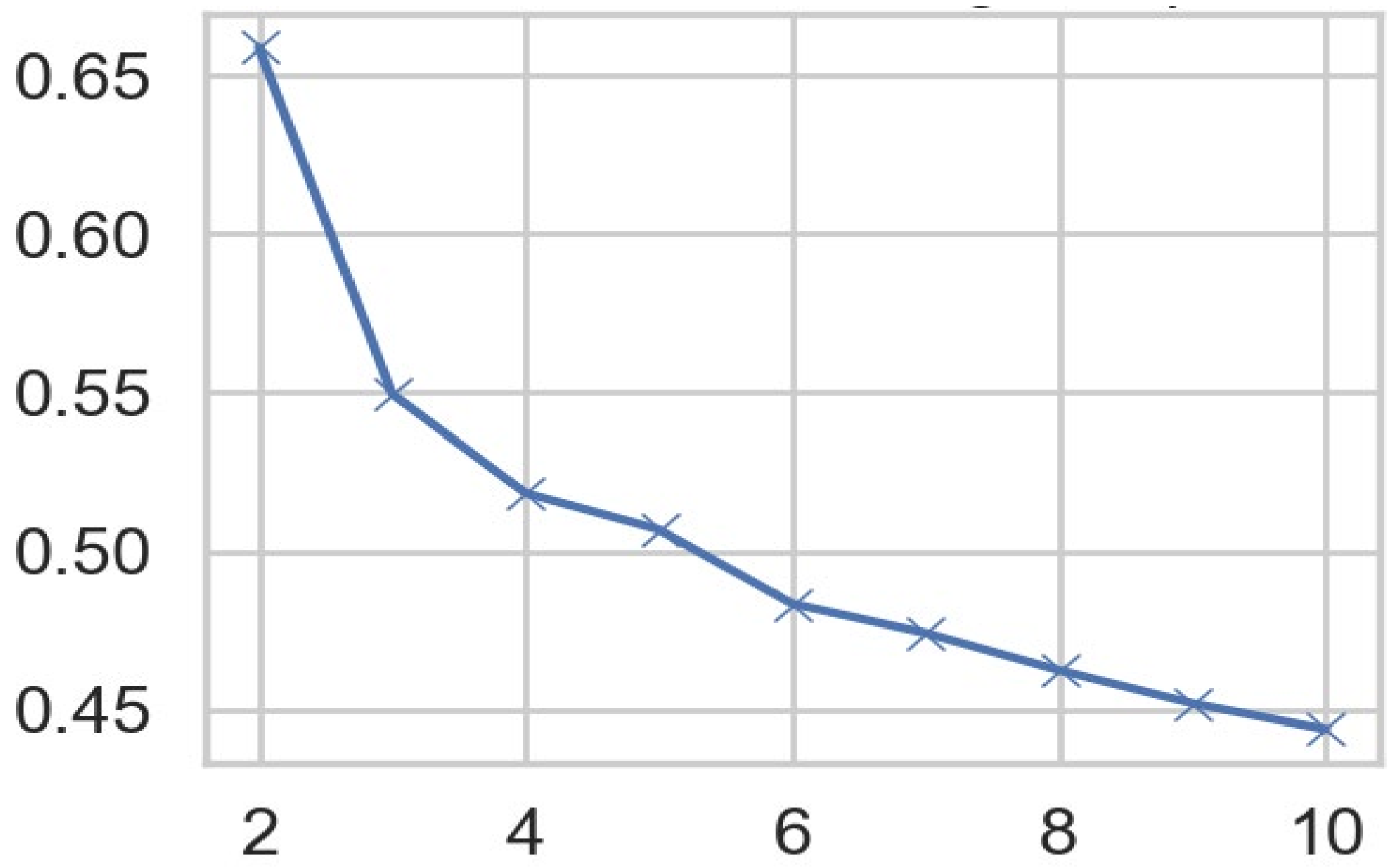

(**B**).

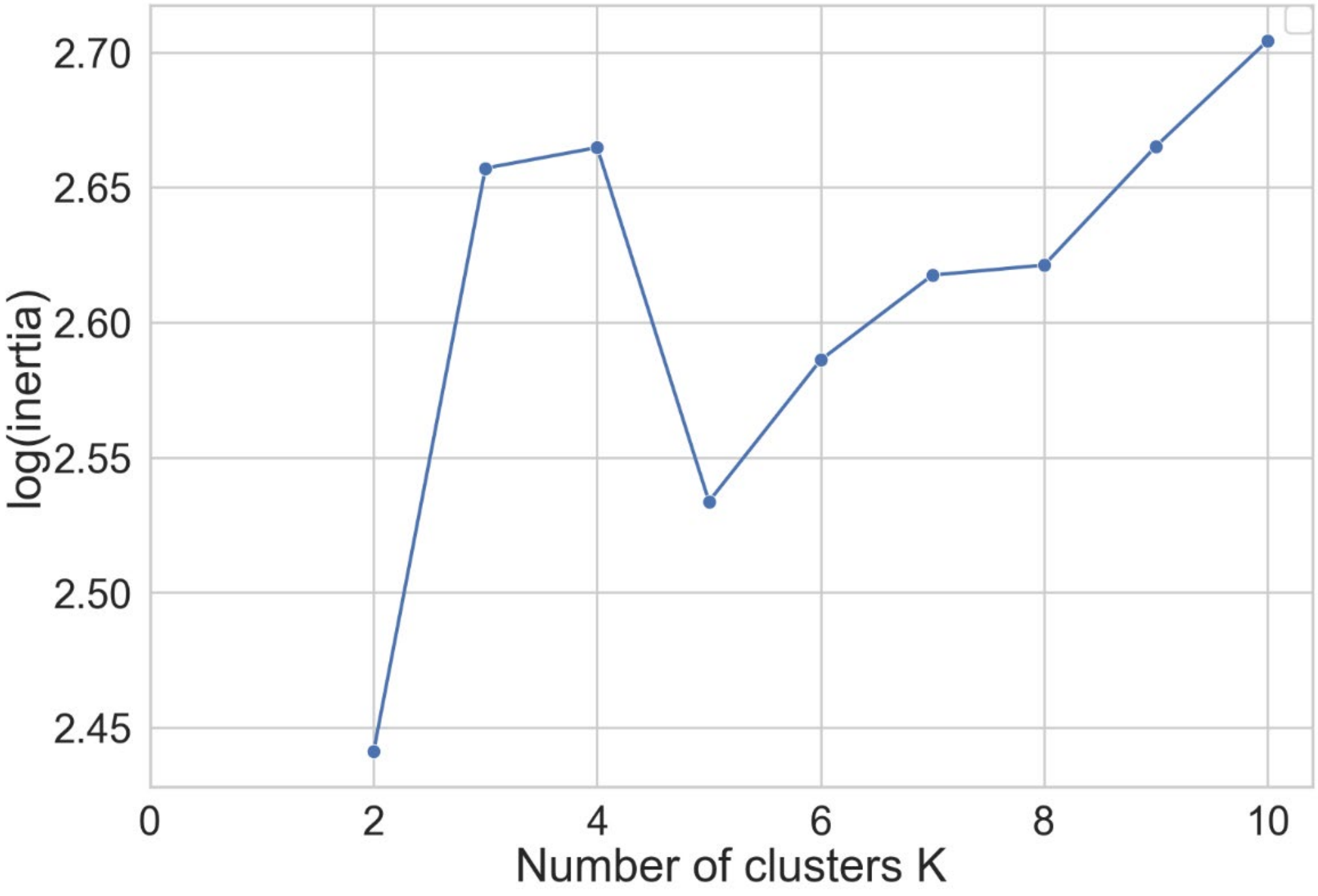

(A) Elbow approach shows the optimal number of clusters. The value of k at the "elbow", the point after which the distortion starts decreasing in a linear fashion, suggests the optimal number of clusters. (B) Gap statistic approach shows the optimal number of clusters. Higher gap statistic value suggests the optimal number of clusters.

**eFigure 5. Alluvial plot showing distribution of phenotypes across worst SOFA scores of patients within first 24 hours of admission in the training cohort**

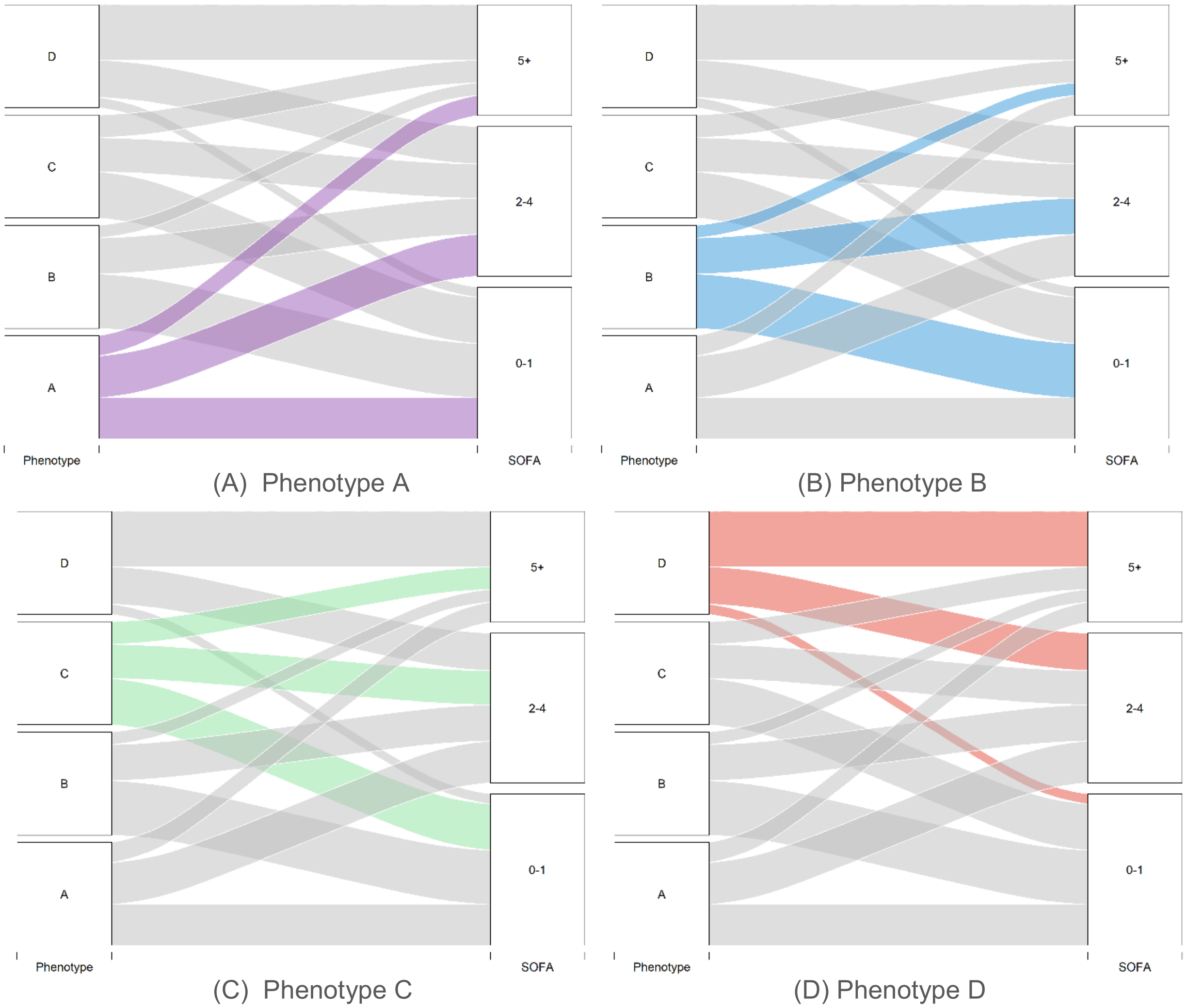

For each phenotype, the larger percentage of patients with that score, the broader the ribbon.

**eFigure 6. Chord diagrams showing the distribution of patients with higher SOFA scores (i.e., 2+) within first 24 hours of admission of six organ systems by phenotypes in the training cohort**

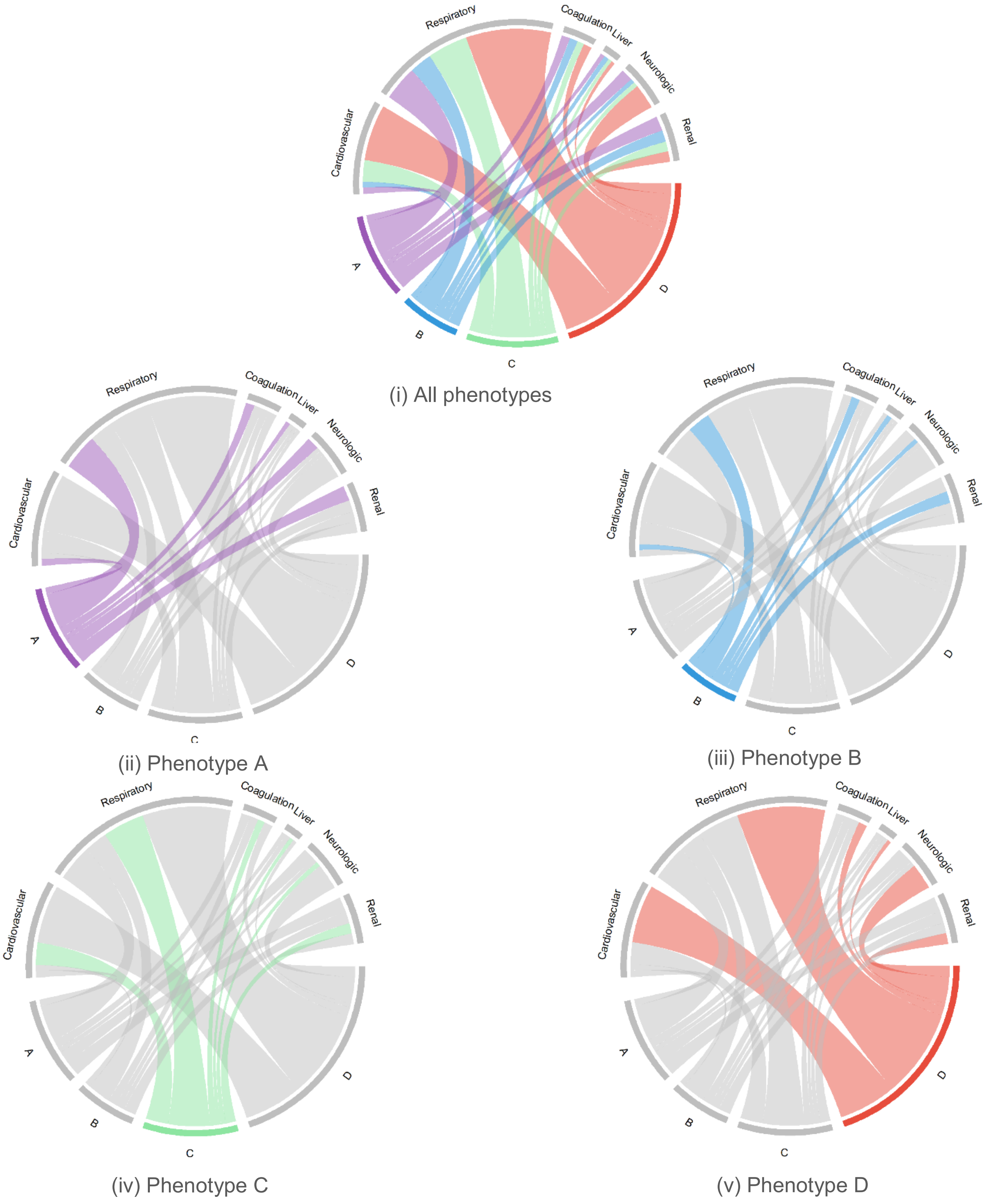

For each phenotype, the larger percentage of patients with higher score of that organ system, the border the ribbon.

**eFigure 7. Survival curves and Cox proportional hazards modeling by phenotypes in the training cohort**

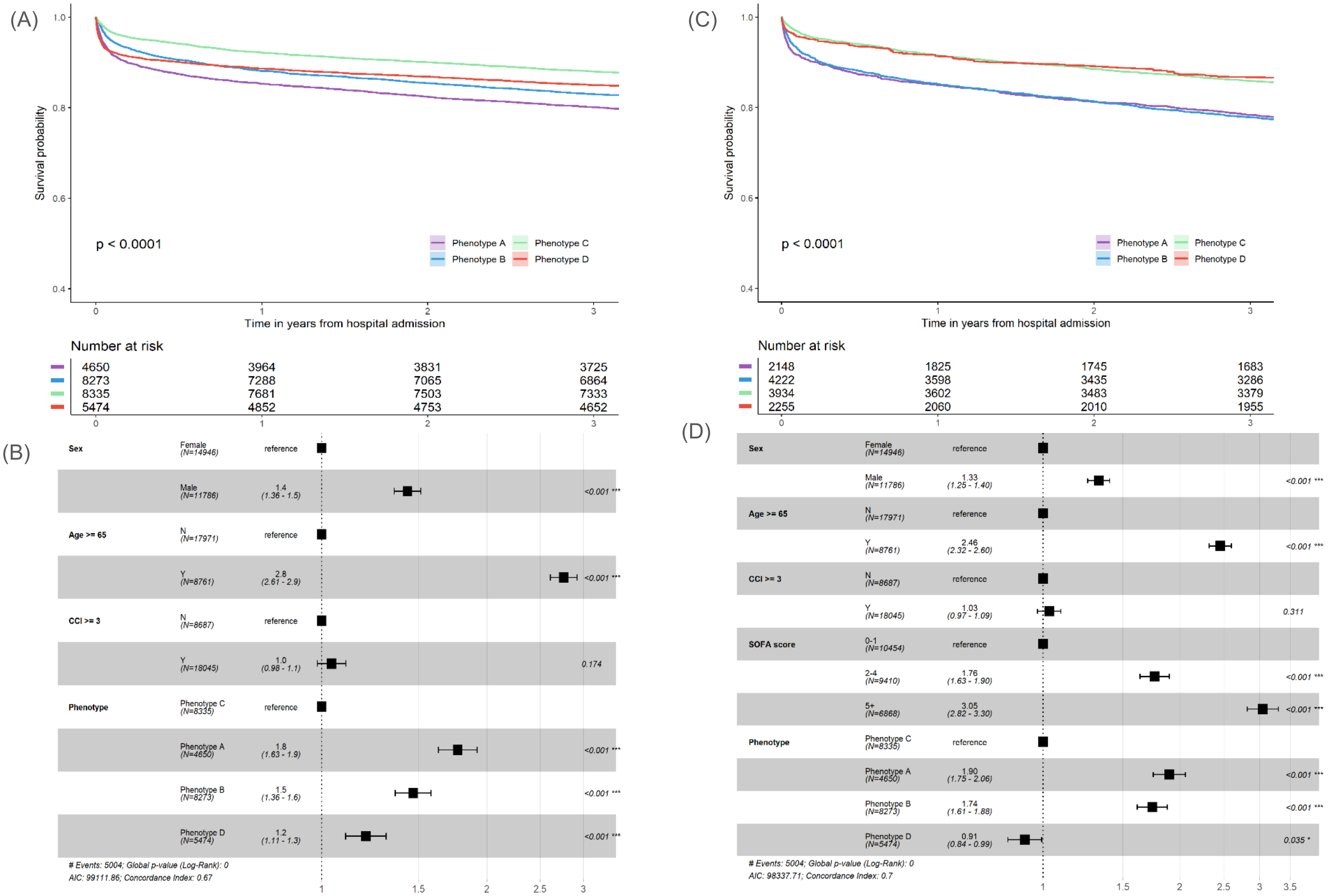

(A) Phenotype survival curves adjusted using demographic information and comorbidities. (B) Adjusted Cox proportional hazards models using demographic information and comorbidities. (C) Phenotype survival curves adjusted using demographic information, comorbidities, and SOFA scores. (D) Adjusted Cox proportional hazards model using demographic information, comorbidities, and SOFA scores. Abbreviation: CCI: charlson comorbidity index; SOFA: sequential organ failure assessment.

**eFigure 8. Survival curves and Cox proportional hazards modeling by sepsis patients in the training cohort**

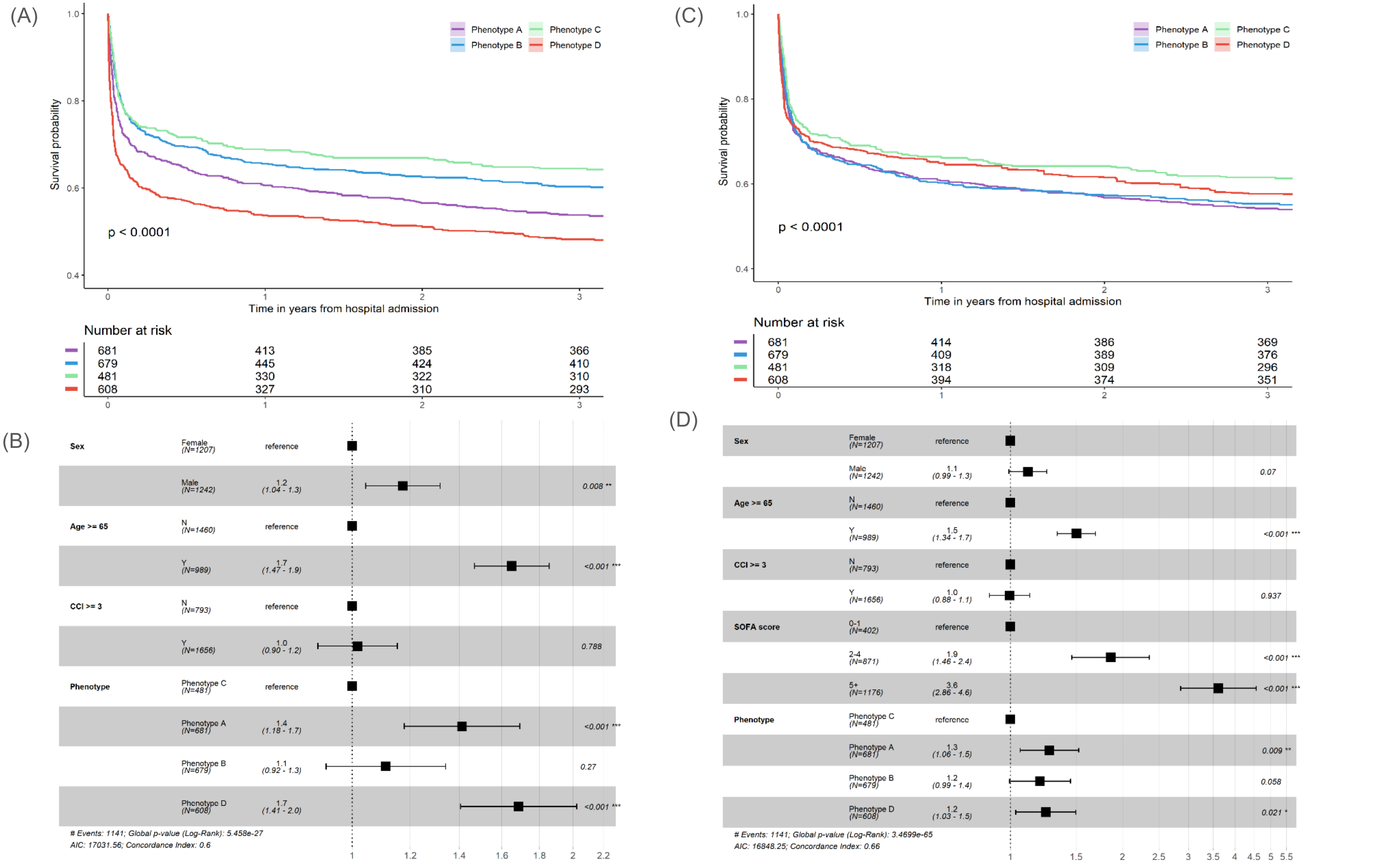

(A) Phenotype survival curves adjusted using demographic information and comorbidities. (B) Adjusted Cox proportional hazards models using demographic information and comorbidities. (C) Phenotype survival curves adjusted using demographic information, comorbidities, and SOFA scores. (D) Adjusted Cox proportional hazards model using demographic information, comorbidities, and SOFA scores. Abbreviation: CCI: charlson comorbidity index; SOFA: sequential organ failure assessment.

## eFigure 9. Survival curves and Cox proportional hazards modeling by AKI patients in training cohort

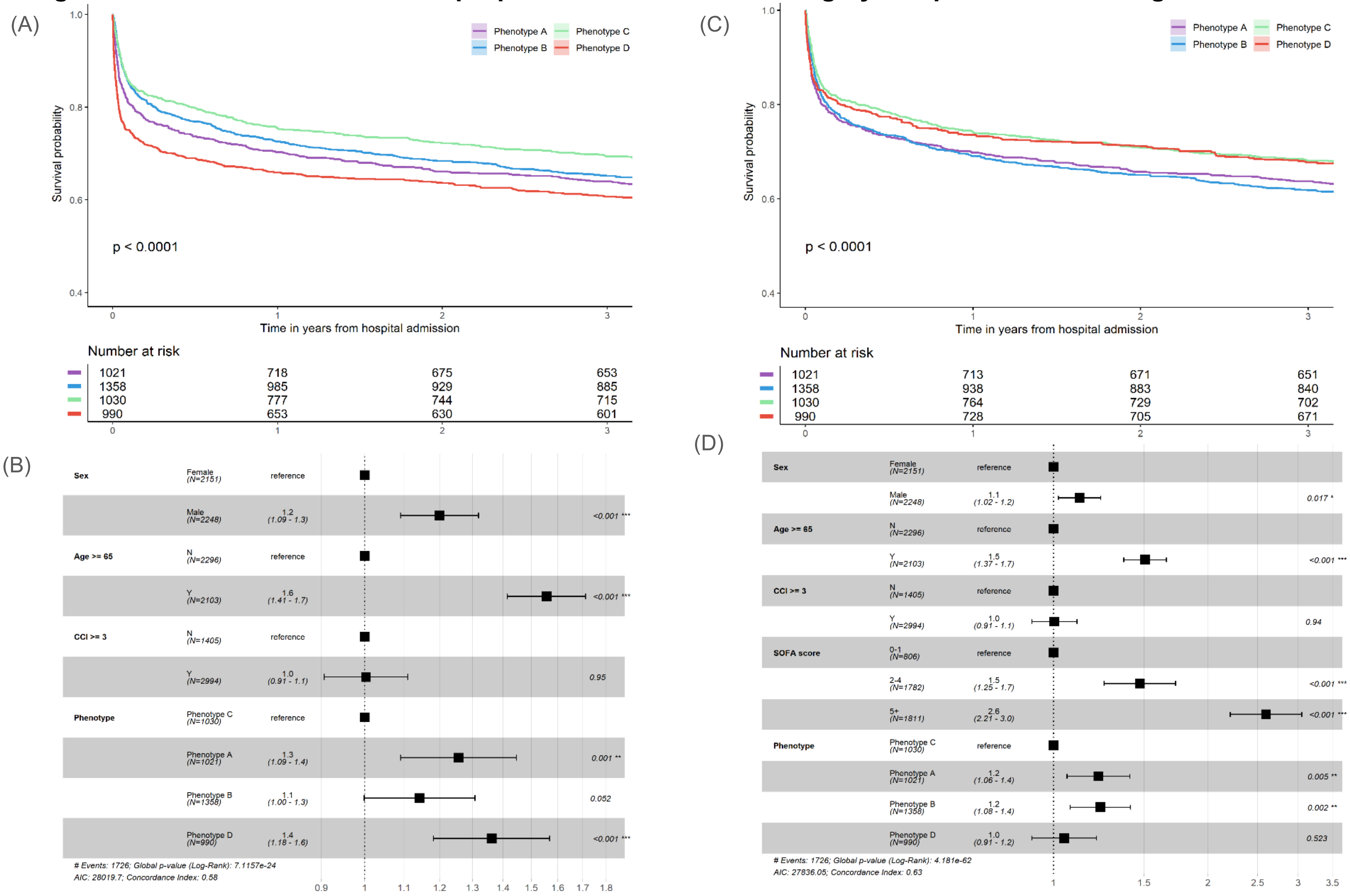

(A) Phenotype survival curves adjusted using demographic information and comorbidities. (B) Adjusted Cox proportional hazards models using demographic information and comorbidities. (C) Phenotype survival curves adjusted using demographic information, comorbidities, and SOFA scores. (D) Adjusted Cox proportional hazards model using demographic information, comorbidities, and SOFA scores. Abbreviation: CCI: charlson comorbidity index; SOFA: sequential organ failure assessment.

# eFigure 10. Survival curves and Cox proportional hazards modeling by surgical patients in training cohort

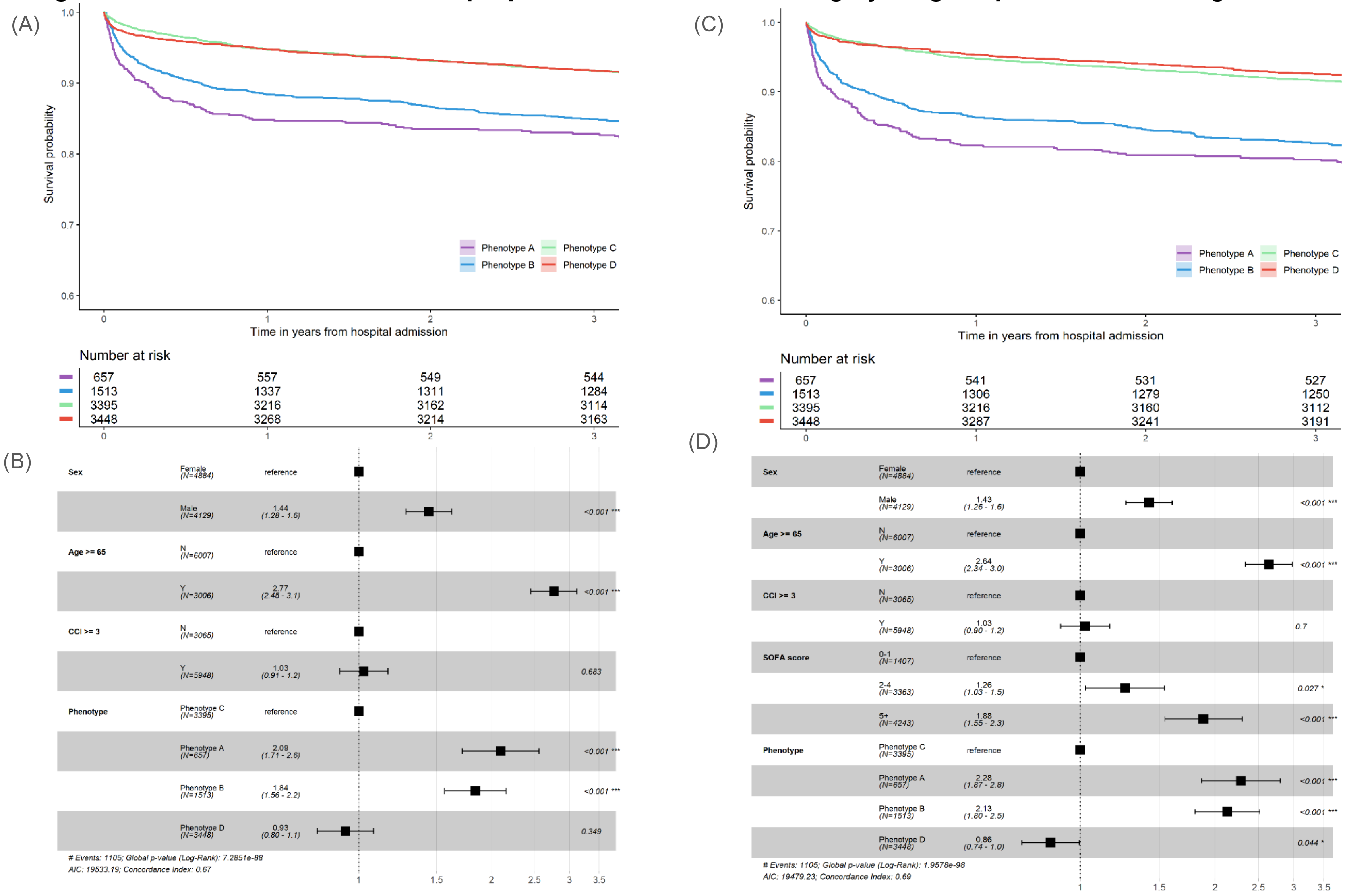

(A) Phenotype survival curves adjusted using demographic information and comorbidities. (B) Adjusted Cox proportional hazards models using demographic information and comorbidities. (C) Phenotype survival curves adjusted using demographic information, comorbidities, and SOFA scores. (D) Adjusted Cox proportional hazards model using demographic information, comorbidities, and SOFA scores. Abbreviation: CCI: charlson comorbidity index; SOFA: sequential organ failure assessment.

**eFigure 11. Distribution of vital signs during the first six hours of hospital admission in the testing cohort**

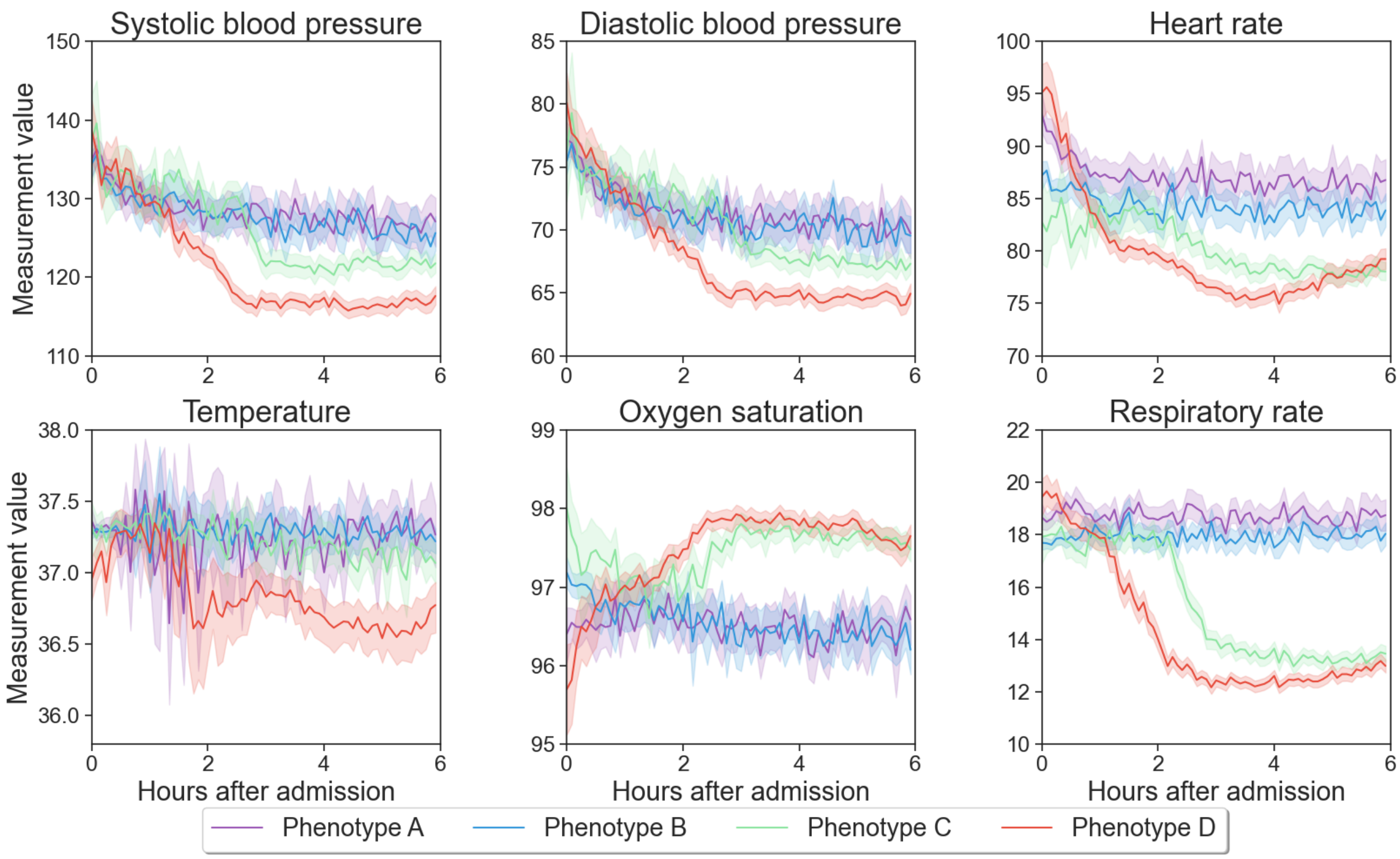

## eFigure 12. t-SNE plot of phenotype assignments in the testing cohort

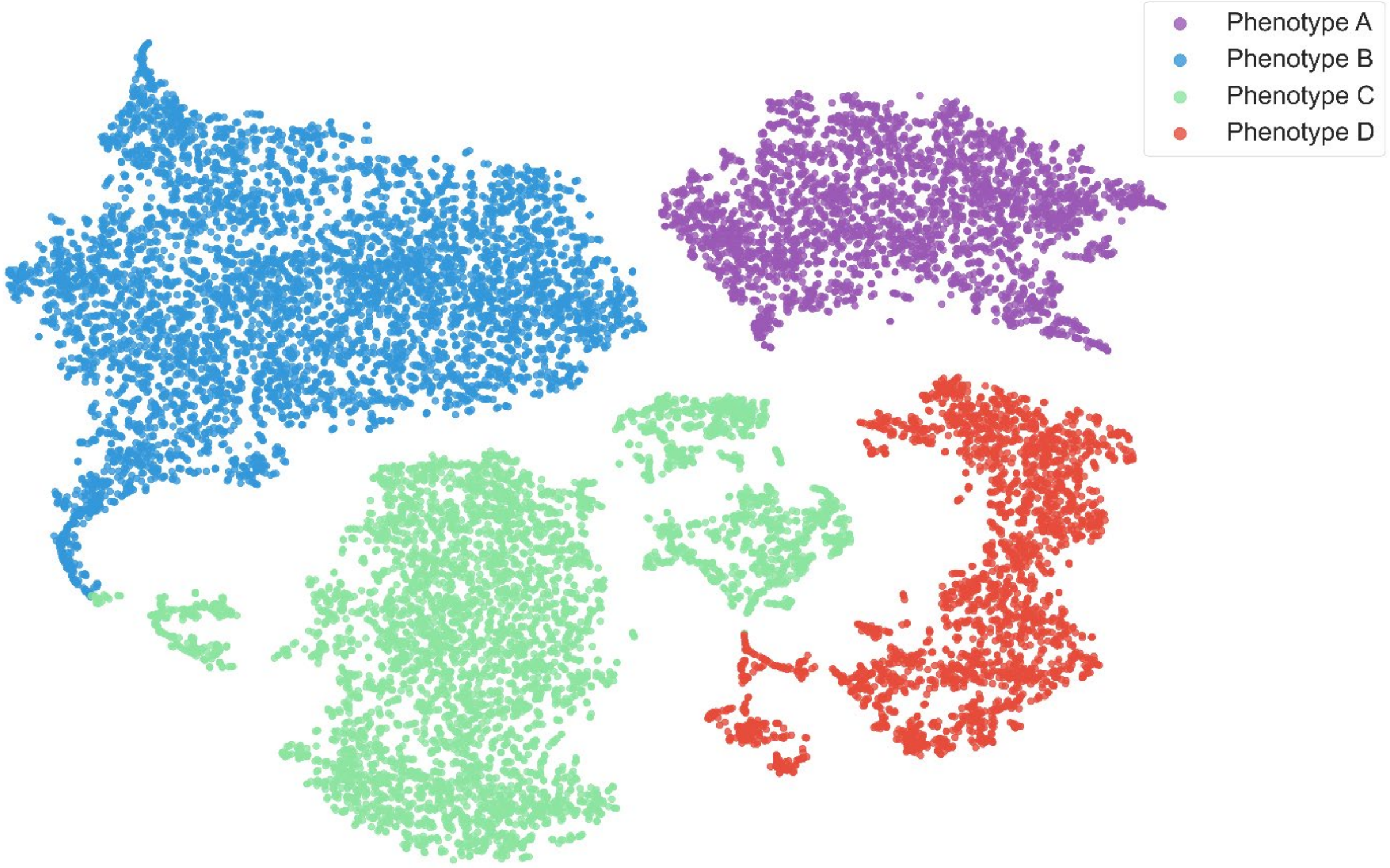

Starting from the original 128 dimensional vital sign representations, we run the t-SNE to reduce to 2 dimensions. Each dot represents a patient. Phenotypes are show in separate colors.

**eFigure 13. Alluvial plot showing distribution of phenotypes across worst SOFA scores of patients within first 24 hours of admission in the testing cohort**

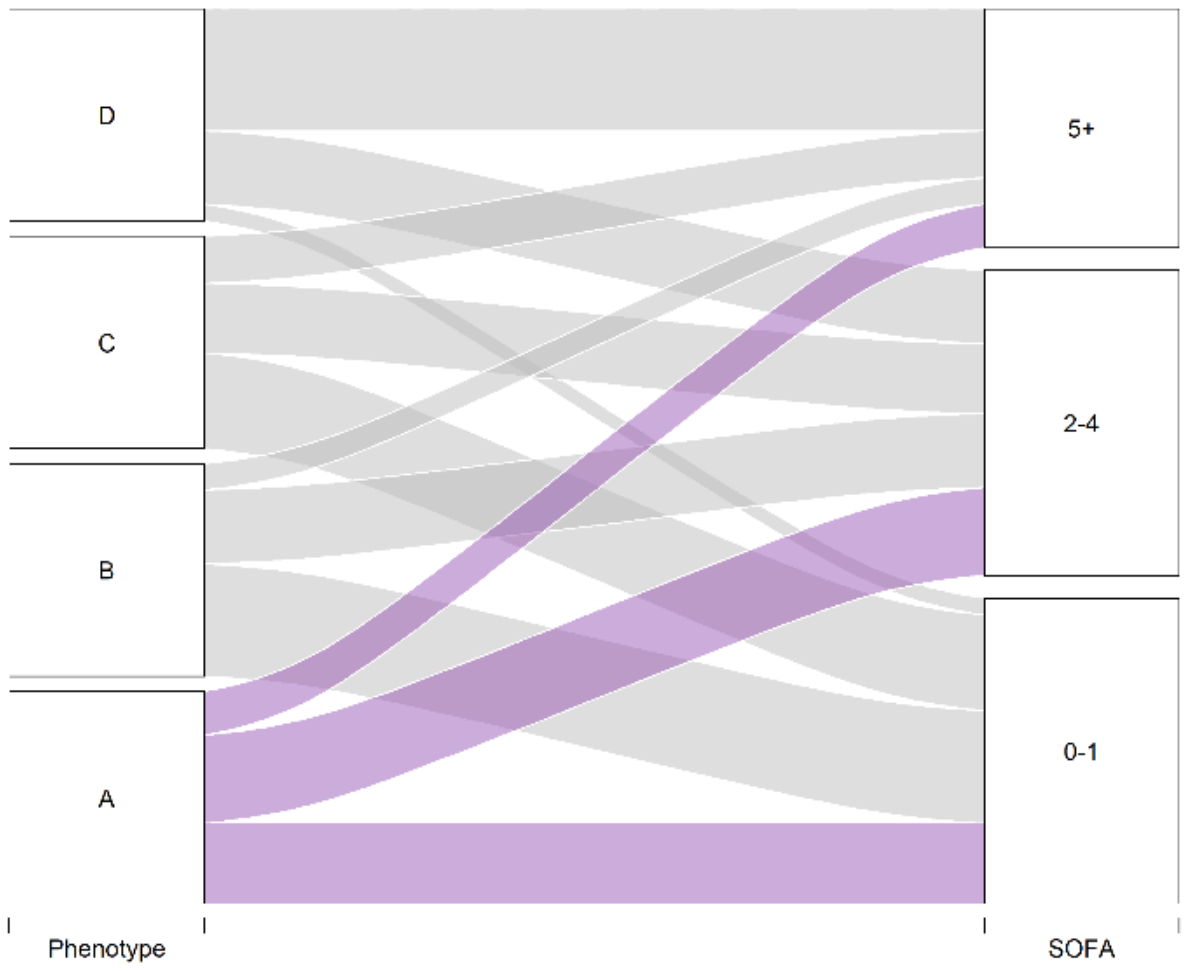

(A) Phenotype A

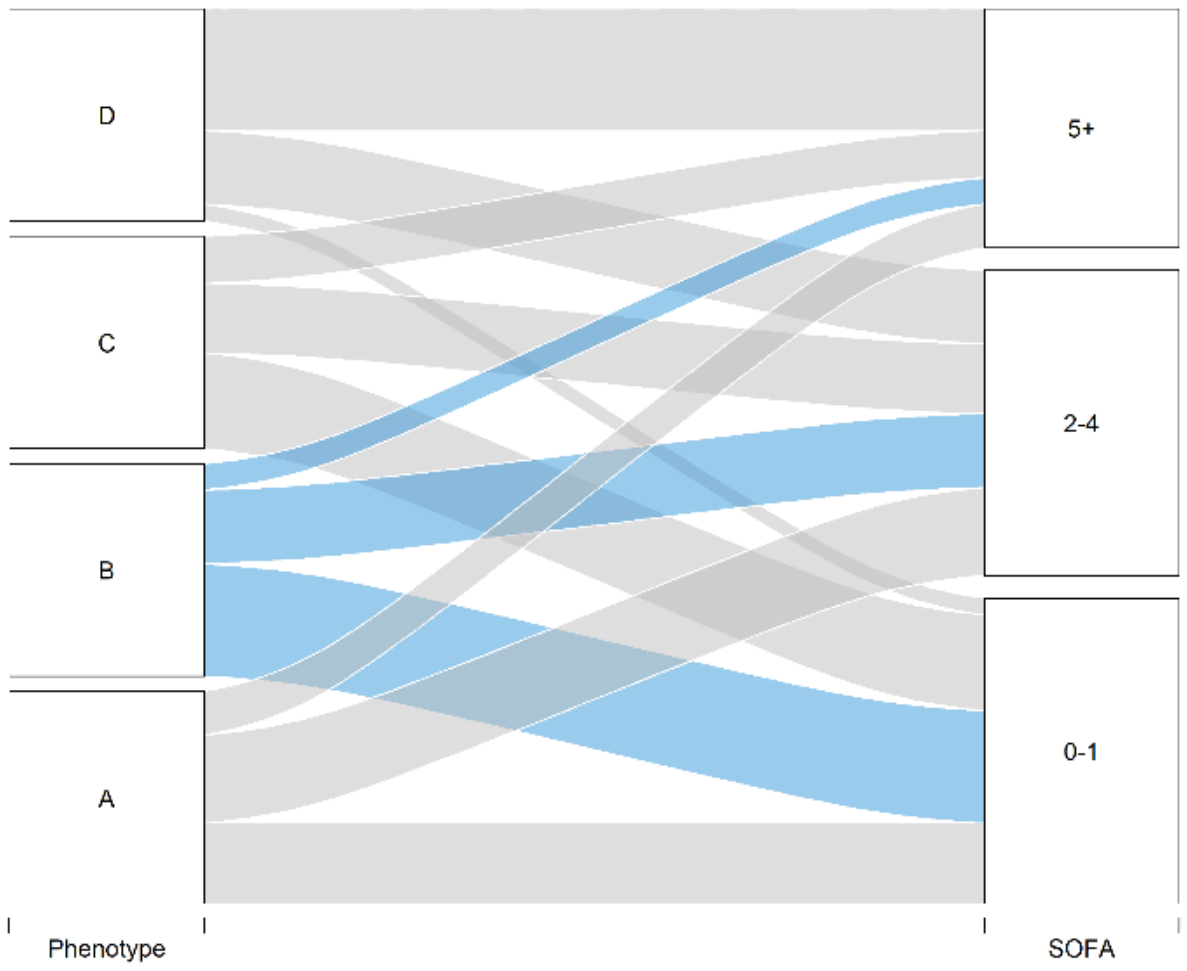

(B) Phenotype B

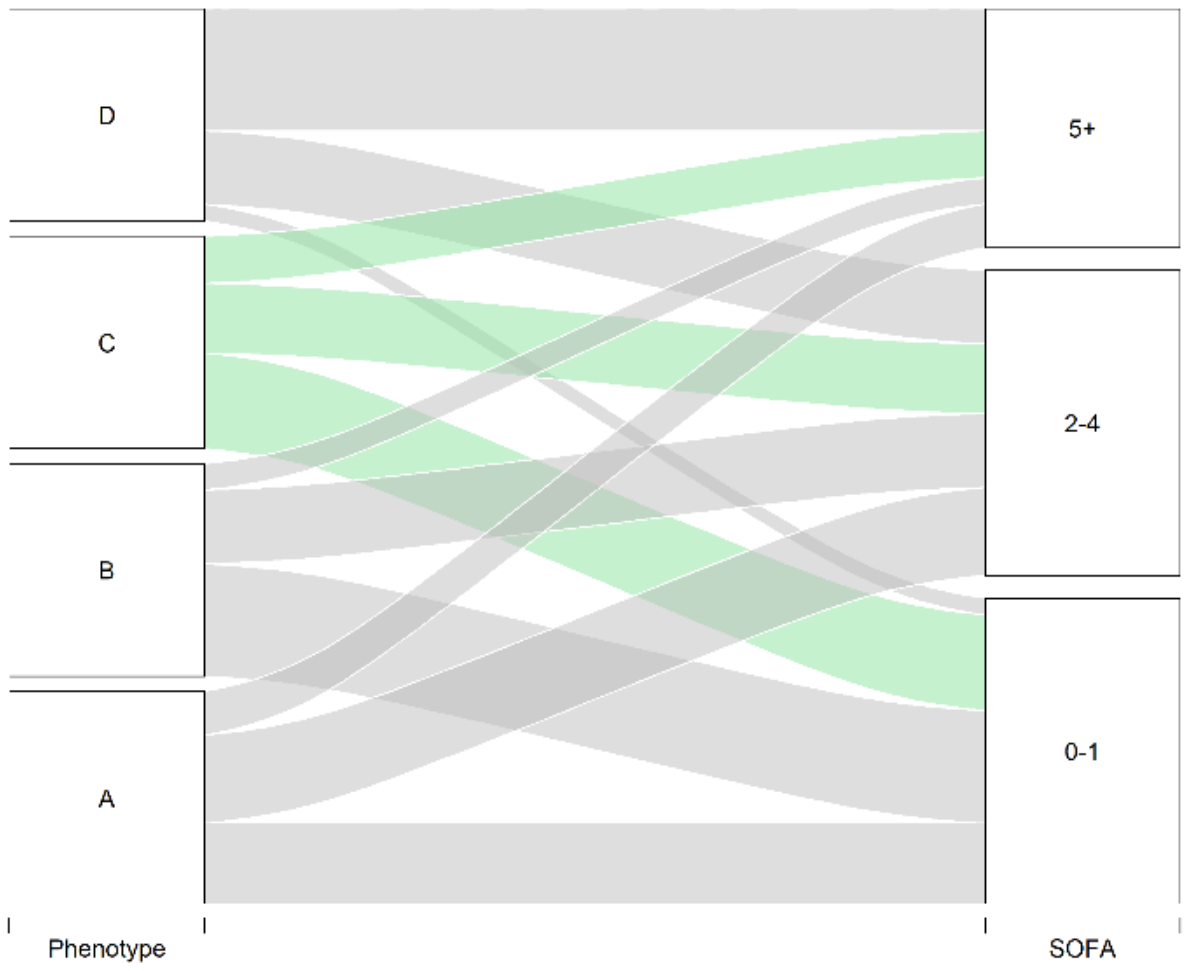

(C) Phenotype C

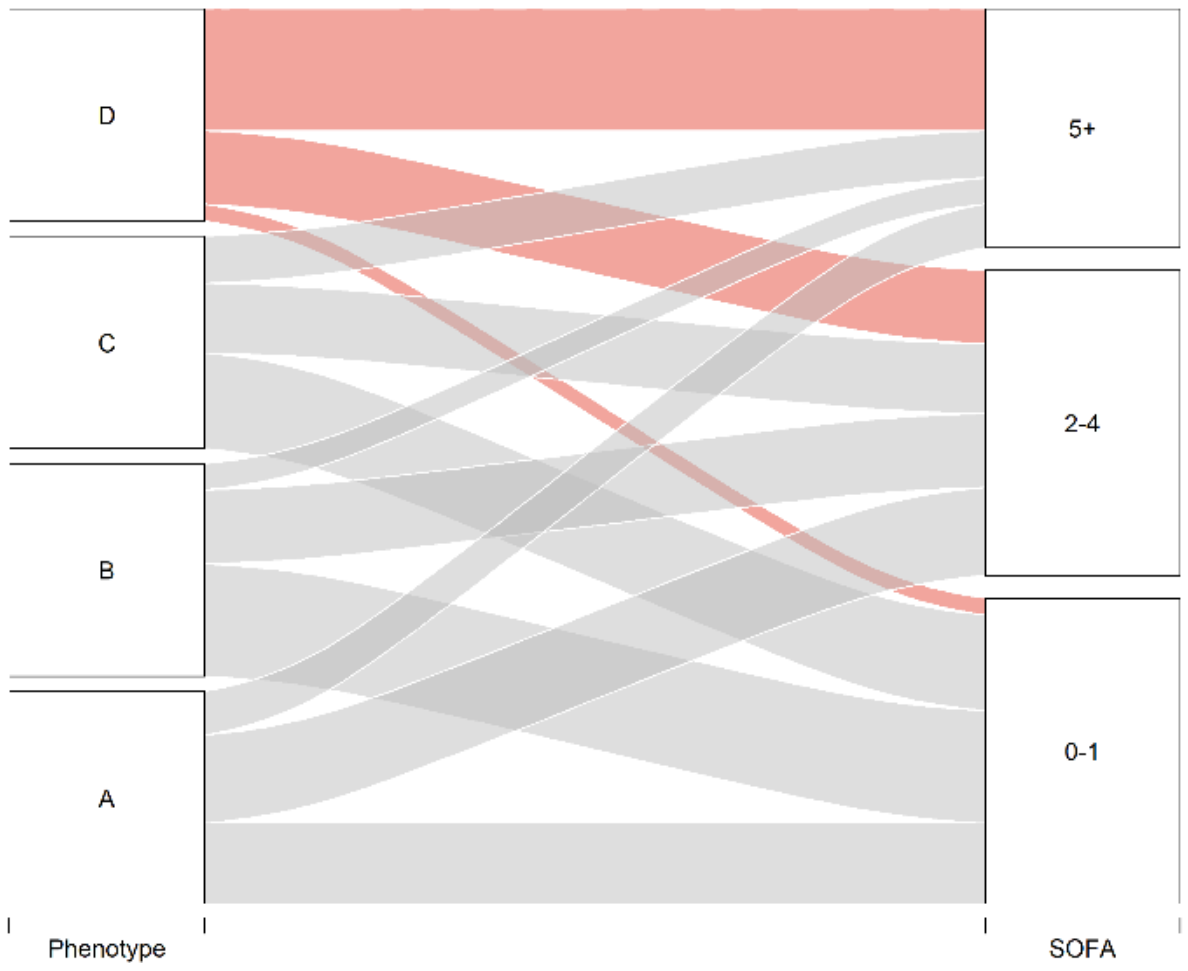

(D) Phenotype D

For each phenotype, the larger percentage of patients with that score, the broader the ribbon.

## eFigure 14. Chord diagrams showing the distribution of patients with higher SOFA scores (i.e., 2+) within first 24 hours of admission of six organ systems by phenotypes in the testing cohort

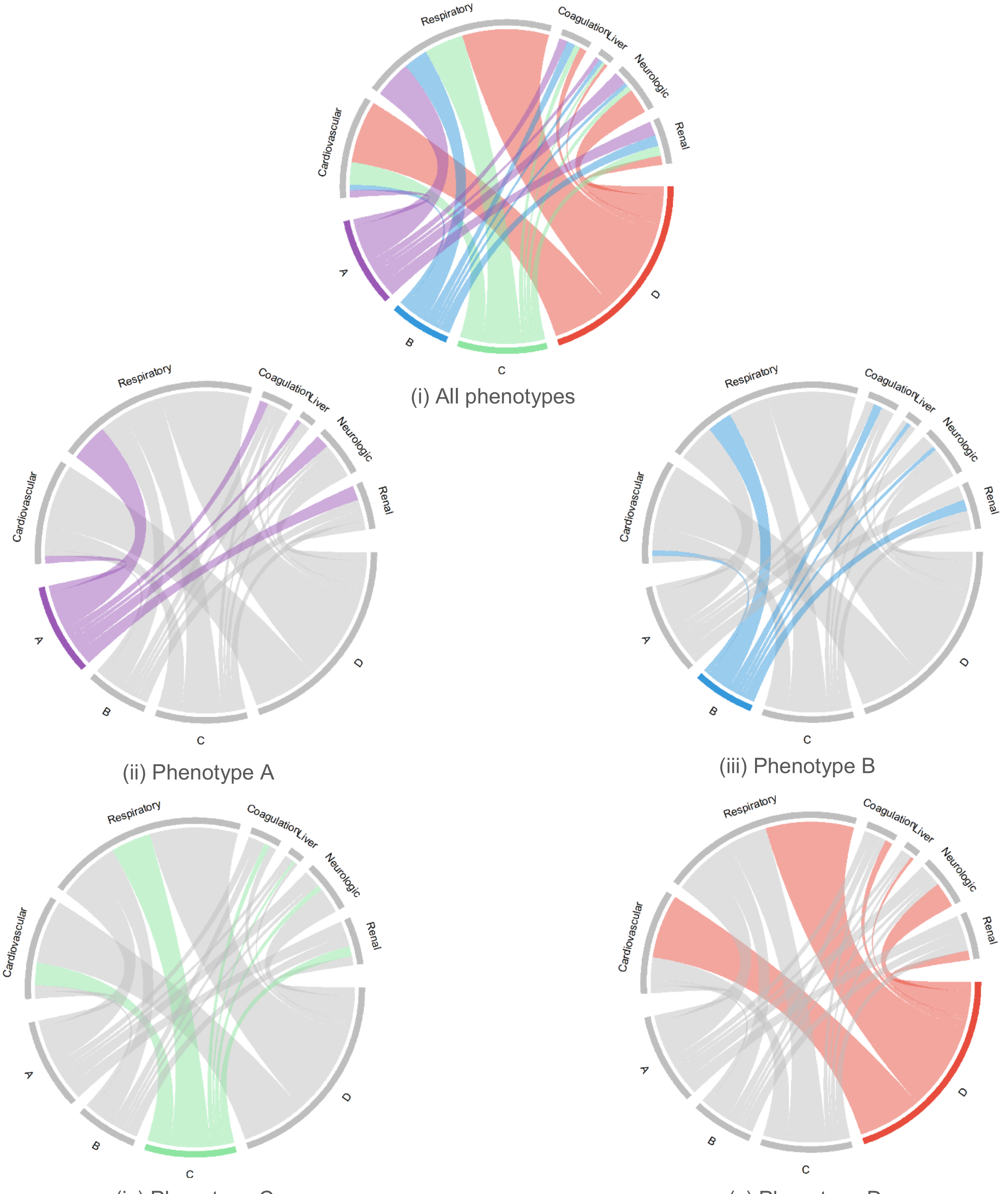

(i) All phenotypes

(ii) Phenotype A

(iii) Phenotype B

(iv) Phenotype C

(v) Phenotype D

For each phenotype, the larger percentage of patients with higher score of that organ system, the border the ribbon.

## eFigure 15. Survival curves and Cox proportional hazards modeling by phenotypes in the testing cohort

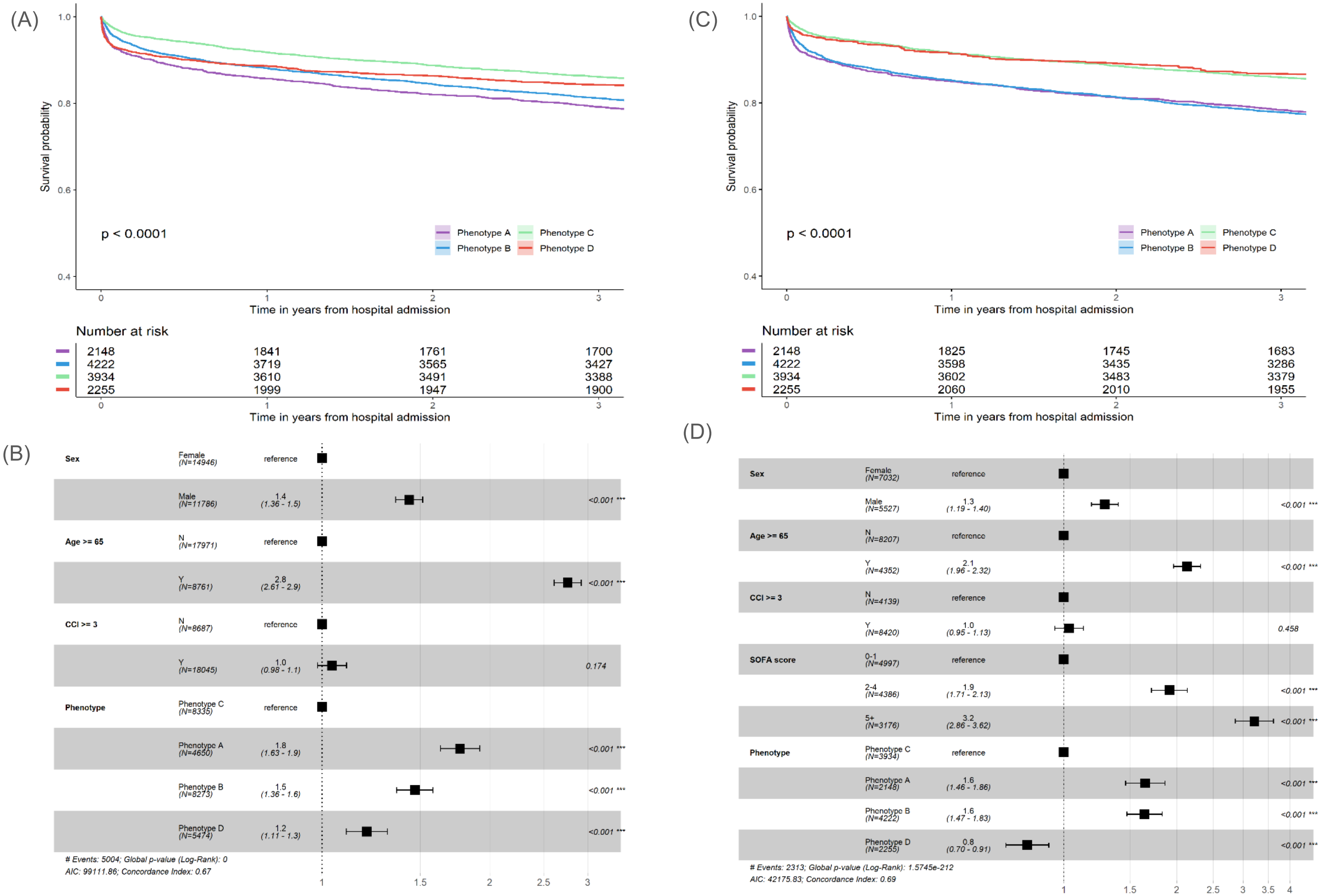

(A) Phenotype survival curves adjusted using demographic information and comorbidities. (B) Adjusted Cox proportional hazards models using demographic information and comorbidities. (C) Phenotype survival curves adjusted using demographic information, comorbidities, and SOFA scores. (D) Adjusted Cox proportional hazards model using demographic information, comorbidities, and SOFA scores. Abbreviation: CCI: charlson comorbidity index; SOFA: sequential organ failure assessment.

**eFigure 16. Chord diagrams showing the distribution of nine most common admission diagnosis groups by phenotype in the training cohort**

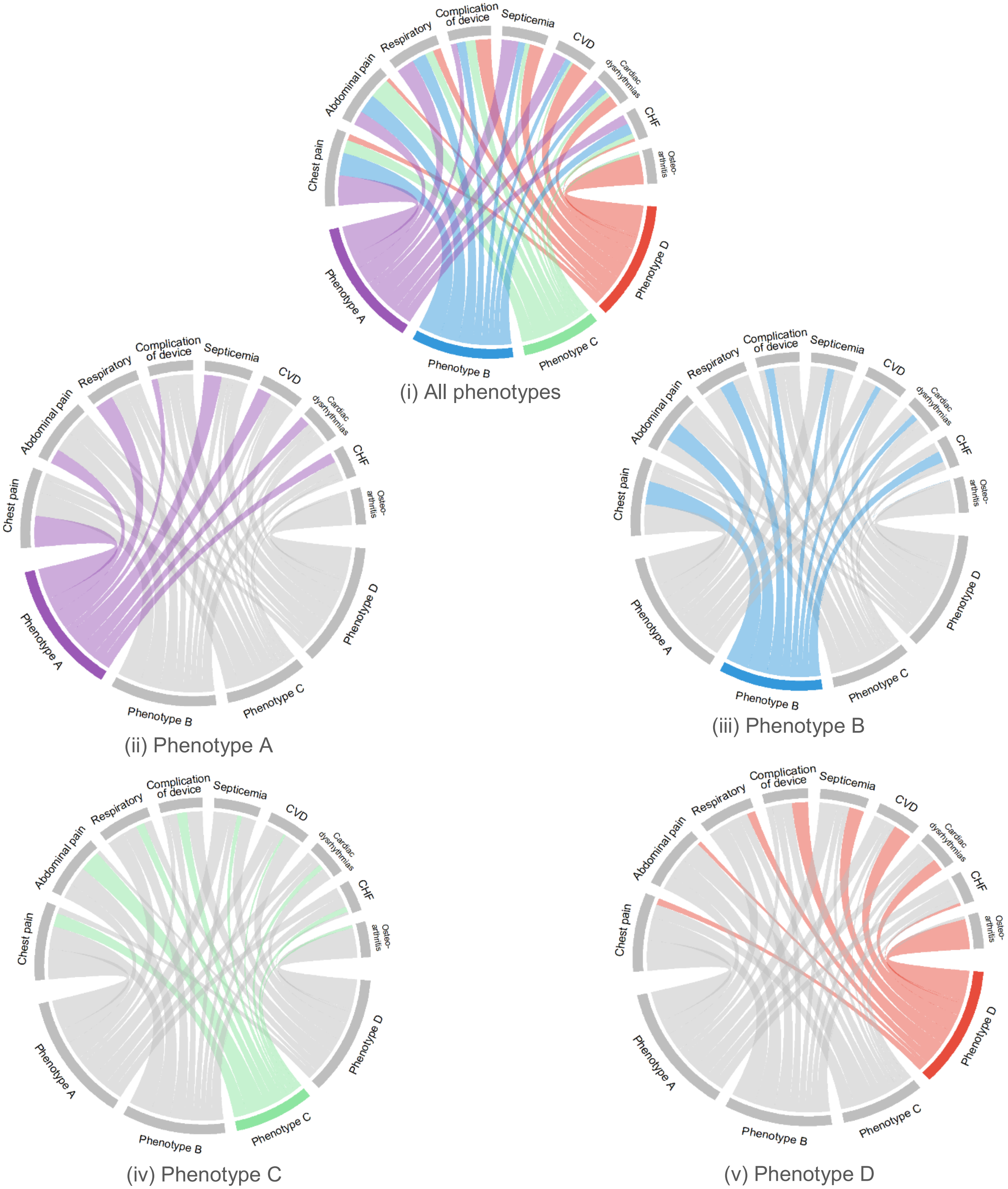

Diagnosis groups are shown in order of frequencies of all patients. For each phenotype, the larger percentage of patients with that diagnosis, the border the ribbon. Detailed diagnosis groups from left to right are: Nonspecific chest pain, Abdominal pain, Other and unspecific lower respiratory disease, Complication of device; implant or graft, Speticemia (except in labor), Acute cerebrovascular disease, Cardiac dysrhythmias, Congestive heart failure; nonhypertensive, and Osteoarthritis.

## eFigure 17. Chord diagrams showing the distribution of nine most common admission diagnosis groups by phenotype in the training cohort

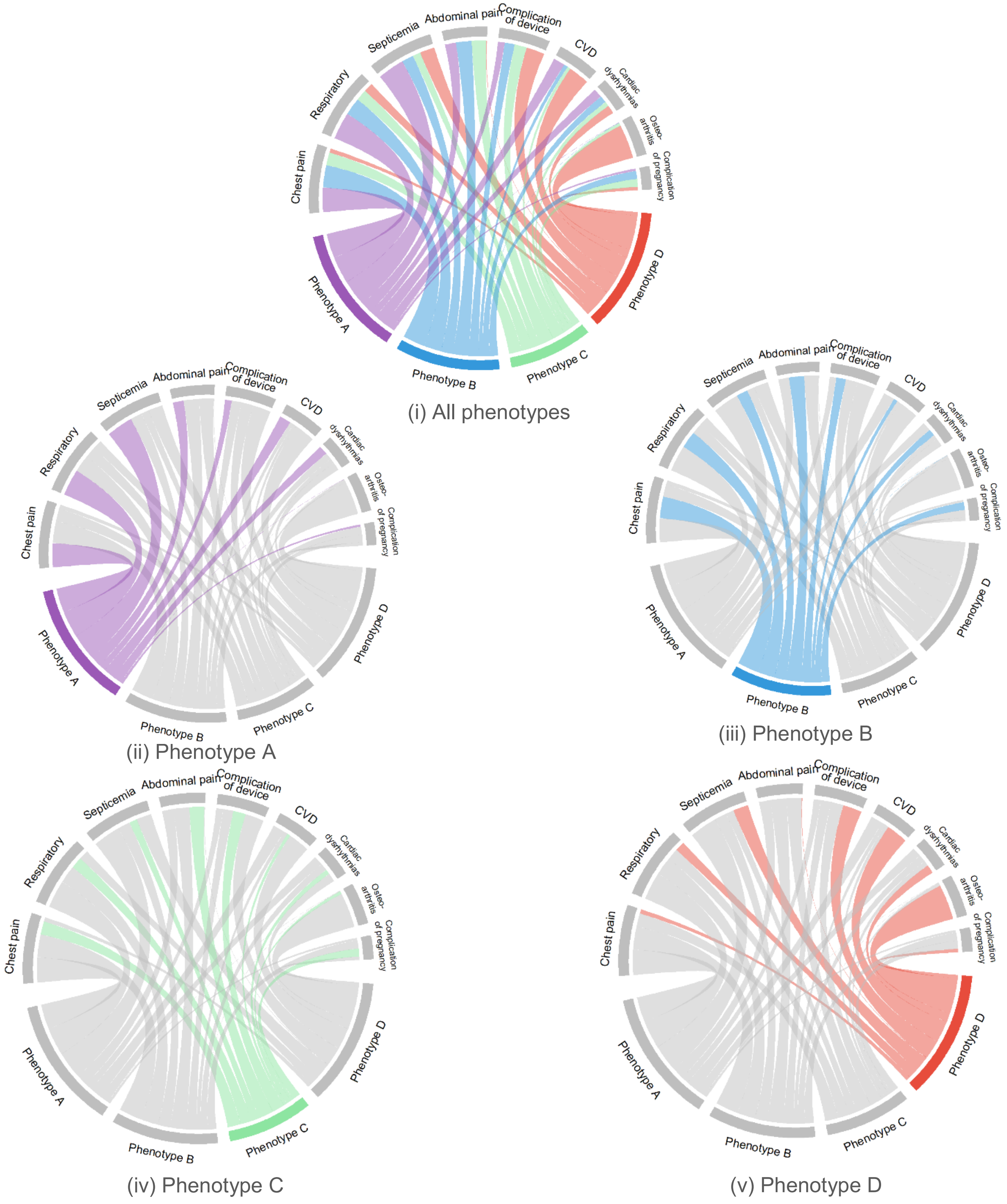

Diagnosis groups are shown in order of frequencies of all patients. For each phenotype, the larger percentage of patients with that diagnosis, the border the ribbon. Detailed diagnosis groups from left to right are: Nonspecific chest pain, Other and unspecific lower respiratory disease, Speticemia (except in labor), Abdominal pain, Complication of device; implant or graft, Acute cerebrovascular disease, Cardiac dysrhythmias, Osteoarthritis, and Other complications of pregnancy.

**eFigure 18. Reconstruction error of physiologic signatures measured within six hours of hospital admission in training cohort using deep temporal interpolation and clustering network**

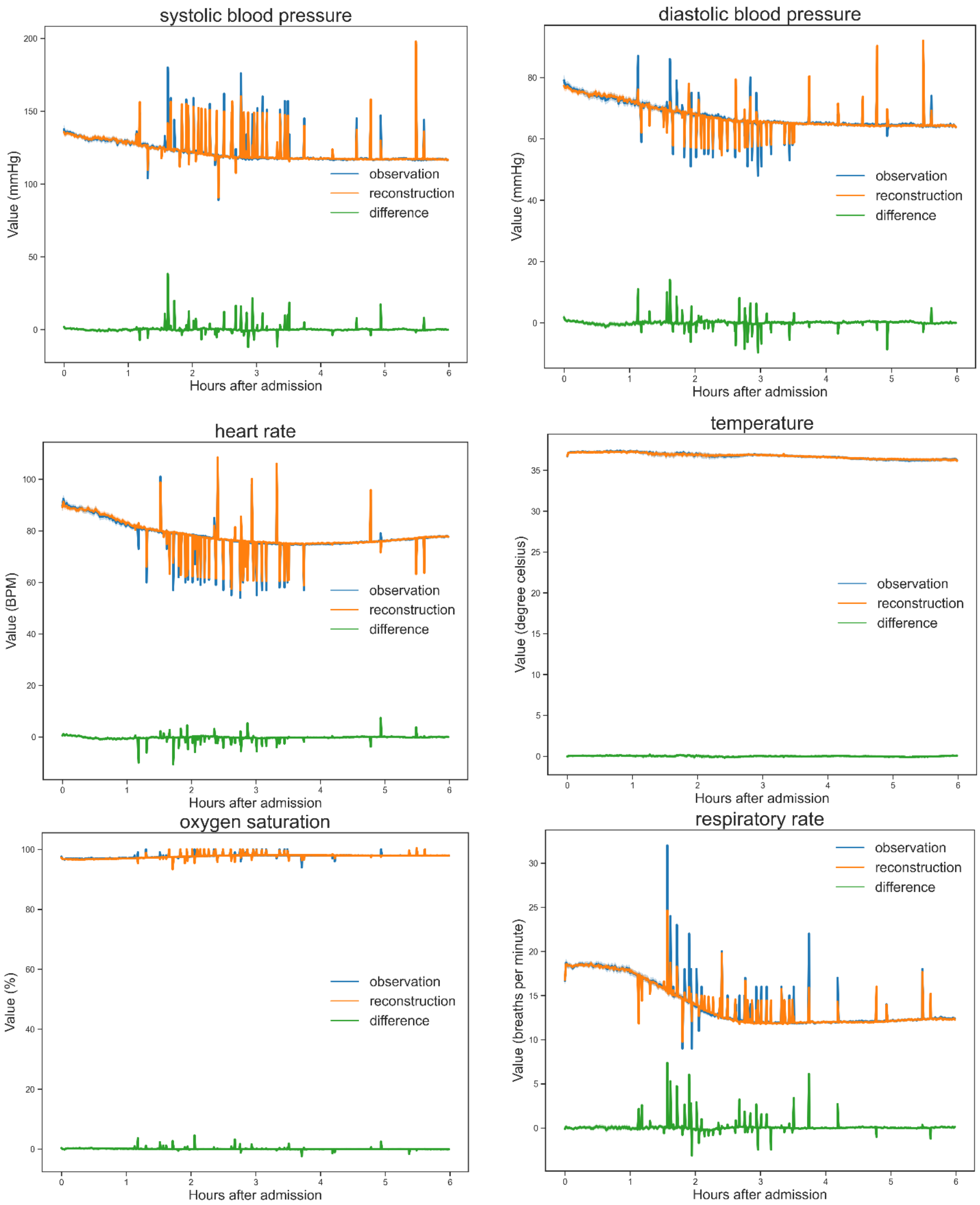

**eFigure 19. Reconstruction error of physiologic signatures measured within six hours of hospital admission in testing cohort using deep temporal interpolation and clustering network**

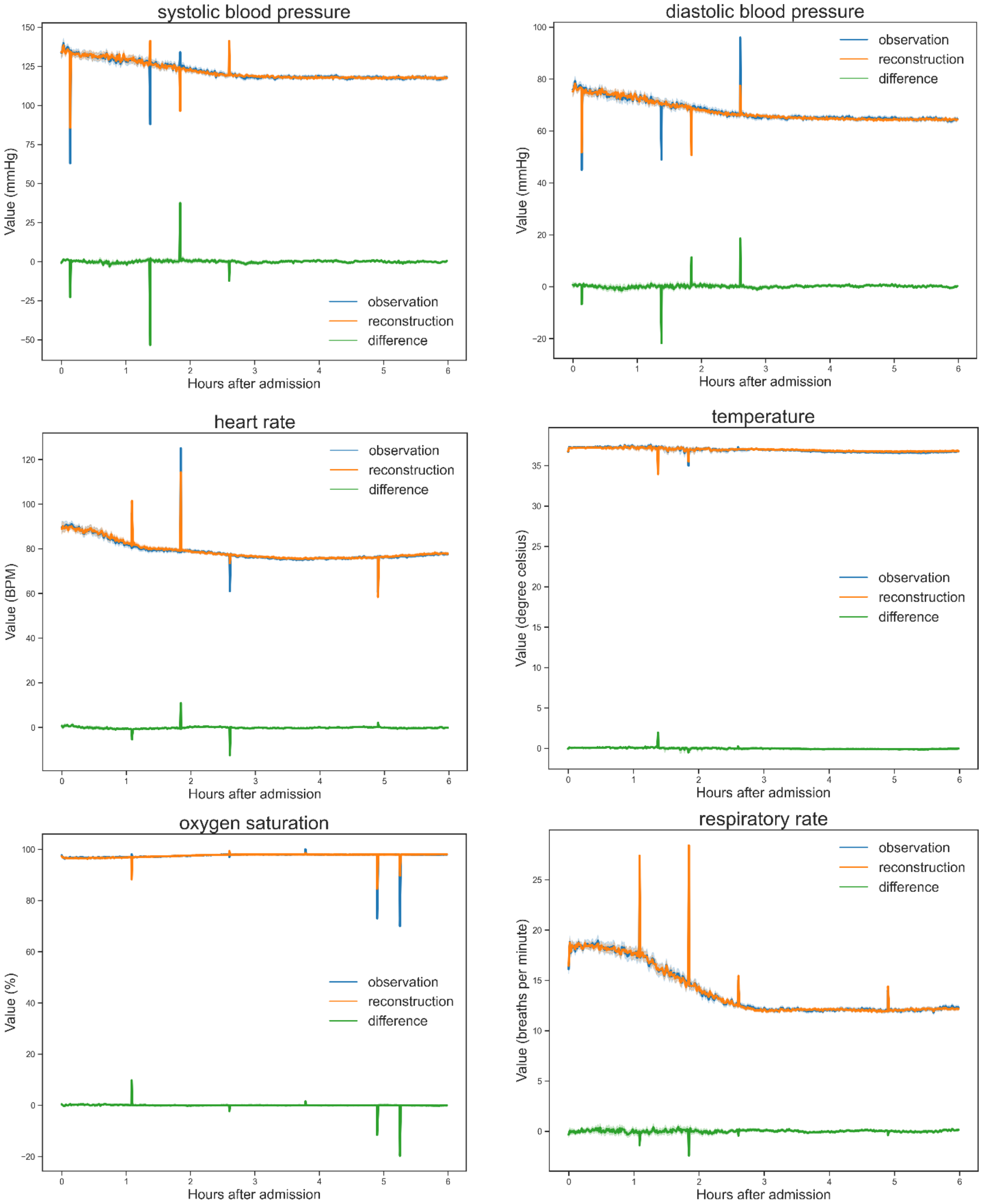

## eTable 1. Processing of vital sign time series

| Variables | Unit | Non-outlier range [min, max][a] | Frequency (per hour) | Missing any measurement (N = 75,762), n (%) | Values used to impute variable completely missing[b] | Normal distribution |
|---|---|---|---|---|---|---|
| **Systolic blood pressure** | mmHg | (20, 300) | 2 | 0 (0) | 119.1 | Yes |
| **Diastolic blood pressure** | mmHg | (5, 225) | 2 | 0 (0) | 66.1 | Yes |
| **Heart rate** | beats per minute | (0, 300] | 2 | 9 (0) | 76.8 | Yes |
| **Temperature** | degree Celsius | (24, 45) | 1 | 10,713 (14) | 36.7 | No |
| **Peripheral capillary oxygen saturation** | % | (1, 100] | 2 | 3,744 (5) | 97.8 | No |
| **Respiratory rate** | breaths per minute | (0, 60] | 2 | 470 (1) | 12.7 | Yes |

[a] Derived from expert-defined ranges. Open brackets ")" indicate value is not included and closed brackets "]" indicate value is included in the interval.

[b] For time series data missing entirely, including instances in which a variable was missing entirely from an admission, mean values of corresponding variables measured values in the training cohort were imputed.

## eTable 2. Used LOINCS, range of values, direction of abnormal values for lab variables

| Lab variables | LOINCS | LOINC Description | Plausible Range[a] | Direction of abnormal value | Missingness in all cohort (N = 75,762), n (%) | Missingness in training cohort (N = 41,502), n (%) | Missingness in validation cohort (N = 17,415), n (%) | Missingness in testing cohort (N = 16,845), n (%) | Normal distribution |
|---|---|---|---|---|---|---|---|---|---|
| **Basic metabolic Panel (BMP)** | **89044-2/24321-2** | Basic metabolic and albumin panel - Serum or Plasma/Basic metabolic 2000 panel - Serum or Plasma | | | | | | | |
| Glucose | 2339-0, 2340-8, 2345-7, 41651-1*, 41652-9*, 41653-7*, 74774-1*, | Glucose in serum or plasma/blood | 25 - 1400 | Maximum, Minimum | 9,090 (12) | 4,783 (12) | 2,129 (12) | 2,178 (13) | No |
| Creatinine | 2160-0, 38483-4 | Creatinine in blood | 0 - 30 | Maximum | 10,045 (13) | 5,276 (13) | 2,353 (14) | 2,416 (14) | No |
| Bilirubin | 1975-2 | Bilirubin total in serum or plasma | 0 - 50 | Maximum | 38,086 (50) | 20,319 (49) | 8,853 (51) | 8,914 (53) | No |
| Albumin | 1751-7, 2862-1, 61151-7 | Albumin in serum or plasma | 0.6 - 6.0 | Minimum | 37,735 (50) | 20,134 (49) | 8,779 (50) | 8,822 (52) | Yes |
| Anion Gap | 33037-3, 10366-1 | Anion gap in Serum or Plasma | 1 - 40 | Maximum | 14,876 (20) | 9,531 (23) | 2,645 (15) | 2,700 (16) | Yes |

| Lab variables | LOINCS | LOINC Description | Plausible Range[a] | Direction of abnormal value | Missingness in all cohort (N = 75,762), n (%) | Missingness in training cohort (N = 41,502), n (%) | Missingness in validation cohort (N = 17,415), n (%) | Missingness in testing cohort (N = 16,845), n (%) | Normal distribution |
|---|---|---|---|---|---|---|---|---|---|
| **CBC Panel** | 57021-8 | CBC W Auto Differential panel - Blood | | | | | | | |
| White Blood Cell Count | 26464-8, 6690-2 | Leukocytes [#/volume] in Blood | 0.1 - 240 | Maximum, Minimum | 7,118 (9) | 3,676 (9) | 1,665 (10) | 1,777 (11) | No |
| Hemoglobin | 718-7, 14775-1*, 30313-1*, 30352-9*, | Hemoglobin [Mass/volume] in Blood | 3 - 23 | Minimum | 6,108 (8) | 3,112 (7) | 1,443 (8) | 1,553 (9) | Yes |
| Platelets | 26515-7, 777-3, 49497-1* | Platelets [#/volume] in Blood | 2 - 1900 | Minimum | 7,143 (9) | 3,684 (9) | 1,671 (10) | 1,788 (11) | Yes |
| Bands % | 26508-2, ,35332-6, 764-1* | Band form neutrophils/100 leukocytes in blood | 0.9 - 90 | Maximum | 72,187 (95) | 39,455 (95) | 16,625 (95) | 16,107 (96) | Yes |
| Lymphocytes % | 736-9, 737-7 | Lymphocytes/100 leukocytes in blood | 0 - 100 | Maximum, Minimum | 22,653 (30) | 12,079 (29) | 5,275 (30) | 5,299 (31) | Yes |
| **Gas Panel** | | | | | | | | | |
| **Gas Panel-Arteri** | 24336-0 | | | | | | | | |

| Lab variables | LOINCS | LOINC Description | Plausible Range[a] | Direction of abnormal value | Missingness in all cohort (N = 75,762), n (%) | Missingness in training cohort (N = 41,502), n (%) | Missingness in validation cohort (N = 17,415), n (%) | Missingness in testing cohort (N = 16,845), n (%) | Normal distribution |
|---|---|---|---|---|---|---|---|---|---|
| **al blood** | | | | | | | | | |
| PH | 2744-1 | pH of arterial blood | 5 - 8 | Maximum, Minimum | 64,519 (85) | 35,387 (85) | 14,894 (86) | 14,238 (85) | No |
| PO2 | 2703-7 | Oxygen [Partial pressure] in arterial blood | 0 - 800 | Minimum | 64,519 (85) | 35,386 (85) | 14,894 (86) | 14,239 (85) | No |
| Base deficit | 1922-4 | Base deficit in Arterial blood | 0 - 30 | Maximum | 69,565 (92) | 38,135 (92) | 15,986 (92) | 15,444 (92) | No |
| RDW | 788-0, 21000-5 | Erythrocyte distribution width [Ratio] | 2 - 40 | Maximum, Minimum | 7,113 (9) | 3,672 (9) | 1,664 (10) | 1,777 (11) | Yes |
| **Others** | | | | | | | | | |
| C-Reactive Protein (all sensitivity levels) | 30522-7, 1988-5 | C reactive protein in serum or plasma | 0 - 280 | Maximum | 65,389 (86) | 35,640 (86) | 15,159 (87) | 14,590 (87) | Yes |
| Lactate | 2518-9, 2524-7, 32693-4, 14118-4, 30242-2 | Lactate in blood | 0.3 - 28 | Maximum | 47,936 (63) | 26,055 (63) | 11,178 (64) | 10,703 (64) | No |
| ESR | 4537-7, 30341-2, 18184-2, | Erythrocyte sedimentation rate | 1 - 140 | Maximum | 69,214 (91) | 37,599 (91) | 16,033 (92) | 15,582 (93) | Yes |

| Lab variables | LOINCS | LOINC Description | Plausible Range[a] | Direction of abnormal value | Missingness in all cohort (N = 75,762), n (%) | Missingness in training cohort (N = 41,502), n (%) | Missingness in validation cohort (N = 17,415), n (%) | Missingness in testing cohort (N = 16,845), n (%) | Normal distribution |
|---|---|---|---|---|---|---|---|---|---|
| | 43402-7, 4538-5, 4539-3, 82477-1 | | | | | | | | |
| INR | 34714-6, 6301-6 | International normalized ratio | 0.8 - 18 | Maximum | 40,755 (54) | 21,145 (51) | 9,835 (56) | 9,775 (58) | No |
| Troponin (TnT, TnI) | 6598-7, 48425-3, 6597-9, 67151-1,10839-9, 42757-5, 49563-0 | Troponin T.cardiac in blood and Troponin I.cardiac in blood | 0 - 49 | Maximum | 49,379 (65) | 26,886 (65) | 11,553 (66) | 10,940 (65) | No |

[a] Values out of the range values were removed.
Abbreviations: BUN: blood urea nitrogen; CO2: carbon dioxide; AST: aspartate transaminase; ALT: alanine transaminase; PO2: partial pressure of oxygen; RDW: red cell distribution width; ESR: erythrocyte sedimentation rate; INR: international normalized ratio.
*Included LOINCs that have same description from other panels.

## eTable 3. Clinic characteristics and biomarkers of the cohorts

| Variables | Overall cohort | Training cohort | Validation Cohort | Testing Cohort |
|---|---|---|---|---|
| Number of encounters (%) | 75,762 | 41,502 (55) | 17,415 (23) | 16,845 (22) |
| **Preadmission clinical characteristics** | | | | |
| Age, mean (SD) | 54 (19) | 54 (19) | 54 (19)[a] | 55 (19)[a,b] |
| Female sex, n (%) | 41,449 (55) | 22,745 (55) | 9,499 (55) | 9,205 (55) |
| Race, n (%) | | | | |
| White | 53,101 (70) | 29,076 (70) | 12,171 (70) | 11,854 (70) |
| African American | 17,432 (23) | 9,634 (23) | 3,953 (23) | 3,845 (23) |
| Primary insurance, n (%) | | | | |
| Private | 17,641 (23) | 9,591 (23) | 4,115 (24) | 3,935 (23) |
| Medicare | 33,969 (45) | 18,499 (45) | 7,625 (44) | 7,845 (47)[a,b] |
| Medicaid | 16,742 (22) | 9,231 (22) | 3,919 (23) | 3,592 (21)[a,b] |
| Uninsured | 7,410 (10) | 4,181 (10) | 1,756 (10) | 1,473 (9)[a,b] |
| Residency area characteristics | | | | |
| Total proportion of African American (%), mean (SD) | 18.8 (17.5) | 18.7 (17.5) | 18.9 (17.5) | 18.7 (17.4) |
| Proportion below poverty (%), mean (SD) | 22.6 (10.1) | 22.7 (10.1) | 22.7 (10.3) | 22.5 (10.2)[a] |
| Distance from hospital (mile), median (IQR) | 18 (3, 34) | 18 (3, 34) | 18 (3, 34) | 18 (3, 34) |
| **Comorbidities** | | | | |
| Hypertension, n (%) | 38,985 (51) | 21,639 (52) | 8,878 (51)[a] | 8,468 (50)[a] |
| Cardiovascular disease, n (%)[c] | 21,743 (29) | 12,058 (29) | 4,983 (29) | 4,702 (28)[a] |
| Diabetes mellitus, n (%) | 18,127 (24) | 10,111 (24) | 4,071 (23)[a] | 3,945 (23)[a] |
| Chronic kidney disease, n (%) | 12,357 (16) | 6,518 (16) | 2,947 (17)[a] | 2,892 (17)[a] |
| **Admission characteristics of patients** | | | | |
| Emergent admission, n (%) | 55,008 (73) | 30,177 (73) | 12,542 (72) | 12,289 (73) |
| Transfer from another hospital, n (%) | 13,569 (18) | 7,115 (17) | 3,595 (21)[a] | 2,859 (17)[b] |
| **Primary admission diagnostic groups** | | | | |
| Diseases of the circulatory system, n (%) | 13,670 (18) | 7,719 (19) | 2,968 (17)[a] | 2,983 (18)[a] |
| Respiratory and infectious diseases, n (%) | 6,016 (8) | 3,306 (8) | 1,185 (7)[a] | 1,525 (9)[a,b] |

| Variables | Overall cohort | Training cohort | Validation Cohort | Testing Cohort |
|---|---|---|---|---|
| Complications of pregnancy and childbirth, n (%) | 5,760 (8) | 3,148 (8) | 1,366 (8) | 1,246 (7) |
| Diseases of the digestive/genitourinary systems, n (%) | 9,532 (13) | 5,184 (12) | 2,201 (13) | 2,147 (13) |
| Diseases of the musculoskeletal/connective tissue and skin, n (%) | 6,591 (9) | 3,651 (9) | 1,522 (9) | 1,418 (8) |
| Neoplasms, n (%) | 4,953 (7) | 2,743 (7) | 1,136 (7) | 1,074 (6) |
| **Clinical biomarkers and interventions within 24 hours of admission** | | | | |
| Surgery on admission day, n (%) | 15,996 (21) | 8,644 (21) | 3,801 (22)[a] | 3,551 (21) |
| ICU/IMC admission within first 24 hours, n (%) | 17,163 (23) | 9,426 (23) | 3,899 (22) | 3,838 (23) |
| **Cardiovascular system** | | | | |
| Hypotension (MAP < 60 mmHg) at any time, n (%) | 26,400 (35) | 14,470 (35) | 6,014 (35) | 5,916 (35) |
| Duration, median (IQR), minutes | 57 (15, 165) | 57 (15, 168) | 53 (14, 157) | 60 (15, 167)[b] |
| Vasopressors used, n (%) | 13,991 (18) | 7,531 (18) | 3,294 (19) | 3,166 (19) |
| Out of operating room | 2,596 (3) | 1,403 (3) | 625 (4) | 568 (3) |
| Hypertension (SBP > 160 mmHg) at any time, n (%) | 2,7267 (36) | 14,838 (36) | 6,222 (36) | 6,207 (37)[a] |
| Duration, median (IQR), minutes | 120 (26, 358) | 120 (27, 356) | 120 (25, 373) | 120 (26, 352) |
| Troponin, tested, n (%) | 26,383 (35) | 14,616 (35) | 5,862 (34)[a] | 5,905 (35)[b] |
| Abnormal result among tested, n (%) | 5,842 (22) | 3,398 (23) | 1,239 (21)[a] | 1,205 (20)[a] |
| **Respiratory system** | | | | |
| Highest administered FiO2, median (IQR) | 0.21 (0.21, 0.40) | 0.21 (0.21, 0.40) | 0.21 (0.21, 0.40) | 0.21 (0.21, 0.40) |
| Room air only, n (%) | 43,887 (58) | 23,963 (58) | 10,242 (59)[a] | 9,682 (57)[b] |
| 0.22 - 0.40, n (%) | 26,997 (36) | 14,790 (36) | 6,125 (35) | 6,082 (36) |
| > 0.4, n (%) | 4,878 (6) | 2,749 (7) | 1,048 (6)[a] | 1,081 (6) |
| PaO2/FiO2, tested with arterial blood gas, n (%) | 11,235 (15) | 6,113 (15) | 2,519 (14) | 2,603 (15)[b] |
| <200 among tested, n (%) | 4,137 (37) | 2,265 (37) | 908 (36) | 964 (37) |
| Mechanical ventilation, n (%) | 3,924 (5) | 2,123 (5) | 883 (5) | 918 (5) |
| **Kidney and acid-base status** | | | | |
| Preadmission estimated glomerular filtration rate[d] (mL/min per 1.73 m2), median (IQR) | 95 (77, 111) | 95 (78, 111) | 95 (76, 111) | 94 (76, 110)[a] |
| Highest / reference creatinine[d], mean (SD) | 1.24 (0.75) | 1.24 (0.66) | 1.24 (0.74)[a] | 1.23 (0.95)[a] |

| Variables | Overall cohort | Training cohort | Validation Cohort | Testing Cohort |
|---|---|---|---|---|
| Renal replacement therapy, n (%) | 1,166 (1.5) | 641 (1.5) | 257 (1.5) | 268 (1.6) |
| Highest anion gap, median (IQR), mmol/L | 14 (12, 17) | 14 (12, 17) | 14 (13, 17)[a] | 14 (12, 17)[a,b] |
| Arterial blood gas tested, n (%) | 11,242 (15) | 6,115 (15) | 2,521 (14) | 2,606 (15)[b] |
| pH < 7.3 among tested, n (%) | 2,580 (23) | 1,437 (23) | 557 (22) | 586 (22) |
| Highest base deficit, mean (SD), mmol/L | 4.8 (4.7) | 4.8 (4.7) | 4.9 (4.6) | 4.6 (4.8)[a,b] |
| Lactate, tested, n (%) | 27,826 (37) | 15,447 (37) | 6,237 (36)[a] | 6,142 (36) |
| 2 - 4 mmol/L among tested, n (%) | 6,717 (24) | 3,739 (24) | 1,498 (24) | 1,480 (24) |
| > 4 mmol/L among tested, n (%) | 2,532 (9) | 1,374 (9) | 578 (9) | 580 (9) |
| **Inflammation** | | | | |
| Highest white blood cell count, median (IQR), x10^9/L | 9 (7, 13) | 9 (7, 13) | 9 (7, 13)[a] | 9 (7, 12)[a] |
| Highest premature neutrophils (bands)), median (IQR), % | 10 (4, 20) | 10 (4, 20) | 9 (3, 19) | 9 (4, 18) |
| Lowest lymphocytes, median (IQR), % | 16 (9, 24) | 16 (9, 24) | 16 (9, 24) | 16 (9, 24) |
| C-reactive protein, tested, n (%) | 10373 (14) | 5862 (14) | 2256 (13)[a] | 2255 (13) |
| Highest C-reactive protein, median (IQR), mg/L | 18 (5, 81) | 18 (5, 77) | 17 (4, 80) | 28 (5, 93)[a,b] |
| Erythrocyte sedimentation rate, tested, n (%) | 6548 (9) | 3903 (9) | 1382 (8)[a] | 1263 (7)[a] |
| Highest erythrocyte sedimentation rate, median (IQR), mm/h | 40 (19, 74) | 40 (19, 73) | 42 (19, 77) | 41 (20, 73) |
| Highest temperature, mean (SD), celsius | 37.7 (0.6) | 37.7 (0.6) | 37.7 (0.6) | 37.7 (0.6)[a,b] |
| 38 - 39, n (%) | 15,774 (21) | 8,633 (21) | 3,563 (20) | 3,578 (21) |
| > 39, n (%) | 2,779 (4) | 1,548 (4) | 604 (3) | 627 (4) |
| Lowest temperature, mean (SD), celsius | 36.7 (0.9) | 36.7 (1.0) | 36.7 (0.9)[a] | 36.8 (0.9)[a,b] |
| **Hematologic** | | | | |
| Lowest hemoglobin, mean (SD), g/dL | 11.4 (2.3) | 11.5 (2.3) | 11.4 (2.3)[a] | 11.2 (2.3)[a,b] |
| Highest RDW, mean (SD), % | 15.4 (2.1) | 15.5 (2.1) | 15.2 (2.1)[a] | 15.3 (2.1)[a,b] |
| Lowest platelets, median (IQR), x10^9/L | 208 (160, 266) | 210 (161, 269) | 204 (157, 260)[a] | 207 (160, 266)[a,b] |
| Platelets < 200, n (%) | 31,078 (41) | 16,707 (40) | 7,489 (43)[a] | 6,882 (41)[a] |
| < 100 | 4,779 (15) | 2,643 (16) | 1,128 (15) | 1,008 (15) |
| 100 - 200 | 26,299 (85) | 14,064 (84) | 6,361 (85) | 5,874 (85) |
| International normalized ratio, tested, n (%) | 35,007 (46) | 20,357 (49) | 7,580 (44)[a] | 7,070 (42)[a,b] |

| Variables | Overall cohort | Training cohort | Validation Cohort | Testing Cohort |
|---|---|---|---|---|
| >= 2 | 3,291 (9) | 1,836 (9) | 757 (10)[a] | 698 (10) |
| **Neurologic** | | | | |
| Glasgow Coma Scale score, n (%) | | | | |
| Moderate (9 - 12) | 3,125 (4) | 1,708 (4) | 687 (4) | 730 (4) |
| Severe (<= 8) | 2,662 (4) | 1,482 (4) | 587 (3) | 593 (4) |
| **Liver and metabolic** | | | | |
| Bilirubin, tested, n (%) | 37,676 (50) | 21,183 (51) | 8,562 (49)[a] | 7931 (47)[a,b] |
| >= 2 mg/dL, n (%) | 2,530 (7) | 1,427 (7) | 607 (7) | 496 (6) |
| Highest glucose, median (IQR), mg/dL | 127 (104, 170) | 126 (104, 170) | 126 (104, 169) | 128 (104, 172)[a,b] |
| Albumin, tested, n (%) | 38,027 (50) | 21,368 (51) | 8,636 (50)[a] | 8,023 (48)[a,b] |
| < 2.5 | 2,159 (6) | 1,243 (6) | 471 (5) | 445 (6) |
| 2.5 - 3.5 | 12,118 (32) | 6,904 (32) | 2,665 (31)[a] | 2,549 (32) |

Abbreviation: ICU: intensive care unit; IMC: intermediate care unit; MAP: mean arterial pressure; SBP: systolic blood pressure; RDW: red cell distribution width; SD: standard deviation; IQR: interquartile range.
All p values were adjusted for multiple comparisons using the Bonferroni method.
[a] $p < 0.05$ compared to training cohort.
[b] $p < 0.05$ compared to validation cohort.
[c] Cardiovascular disease was considered if there was a history of congestive heart failure, coronary artery disease, or peripheral vascular disease.
[d] Reference glomerular filtration rate and reference creatinine were derived without use of race correction (see eMethods for details).

## eTable 4. Illness severity, clinical outcomes, and resource use of the cohorts

| Variables | Overall cohort | Training cohort | Validation Cohort | Testing Cohort |
|---|---|---|---|---|
| Number of encounters (%) | 75,762 | 41,502 (55) | 17,415 (23) | 16,845 (22) |
| **Acuity scores within 24h of admission** | | | | |
| SOFA score > 6, n (%) | 6,463 (9) | 3,506 (8) | 1,503 (9) | 1,454 (9) |
| Patients in ICU/IMC, SOFA score <= 6, n (%) | 12,477 (16) | 6,882 (17) | 2,795 (16) | 2,800 (17) |
| Patients in ICU/IMC, SOFA score > 6, n (%) | 4,686 (6) | 2,544 (6) | 1,104 (6) | 1,038 (6) |
| Patients in ward, SOFA score <= 6, n (%) | 56,822 (75) | 31,114 (75) | 13,117 (75) | 12,591 (75) |
| Patients in ward, SOFA score > 6, n (%) | 1,777 (2) | 962 (2) | 399 (2) | 416 (2) |
| MEWS score > 4, n (%) | 5,033 (7) | 2,828 (7) | 1,115 (6) | 1,090 (6) |
| Patients in ICU/IMC, MEWS score <= 4, n (%) | 13,316 (18) | 7,235 (17) | 3,041 (17) | 3,040 (18) |
| Patients in ICU/IMC, MEWS score > 4, n (%) | 3,847 (5.1) | 2,191 (5.3) | 858 (4.9) | 798 (4.7)[a] |
| Patients in ward, MEWS score <= 4, n (%) | 57,413 (76) | 31,439 (76) | 13,259 (76) | 12,715 (75) |
| Patients in ward, MEWS score > 4, n (%) | 1,186 (2) | 637 (2) | 257 (1) | 292 (2) |
| **Resource use during hospitalization** | | | | |
| Hospital days, median (IQR) | 4 (2, 7) | 4 (2, 7) | 4 (2, 7) | 4 (2, 7) |
| Surgery at any time, n (%) | 21,436 (28) | 11,634 (28) | 5,084 (29)[a] | 4,718 (28)[b] |
| Admitted to ICU/IMC[c], n (%) | 20,380 (27) | 11,121 (27) | 4,643 (27) | 4,616 (27) |
| Days in ICU/IMC[d], median (IQR) | 4 (2, 7) | 4 (2, 7) | 4 (2, 7) | 4 (2, 7)[a,b] |
| Days in ICU/IMC greater than 48 hrs, n (%) | 15,201 (75) | 8,332 (75) | 3,468 (75) | 3,401 (74) |
| Mechanical ventilation, n (%) | 5,970 (8) | 3,218 (8) | 1,403 (8) | 1,349 (8) |
| Mechanical ventilation hours, median (IQR)[e] | 32 (12, 108) | 35 (14, 113) | 31 (11, 105)[a] | 28 (10, 92)[a] |
| Mechanical ventilation greater than 2 calendar days, n (%) | 2,993 (50) | 1,661 (52) | 699 (50) | 633 (47)[a] |
| Renal replacement therapy, n (%) | 2,316 (3) | 1,262 (3) | 524 (3) | 530 (3) |
| **Complications** | | | | |
| Acute kidney injury overall, n (%) | 12,547 (17) | 6,905 (17) | 2,901 (17) | 2,741 (16) |
| Community-acquired AKI, n (%) | 7,007 (56) | 3,839 (56) | 1,603 (55) | 1,565 (57) |
| Hospital-acquired AKI, n (%) | 5,540 (44) | 3,066 (44) | 1,298 (45) | 1,176 (43) |
| Worst AKI staging, n (%) | | | | |

| Variables | Overall cohort | Training cohort | Validation Cohort | Testing Cohort |
|---|---|---|---|---|
| Stage 1 | 8,036 (64) | 4,360 (63) | 1,878 (65) | 1,798 (66) |
| Stage 2 | 2,407 (19) | 1,362 (20) | 533 (18) | 512 (19) |
| Stage 3 | 1,496 (12) | 848 (12) | 348 (12) | 300 (11) |
| Stage 3 with RRT | 608 (5) | 335 (5) | 142 (5) | 131 (5) |
| Venous thromboembolism, n (%) | 2,902 (4) | 1,257 (3) | 708 (4)[a] | 937 (6)[a,b] |
| Sepsis, n (%) | 7,322 (10) | 3,750 (9) | 1,659 (10) | 1,913 (11)[a,b] |
| Hospital disposition, n (%) | | | | |
| Hospital mortality | 2,134 (2.8) | 1,141 (2.7) | 480 (2.8) | 513 (3.0) |
| Another hospital, LTAC, SNF, Hospice | 8,423 (11.1) | 4,475 (10.8) | 2,002 (11.5)[a] | 1,946 (11.6)[a] |
| Home or short-term rehabilitation | 65,205 (86.1) | 35,886 (86.5) | 14,933 (85.7) | 14,386 (85.4)[a] |
| 30-day mortality, n (%) | 2,984 (4) | 1,633 (4) | 646 (4) | 705 (4) |
| Three-year mortality, n (%) | 14,634 (19) | 8,013 (19) | 3,297 (19) | 3,324 (20) |

Abbreviation: SOFA: sequential organ failure assessment; MEWS: modified early warning score; ICU: intensive care unit; IMC: intermediate care unit; SD: standard deviation; IQR: interquartile range.

All p-values were adjusted for multiple comparisons using the Bonferroni method.

[a] $p < 0.05$ compared to training cohort.

[b] $p < 0.05$ compared to validation cohort.

[c] At any time during hospitalization.

[d] Values were calculated among patients admitted to ICU/IMC.

[e] Values were calculated among patients requiring MV.

## eTable 5. Statistic output from the deep interpolation network modeling in the training cohort.

| | Statics | | Class size (N = 41,502), n (%) | | | | | | | | | |
|---|---|---|---|---|---|---|---|---|---|---|---|---|
| Class number | Sihouette[a] | Davies-Bouldin Index[b] | 1 | 2 | 3 | 4 | 5 | 6 | 7 | 8 | 9 | 10 |
| 2 | 0.60 | 0.77 | 34,750 (84) | 6,752 (16) | . | . | . | . | . | . | . | . |
| 3 | 0.33 | 1.04 | 6,110 (15) | 17,450 (42) | 17,942 (43) | . | . | . | . | . | . | . |
| 4 | 0.30 | 1.27 | 14,356 (35) | 4,238 (10) | 5,078 (12) | 17,830 (43) | . | . | . | . | . | . |
| 5 | 0.30 | 1.24 | 5,016 (12) | 17,401 (42) | 4,844 (12) | 13,886 (33) | 355 (0.9) | . | . | . | . | . |
| 6 | 0.24 | 1.42 | 14,307 (34) | 3,985 (10) | 2,631 (6) | 8,261 (20) | 11,965 (29) | 353 (0.9) | . | . | . | . |
| 7 | 0.23 | 1.42 | 8,496 (20) | 3,030 (7) | 11,834 (29) | 1,956 (5) | 14,019 (34) | 354 (0.9) | 1,813 (4) | . | . | . |
| 8 | 0.24 | 1.453 | 11,824 (28) | 1,229 (3) | 13,990 (34) | 2,841 (7) | 8,477 (20) | 345 (0.8) | 1,068 (3) | 1,728 (4) | . | . |
| 9 | 0.19 | 1.60 | 10,106 (24) | 1,227 (3) | 10,218 (25) | 9,701 (23) | 344 (0.8) | 2,769 (7) | 1,038 (3) | 1,589 (4) | 4,510 (11) | . |
| 10 | 0.18 | 1.63 | 2,746 (7) | 6,096 (15) | 10,041 (24) | 1,582 (4) | 1,221 (3) | 5,690 (14) | 9,298 (22) | 344 (0.8) | 999 (2) | 3,485 (8) |

[a] Higher Sihouette score indicates better clustering.
[b] Lower Davies-Bouldin Index score indicates better clustering.

## eTable 6. Phenotype clinical characteristics and biomarkers in the training cohort

| Variables | Total | Acute Illness Phenotypes | | | |
|---|---|---|---|---|---|
| | | Phenotype A | Phenotype B | Phenotype C | Phenotype D |
| Number of encounters (%) | 41,502 | 7,647 (18) | 13,710 (33) | 12,901 (31) | 7,244 (17) |
| **Preadmission clinical characteristics** | | | | | |
| Age, mean (SD), years | 54 (19) | 57 (19)[a,c] | 53 (19)[a,b] | 51 (19) | 57 (17)[a] |
| Female sex, n (%) | 22,745 (55) | 3,963 (52)[a,c] | 7,595 (55)[a,b] | 7,391 (57) | 3,796 (52)[a] |
| Race, n (%) | | | | | |
| White | 29,076 (70) | 5,203 (68)[a,b] | 9,421 (69)[b] | 9,021 (70) | 5,431 (75)[a] |
| African American | 9,634 (23) | 2,036 (27)[a,b,c] | 3,411 (25)[a,b] | 2,930 (23) | 1,257 (17)[a] |
| Primary insurance, n (%) | | | | | |
| Private | 9,591 (23) | 1,323 (17)[a,b,c] | 2,917 (21)[a,b] | 3,314 (26) | 2,037 (28)[a] |
| Medicare | 18,499 (45) | 3,839 (50)[a,b,c] | 6,120 (45)[a,b] | 5,158 (40) | 3,382 (47)[a] |
| Medicaid | 9,231 (22) | 1,641 (21)[a,b,c] | 3,213 (23)[b] | 3,104 (24) | 1,273 (18)[a] |
| Uninsured | 4,181 (10) | 844 (11)[b] | 1,460 (11)[b] | 1,325 (10) | 552 (8)[a] |
| Residing neighborhood characteristics | | | | | |
| Proportion of African Americans (%), mean (SD) | 18.7 (17.5) | 19.6 (17.8)[a,b] | 19.3 (17.8)[a,b] | 18.6 (17.5) | 17.2 (16.1)[a] |
| Proportion below poverty (%), mean (SD) | 22.7 (10.1) | 23.8 (10.1)[a,b,c] | 23.1 (10.1)[a,b] | 22.5 (10.0) | 21.2 (9.8)[a] |
| Distance from hospital (mile), median (IQR) | 18 (3, 34) | 14 (3, 27)[a,b,c] | 14 (3, 32)[a,b] | 18 (3, 36) | 23 (9, 40)[a] |
| **Comorbidities** | | | | | |
| Hypertension, n (%) | 21,639 (52) | 4,129 (54)[a,b] | 7,205 (53)[b] | 6,704 (52) | 3,601 (50)[a] |
| Cardiovascular disease, n (%)[b] | 12,058 (29) | 2,413 (32)[a,b,c] | 3,991 (29)[b] | 3,682 (29) | 1,972 (27) |
| Diabetes mellitus, n (%) | 10,111 (24) | 1,972 (26)[a,b] | 3,370 (25) | 3,100 (24) | 1,669 (23) |
| Chronic kidney disease, n (%) | 6,518 (16) | 1,450 (19)[a,b] | 2,467 (18)[a,b] | 1,802 (14) | 799 (11)[a] |
| **Admission characteristics of patients** | | | | | |
| Emergent admission, n (%) | 30,177 (73) | 7,257 (95)[a,b,c] | 11,764 (86)[a,b] | 8,064 (63) | 3,092 (43)[a] |
| Transfer from another hospital, n (%) | 7,115 (17) | 1,986 (26)[a,b,c] | 3,014 (22)[a,b] | 1,087 (8) | 1,028 (14)[a] |
| **Primary admission diagnostic groups** | | | | | |
| Diseases of the circulatory system | 7,719 (19) | 1,934 (25)[a,b,c] | 2,425 (18)[a,b] | 1,834 (14) | 1,526 (21)[a] |

| Variables | Total | Acute Illness Phenotypes | | | |
|---|---|---|---|---|---|
| | | Phenotype A | Phenotype B | Phenotype C | Phenotype D |
| Respiratory and infectious diseases | 3,306 (8) | 961 (13)[a,b,c] | 1,161 (8)[a,b] | 705 (5) | 479 (7)[a] |
| Complications of pregnancy and childbirth | 3,148 (8) | 391 (5)[a,c] | 1,147 (8)[a,b] | 1,248 (10) | 362 (5)[a] |
| Diseases of the digestive/genitourinary systems | 5,184 (12) | 789 (10)[a,c] | 1686 (12)[a] | 1,876 (15) | 833 (11)[a] |
| Diseases of the musculoskeletal/connective tissue and skin | 3,651 (9) | 317 (4)[a,b,c] | 1042 (8)[a,b] | 1,222 (9) | 1,070 (15)[a] |
| Neoplasms | 2,743 (7) | 93 (1)[a,b,c] | 665 (5)[a,b] | 1,138 (9) | 847 (12)[a] |
| **Clinical biomarkers and interventions within 24 hours of admission** | | | | | |
| Surgical procedure on admission day, n (%) | 8,644 (21) | 272 (4)[a,b,c] | 813 (6)[a,b] | 3,466 (27) | 4,093 (57)[a] |
| ICU/IMC admission within first 24 hours, n (%) | 9,426 (23) | 2,372 (31)[a,b,c] | 1,979 (14)[b] | 1,921 (15) | 3,154 (44)[a] |
| **Cardiovascular system** | | | | | |
| Hypotension (MAP < 60 mmHg) at any time, n (%) | 14,470 (35) | 2,234 (29)[a,b,c] | 2,903 (21)[a,b] | 4,445 (34) | 4,888 (67)[a] |
| Duration, median (IQR), minutes | 57 (15, 168) | 86 (30, 224)[a,b] | 92 (30, 233)[a,b] | 33 (10, 120) | 37 (10, 129) |
| Vasopressors used, n (%) | 7,531 (18) | 421 (6)[a,b,c] | 641 (5)[a,b] | 2633 (20) | 3836 (53)[a] |
| Out of operating room | 1,403 (3) | 242 (3)[a,b,c] | 146 (1)[a,b] | 229 (2) | 786 (11)[a] |
| Hypertension (SBP > 160 mmHg) at any time, n (%) | 14,838 (36) | 2,923 (38)[a,b,c] | 3,684 (27)[a,b] | 4,272 (33) | 3,959 (55)[a] |
| Duration, median (IQR), minutes | 120 (27, 356) | 174 (52, 445)[a,b,c] | 214 (73, 477)[a,b] | 114 (19, 336) | 44 (9, 165)[a] |
| Troponin, tested, n (%) | 14,616 (35) | 4,502 (59)[a,b,c] | 4,862 (35)[a,b] | 3,055 (24) | 2,197 (30)[a] |
| Abnormal result among tested, n (%) | 3,398 (23) | 1,109 (25)[a,b,c] | 884 (18)[b] | 585 (19) | 820 (37)[a] |
| **Respiratory system** | | | | | |
| Highest administered FiO2, median (IQR), % | 0.21 (0.21, 0.40) | 0.21 (0.21, 0.29)[a,b,c] | 0.21 (0.21, 0.28)[a,b] | 0.21 (0.21, 0.40) | 0.40 (0.29, 0.40)[a] |
| Room air only, n (%) | 23,963 (58) | 4,615 (60)[b,c] | 10,130 (74)[a,b] | 7,874 (61) | 1,344 (19)[a] |
| 0.22 – 0.40, n (%) | 14,790 (36) | 2,496 (33)[a,b,c] | 3,252 (24)[a,b] | 4,484 (35) | 4,558 (63)[a] |
| > 0.40, n (%) | 2,749 (7) | 536 (7)[a,b,c] | 328 (2)[a,b] | 543 (4) | 1,342 (19)[a] |
| $P_aO2/FiO_2$, tested with arterial blood gas, n (%) | 6,113 (15) | 1,352 (18)[a,b,c] | 1,033 (8)[a,b] | 1,273 (10) | 2,455 (34)[a] |
| <200 among tested, n (%) | 2,265 (37) | 496 (37)[b,c] | 301 (29)[b] | 430 (34) | 1,038 (42)[a] |
| Mechanical ventilation, n (%) | 2,123 (5) | 434 (6)[a,b,c] | 191 (1)[a,b] | 314 (2) | 1,184 (16)[a] |
| **Kidney and acid-base status** | | | | | |

| Variables | Total | Acute Illness Phenotypes | | | |
|---|---|---|---|---|---|
| | | Phenotype A | Phenotype B | Phenotype C | Phenotype D |
| Preadmission estimated glomerular filtration rate[c] (mL/min per 1.73 $m^2$), median (IQR) | 95 (78, 111) | 93 (74, 109)[a,c] | 96 (77, 111)[a,b] | 97 (80, 113) | 93 (79, 106)[a] |
| Highest /reference creatinine[c] ratio, mean (SD) | 1.24 (0.66) | 1.30 (0.70)[a,b,c] | 1.22 (0.59)[a,b] | 1.20 (0.67) | 1.28 (0.74)[a] |
| Renal replacement therapy, n (%) | 641 (2) | 160 (2)[a,c] | 168 (1)[b] | 175 (1) | 138 (2)[a] |
| Highest anion gap, median (IQR), mmol/L | 14 (12, 17) | 15 (12, 18)[a,c] | 14 (12, 16)[b] | 14 (11, 16) | 15 (12, 18)[a] |
| Arterial blood gas tested, n (%) | 6,115 (15) | 1,353 (18)[a,b,c] | 1,033 (8)[a,b] | 1,274 (10) | 2,455 (34)[a] |
| pH < 7.3 among tested, n (%) | 1437 (23) | 298 (22)[b,c] | 160 (15)[b] | 242 (19) | 737 (30)[a] |
| Highest base deficit among tested, mean (SD), mmol/L | 4.8 (4.7) | 5.4 (5.0)[a,c] | 4.5 (4.6) | 4.4 (4.3) | 4.8 (4.6) |
| Lactate, tested, n (%) | 15,447 (37) | 3,935 (51)[a,b,c] | 4,308 (31)[a,b] | 3,706 (29) | 3,498 (48)[a] |
| 2 – 4 mmol/L among tested, n (%) | 3,739 (24) | 978 (25)[c] | 956 (22)[b] | 870 (23) | 935 (27)[a] |
| > 4 mmol/L among tested, n (%) | 1,374 (9) | 331 (8)[a,b,c] | 207 (5)[b] | 216 (6) | 620 (18)[a] |
| **Inflammation** | | | | | |
| Highest white blood cell count, median (IQR), x$10^9$/L | 9 (7, 13) | 9 (7, 13)[b,c] | 9 (6, 12)[a,b] | 9 (7, 12) | 11 (8, 15)[a] |
| Highest premature neutrophils (bands), median (IQR), % | 10 (4, 20) | 12 (4, 22)[a,b,c] | 7 (3, 15)[b] | 9 (3, 18) | 15 (7, 26)[a] |
| Lowest lymphocytes, median (IQR), % | 16 (9, 24) | 15 (8, 24)[a,b,c] | 16 (10, 25)[b] | 17 (10, 26) | 10 (6, 18)[a] |
| C-reactive protein, tested, n (%) | 5,862 (14) | 1,246 (16)[a,b] | 2,396 (17)[a,b] | 1,759 (14) | 461 (6)[a] |
| Highest C-reactive protein, median (IQR), mg/L | 18 (5, 77) | 20 (5, 89)[a,b] | 18 (5, 73)[a,b] | 15 (4, 70) | 39 (7, 112)[a] |
| Erythrocyte sedimentation rate, tested, n (%) | 3,903 (9) | 775 (10)[b,c] | 1,591 (12)[a,b] | 1,253 (10) | 284 (4)[a] |
| Highest erythrocyte sedimentation rate, median (IQR), mm/h | 40 (19, 73) | 42 (18, 72) | 41 (20, 75)[b] | 39 (19, 71) | 32 (15, 66) |
| Highest temperature, mean (SD), Celsius | 37.7 (0.6) | 37.7 (0.6)[b,c] | 37.6 (0.6)[a,b] | 37.7 (0.6) | 37.9 (0.6)[a] |
| 38 - 39, n (%) | 8,633 (21) | 1,519 (20)[b,c] | 2,238 (16)[a,b] | 2,486 (19) | 2,390 (33)[a] |
| > 39, n (%) | 1,548 (4) | 354 (5)[a,c] | 453 (3)[b] | 368 (3) | 373 (5)[a] |
| Lowest temperature, mean (SD), Celsius | 36.7 (1.0) | 36.7 (0.8)[b,c] | 36.8 (0.7)[a,b] | 36.7 (0.7) | 36.3 (1.7)[a] |
| **Hematologic** | | | | | |
| Lowest hemoglobin, mean (SD), g/dL | 11.5 (2.3) | 11.6 (2.4)[b] | 11.7 (2.3)[b] | 11.6 (2.2) | 10.9 (2.3)[a] |
| Highest RDW, mean (SD), % | 15.5 (2.1) | 15.6 (2.1)[a,b] | 15.6 (2.3)[a,b] | 15.4 (2.1) | 15.3 (1.9) |

| Variables | Total | Acute Illness Phenotypes | | | |
|---|---|---|---|---|---|
| | | Phenotype A | Phenotype B | Phenotype C | Phenotype D |
| Lowest platelets, median (IQR), $x10^9/L$ | 210 (161, 269) | 209 (160, 271)[a,b,c] | 216 (165, 277)[b] | 214 (165, 270) | 195 (150, 247)[a] |
| Platelets < 200, n (%), $x10^9/L$ | 16,707 (40) | 3,289 (43)[a,b,c] | 5,347 (39)[a,b] | 4,769 (37) | 3,302 (46)[a] |
| < 100 | 2,643 (16) | 526 (7) | 910 (7)[a] | 715 (6) | 492 (7) |
| 100 - 200 | 14,064 (84) | 2,763 (84) | 4,437 (83)[a] | 4,054 (85) | 2,810 (85) |
| International normalized ratio, tested, n (%) | 20,357 (49) | 4,830 (63)[a,b,c] | 6,942 (51)[a,b] | 5,201 (40) | 3,384 (47)[a] |
| >= 2 | 1,836 (9) | 432 (9) | 666 (10) | 433 (8) | 305 (9) |
| **Neurologic** | | | | | |
| Glasgow Coma Scale score, n (%) | | | | | |
| Moderate neurologic dysfunction (9 - 12) | 1,708 (4) | 385 (5)[a,b,c] | 340 (2)[b] | 284 (2) | 699 (10)[a] |
| Severe neurologic dysfunction (<= 8) | 1,482 (4) | 359 (5)[a,b,c] | 144 (1)[a,b] | 207 (2) | 772 (11)[a] |
| **Liver and metabolic** | | | | | |
| Bilirubin tested, n (%), mg/dL | 21,183 (51) | 4,759 (62)[a,b,c] | 8,018 (58)[a,b] | 5,894 (46) | 2,512 (35)[a] |
| ≥ 2 | 1,427 (7) | 271 (6)[b] | 542 (7)[b] | 395 (7) | 219 (9)[a] |
| Highest glucose, median (IQR), mg/dL | 126 (104, 170) | 127 (105, 175)[a,b,c] | 120 (101, 161)[b] | 123 (101, 164) | 144 (116, 187)[a] |
| Albumin, tested, n (%) | 21,368 (51) | 4,780 (63)[a,b,c] | 8,070 (59)[a,b] | 5,951 (46) | 2,567 (35)[a] |
| < 2.5 | 1,243 (6) | 304 (6)[a,b,c] | 419 (5)[b] | 260 (4) | 260 (10)[a] |
| 2.5 - 3.5 | 6,904 (32) | 1,678 (35)[a,b,c] | 2,515 (31)[a,b] | 1,719 (29) | 992 (39)[a] |

Abbreviation: ICU: intensive care unit; IMC: intermediate care unit; MAP: mean arterial pressure; SBP: systolic blood pressure; RDW: red cell distribution width; SD: standard deviation; IQR: interquartile range.
All p-values were adjusted for multiple comparisons using the Bonferroni method.
[a] p < 0.05 compared to Phenotype C .
[b] p < 0.05 compared to Phenotype D.
[c] p < 0.05 compared to Phenotype B.
[d] Cardiovascular disease was considered if there was a history of congestive heart failure, coronary artery disease, or peripheral vascular disease.
[e] Reference glomerular filtration rate and reference creatinine were derived without use of race correction (see eMethods for details).

## eTable 7. Phenotype illness severity, clinical outcomes, and resource use in the training cohort

| Variables | Total | Acute Illness Phenotypes | | | |
|---|---|---|---|---|---|
| | | Phenotype A | Phenotype B | Phenotype C | Phenotype D |
| Number of encounters (%) | 41,502 | 7,647 (18) | 13,710 (33) | 12,901 (31) | 7,244 (17) |
| **Acuity scores within 24h of admission** | | | | | |
| SOFA score > 6, n (%) | 3,506 (8) | 656 (9)[a,b,c] | 508 (4)[a,b] | 768 (6) | 1,574 (22)[a] |
| Patients in ICU/IMC, SOFA score ≤ 6, n (%) | 6,882 (17) | 1,822 (24)[a,c] | 1,690 (12)[b] | 1,514 (12) | 1,856 (26)[a] |
| Patients in ICU/IMC, SOFA score > 6, n (%) | 2,544 (6) | 550 (7)[a,b,c] | 289 (2)[a,b] | 407 (3) | 1,298 (18)[a] |
| Patients on ward, SOFA score ≤ 6, n (%) | 31,114 (75) | 5,169 (68)[a,b,c] | 11,512 (84)[a,b] | 10,619 (82) | 3,814 (53)[a] |
| Patients on ward, SOFA score > 6, n (%) | 962 (2) | 106 (1)[a,b] | 219 (2)[a,b] | 361 (3) | 276 (4)[a] |
| MEWS score ≥ 5, n (%) | 2,828 (7) | 873 (11)[a,b,c] | 575 (4)[a,b] | 387 (3) | 993 (14)[a] |
| Patients in ICU/IMC, MEWS score ≤ 4, n (%) | 7,235 (17) | 1,703 (22)[a,b,c] | 1,618 (12)[b] | 1,643 (13) | 2,271 (31)[a] |
| Patients in ICU/IMC, MEWS score > 4, n (%) | 2,191 (5) | 669 (9)[a,b,c] | 361 (3)[b] | 278 (2) | 883 (12)[a] |
| Patients on ward, MEWS score ≤ 4, n (%) | 31,439 (76) | 5,071 (66)[a,b,c] | 11,517 (84)[b] | 10,871 (84) | 3,980 (55)[a] |
| Patients on ward, MEWS score > 4, n (%) | 637 (2) | 204 (3)[a,b,c] | 214 (2)[a] | 109 (1) | 110 (2)[a] |
| **Resource use during hospitalization** | | | | | |
| Hospital days, median (IQR) | 4 (2, 7) | 4 (2, 7)[a,b,c] | 4 (2, 7)[a,b] | 3 (2, 6) | 4 (3, 7)[a] |
| Surgery at any time, n (%) | 11,634 (28) | 860 (11)[a,b,c] | 2,006 (15)[a,b] | 4,452 (35) | 4,316 (60)[a] |
| Admitted to ICU/IMC[b], n (%) | 11,121 (27) | 2,700 (35)[a,b,c] | 2,673 (19)[b] | 2,446 (19) | 3,302 (46)[a] |
| Days in ICU/IMC[c], median (IQR) | 4 (2, 7) | 4 (3, 7)[a] | 4 (3, 7)[a] | 4 (2, 6) | 4 (3, 8)[a] |
| ICU/IMC stay greater than 48 hrs, n (%) | 8,332 (75) | 2,068 (77)[a] | 2,008 (75)[a] | 1,751 (72) | 2,505 (76)[a] |
| Mechanical ventilation, n (%) | 3,218 (8) | 695 (9)[a,b,c] | 554 (4)[a,b] | 628 (5) | 1341 (19)[a] |
| Mechanical ventilation hours, median (IQR)[d] | 35 (14, 113) | 44 (17, 127)[a,b] | 35 (13, 116) | 25 (11, 101) | 33 (14, 113) |
| Mechanical ventilation greater than 2 calendar days, n (%) | 1,661 (52) | 403 (58)[a,b] | 284 (51) | 302 (48) | 672 (50) |
| Renal replacement therapy, n (%) | 1,262 (3) | 314 (4)[a,b,c] | 396 (3) | 322 (2) | 230 (3)[a] |
| **Complications** | | | | | |
| Acute kidney injury, n (%) | 6905 (17) | 1,598 (21)[a,b,c] | 2,279 (17)[a,b] | 1,680 (13) | 1,348 (19)[a] |
| Community-acquired AKI, n (%) | 3839 (56) | 924 (58)[c] | 1,200 (53)[b] | 897 (53) | 818 (61)[a] |

| Variables | Total | Acute Illness Phenotypes | | | |
|---|---|---|---|---|---|
| | | Phenotype A | Phenotype B | Phenotype C | Phenotype D |
| Hospital-acquired AKI, n (%) | 3066 (44) | 674 (42)[c] | 1,079 (47)[b] | 783 (47) | 530 (39)[a] |
| Worst AKI staging, n (%) | | | | | |
| Stage 1 | 4360 (63) | 961 (60)[a,c] | 1,479 (65)[b] | 1,112 (66) | 808 (60)[a] |
| Stage 2 | 1362 (20) | 346 (22)[a] | 425 (19) | 300 (18) | 291 (22) |
| Stage 3 | 848 (12) | 206 (13) | 270 (12) | 202 (12) | 170 (13) |
| Stage 3 with RRT | 335 (5) | 85 (5) | 105 (5) | 66 (4) | 79 (6) |
| Venous thromboembolism, n (%) | 1257 (3) | 261 (3)[a,b] | 481 (4)[a,b] | 334 (3) | 181 (2) |
| Sepsis, n (%) | 3750 (9) | 1,049 (14)[a,b,c] | 1,102 (8)[a,b] | 754 (6) | 845 (12)[a] |
| Hospital disposition, n (%) | | | | | |
| Hospital mortality | 1141 (3) | 294 (4)[a,b,c] | 278 (2)[a,b] | 184 (1) | 385 (5)[a] |
| Another hospital, LTAC, SNF, Hospice | 4475 (11) | 1,140 (15)[a,b,c] | 1,591 (12)[a] | 932 (7) | 812 (11)[a] |
| Home or short-term rehabilitation | 35886 (86) | 6,213 (81)[a,b,c] | 11,841 (86)[a,b] | 11,785 (91) | 6,047 (83)[a] |
| 30-day mortality, n (%) | 1633 (3.9) | 439 (6)[a,c] | 458 (3)[a,b] | 278 (2) | 458 (6)[a] |
| Three-year mortality, n (%) | 8013 (19) | 1,892 (25)[a,b,c] | 2,861 (21)[a,b] | 1,975 (15) | 1,285 (18)[a] |

Abbreviation: SOFA: sequential organ failure assessment; MEWS: modified early warning score; ICU: intensive care unit; IMC: intermediate care unit; IQR: interquartile range.

[a] The p-values represent difference < 0.05 compared to Phenotype C and were adjusted for multiple comparisons using the Bonferroni method. Supplemental Tables list p values for all within-group comparisons.

[b] At any time during hospitalization.

[c] Values were calculated among patients admitted to ICU/IMC.

[d] Values were calculated among patients requiring MV.

## eTable 8. Phenotype clinical characteristics and biomarkers in the testing cohort

| Variables | Total | Acute Illness Phenotypes | | | |
|---|---|---|---|---|---|
| | | Phenotype A | Phenotype B | Phenotype C | Phenotype D |
| Number of encounters (%) | 16,845 | 3,036 (18) | 5,880 (35) | 5,201 (31) | 2,728 (16) |
| **Preadmission clinical characteristics** | | | | | |
| Age, mean (SD), years | 55 (19) | 57 (19)[a,b,c] | 54 (19)[a,b] | 52 (19) | 58 (17)[a] |
| Female sex, n (%) | 9,205 (55) | 1,592 (52)[a,c] | 3,287 (56)[b] | 2,932 (56) | 1,394 (51)[a] |
| Race, n (%) | | | | | |
| White | 11,854 (70) | 2,096 (69)[b] | 4,063 (69)[b] | 3,646 (70) | 2,049 (75)[a] |
| African American | 3,845 (23) | 780 (26)[a,b] | 1,415 (24)[b] | 1,181 (23) | 469 (17)[a] |
| Primary insurance, n (%) | | | | | |
| Private | 3,935 (23) | 557 (18)[a,b,c] | 1281 (22)[a,b] | 1339 (26) | 758 (28) |
| Medicare | 7,845 (47) | 1,574 (52)[a,c] | 2,717 (46)[a,b] | 2,196 (42) | 1,358 (50)[a] |
| Medicaid | 3,592 (21) | 623 (21)[b] | 1,330 (23)[b] | 1,172 (23) | 467 (17)[a] |
| Uninsured | 1,473 (9) | 282 (9)[b] | 552 (9)[b] | 494 (9) | 145 (5)[a] |
| Residing neighborhood characteristics | | | | | |
| Proportion of African Americans (%), mean (SD) | 18.7 (17.4) | 19.7 (17.9)[a,b] | 19.3 (17.6)[a,b] | 18.8 (17.8) | 16.4 (15.7)[a] |
| Proportion below poverty (%), mean (SD) | 22.5 (10.2) | 23.5 (10.3)[a,b,c] | 22.7 (10.2)[b] | 22.5 (10.2) | 20.7 (9.9)[a] |
| Distance from hospital (mile), median (IQR) | 18 (3, 34) | 14 (3, 27)[a,b,c] | 14 (3, 32)[a,b] | 18 (3, 34) | 24 (9, 40)[a] |
| **Comorbidities** | | | | | |
| Hypertension, n (%) | 8,468 (50) | 1,510 (50) | 2,966 (50) | 2,625 (50) | 1,367 (50) |
| Cardiovascular disease, n (%)[b] | 4,702 (28) | 850 (28) | 1,671 (28) | 1,415 (27) | 766 (28) |
| Diabetes mellitus, n (%) | 3,945 (23) | 761 (25) | 1,334 (23) | 1,203 (23) | 647 (24) |
| Chronic kidney disease, n (%) | 2,892 (17) | 638 (21)[b] | 1,169 (20)[a,b] | 787 (15) | 298 (11)[a] |
| **Admission characteristics of patients** | | | | | |
| Emergent admission, n (%) | 12,289 (73) | 2,888 (95)[a,b,c] | 5,048 (86)[a,b] | 3,353 (64) | 1,000 (37)[a] |
| Transfer from another hospital, n (%) | 2,859 (17) | 743 (24)[a,b,c] | 1287 (22)[a,b] | 464 (9) | 365 (13)[a] |
| **Primary admission diagnostic groups** | | | | | |
| Diseases of the circulatory system | 2,983 (18) | 688 (23)[a,c] | 989 (17)[a,b] | 734 (14) | 572 (21)[a] |

| Variables | Total | Acute Illness Phenotypes | | | |
|---|---|---|---|---|---|
| | | Phenotype A | Phenotype B | Phenotype C | Phenotype D |
| Respiratory and infectious diseases | 1,525 (9) | 436 (14)[a,b,c] | 576 (10)[a,b] | 351 (7) | 162 (6) |
| Complications of pregnancy and childbirth | 1,246 (7) | 151 (5)[a,c] | 471 (8)[a,b] | 497 (10) | 127 (5)[a] |
| Diseases of the digestive/genitourinary systems | 2,147 (13) | 316 (10)[a,c] | 759 (13)[a] | 772 (15) | 300 (11)[a] |
| Diseases of the musculoskeletal/connective tissue and skin | 1,418 (8) | 91 (3)[a,b,c] | 393 (7)[a,b] | 480 (9) | 454 (17)[a] |
| Neoplasms | 1,074 (6) | 54 (2)[a,b,c] | 306 (5)[a,b] | 344 (7) | 370 (14)[a] |
| **Clinical biomarkers and interventions within 24 hours of admission** | | | | | |
| Surgical procedure on admission day, n (%) | 3,551 (21) | 109 (4)[a,b,c] | 329 (6)[a,b] | 1398 (27) | 1,715 (63)[a] |
| ICU/IMC admission within first 24 hours, n (%) | 3,838 (23) | 977 (32)[a,b,c] | 891 (15)[b] | 827 (16) | 1,143 (42)[a] |
| **Cardiovascular system** | | | | | |
| Hypotension (MAP < 60 mmHg) at any time, n (%) | 5,916 (35) | 940 (31)[a,b,c] | 1,307 (22)[a,b] | 1,768 (34) | 1,901 (70)[a] |
| Duration, median (IQR), minutes | 60 (15, 167) | 90 (35, 240)[a,b] | 83 (30, 222)[a,b] | 36 (11, 129) | 33 (10, 120) |
| Vasopressors used, n (%) | 3,166 (19) | 185 (6)[a,b] | 286 (5)[a,b] | 1,088 (21) | 1,607 (59)[a] |
| Out of operating room | 568 (3) | 106 (3)[a,b,c] | 61 (1)[a,b] | 107 (2) | 294 (11)[a] |
| Hypertension (SBP > 160 mmHg) at any time, n (%) | 6,207 (37) | 1,144 (38)[b,c] | 1,643 (28)[a,b] | 1,820 (35) | 1,600 (59)[a] |
| Duration, median (IQR), minutes | 120 (26, 352) | 161 (45, 440)[a,b,c] | 207 (65, 486)[a,b] | 111 (18, 345) | 45 (9, 164)[a] |
| Troponin, tested, n (%) | 5,905 (35) | 1,753 (58)[a,b,c] | 2,174 (37)[a,b] | 1,258 (24) | 720 (26) |
| Abnormal result among tested, n (%) | 1,205 (20) | 398 (23)[b,c] | 357 (16)[b] | 240 (19) | 210 (29)[a] |
| **Respiratory system** | | | | | |
| Highest administered FiO2, median (IQR), % | 0.21 (0.21, 0.40) | 0.21 (0.21, 0.33)[b,c] | 0.21 (0.21, 0.28)[a,b] | 0.21 (0.21, 0.40) | 0.40 (0.40, 0.40)[a] |
| Room air only, n (%) | 9,682 (57) | 1,772 (58)[a,b,c] | 4,269 (73)[a,b] | 3,205 (62) | 436 (16)[a] |
| 0.22 – 0.40, n (%) | 6,082 (36) | 1,029 (34)[b,c] | 1,446 (25)[a,b] | 1,763 (34) | 1,844 (68)[a] |
| > 0.40, n (%) | 1,081 (6) | 235 (8)[a,b,c] | 165 (3)[a,b] | 233 (4) | 448 (16)[a] |
| $P_aO2/FiO_2$, tested with arterial blood gas, n (%) | 2,603 (15) | 621 (20)[a,b,c] | 526 (9)[b] | 537 (10) | 919 (34)[a] |
| <200 among tested, n (%) | 964 (37) | 239 (38) | 172 (33)[b] | 181 (34) | 372 (40) |
| Mechanical ventilation, n (%) | 918 (5) | 207 (7)[a,b,c] | 101 (2)[a,b] | 154 (3) | 456 (17)[a] |
| **Kidney and acid-base status** | | | | | |

| Variables | Total | Acute Illness Phenotypes | | | |
|---|---|---|---|---|---|
| | | Phenotype A | Phenotype B | Phenotype C | Phenotype D |
| Preadmission estimated glomerular filtration rate[c] (mL/min per 1.73 m$^2$), median (IQR) | 94 (76, 110) | 92 (72, 107)[a,c] | 94 (76, 112)[b] | 96 (79, 112) | 92 (78, 106)[a] |
| Highest /reference creatinine[c] ratio, mean (SD) | 1.23 (0.95) | 1.27 (0.70)[a,b,c] | 1.22 (0.63)[a] | 1.22 (1.44) | 1.23 (0.63) |
| Renal replacement therapy, n (%) | 268 (2) | 64 (2)[c] | 75 (1) | 87 (2) | 42 (2) |
| Highest anion gap, median (IQR), mmol/L | 14 (12, 17) | 15 (13, 18)[a,b,c] | 14 (12, 16)[a,b] | 14 (12, 17) | 15 (12, 18)[a] |
| Arterial blood gas tested, n (%) | 2,606 (15) | 622 (20)[a,b,c] | 526 (9)[b] | 538 (10) | 920 (34)[a] |
| pH < 7.3 among tested, n (%) | 586 (22) | 98 (16)[a,b] | 86 (16)[a,b] | 125 (23) | 277 (30)[a] |
| Highest base deficit among tested, mean (SD), mmol/L | 4.6 (4.8) | 4.3 (4.5) | 4.8 (5.5) | 4.4 (4.3) | 4.9 (4.9) |
| Lactate, tested, n (%) | 6,142 (36) | 1,559 (51)[a,b,c] | 1,826 (31)[a,b] | 1,486 (29) | 1,271 (47)[a] |
| 2 – 4 mmol/L among tested, n (%) | 1,480 (24) | 425 (27)[a,c] | 415 (23) | 308 (21) | 332 (26)[a] |
| > 4 mmol/L among tested, n (%) | 580 (9) | 154 (10)[b,c] | 93 (5)[a,b] | 113 (8) | 220 (17)[a] |
| **Inflammation** | | | | | |
| Highest white blood cell count, median (IQR), x10$^9$/L | 9 (7, 12) | 9 (7, 13)[a,b,c] | 9 (6, 12)[a,b] | 9 (7, 12) | 11 (8, 14)[a] |
| Highest premature neutrophils (bands), median (IQR), % | 9 (4, 18) | 9 (5, 17)[b,c] | 7 (2, 15)[b] | 9 (4, 20) | 15 (7, 26)[a] |
| Lowest lymphocytes, median (IQR), % | 16 (9, 24) | 14 (8, 23)[a,b,c] | 16 (9, 25)[a,b] | 17 (10, 26) | 10 (6, 18)[a] |
| C-reactive protein, tested, n (%) | 2,255 (13) | 514 (17)[a,b] | 914 (16)[a,b] | 650 (12) | 177 (6)[a] |
| Highest C-reactive protein, median (IQR), mg/L | 28 (5, 93) | 34 (7, 99)[b] | 20 (5, 86)[b] | 17 (4, 77) | 66 (12, 160)[a] |
| Erythrocyte sedimentation rate, tested, n (%) | 1,263 (7) | 255 (8)[b,c] | 549 (9)[a,b] | 398 (8) | 61 (2)[a] |
| Highest erythrocyte sedimentation rate, median (IQR), mm/h | 41 (20, 73) | 39 (19, 67) | 42 (22, 74) | 39 (19, 75) | 36 (21, 57) |
| Highest temperature, mean (SD), Celsius | 37.7 (0.6) | 37.8 (0.7)[b,c] | 37.7 (0.6)[a,b] | 37.7 (0.5) | 37.9 (0.6)[a] |
| 38 - 39, n (%) | 3,578 (21) | 609 (20)[b,c] | 1,010 (17)[a,b] | 1,051 (20) | 908 (33)[a] |
| > 39, n (%) | 627 (4) | 164 (5)[a,c] | 207 (4)[a] | 127 (2) | 129 (5)[a] |
| Lowest temperature, mean (SD), Celsius | 36.8 (0.9) | 36.8 (0.9)[b] | 36.8 (0.6)[b] | 36.8 (0.7) | 36.4 (1.5)[a] |
| **Hematologic** | | | | | |
| Lowest hemoglobin, mean (SD), g/dL | 11.2 (2.3) | 11.4 (2.4)[b] | 11.3 (2.3)[b] | 11.4 (2.2) | 10.7 (2.2)[a] |
| Highest RDW, mean (SD), % | 15.3 (2.1) | 15.4 (2.1)[a,b] | 15.4 (2.2)[a,b] | 15.2 (1.9) | 15.1 (1.8) |

| Variables | Total | Acute Illness Phenotypes | | | |
|---|---|---|---|---|---|
| | | Phenotype A | Phenotype B | Phenotype C | Phenotype D |
| Lowest platelets, median (IQR), $x10^9/L$ | 207 (160, 266) | 208 (159, 265)[b,c] | 213 (164, 279)[a] | 208 (164, 266) | 192 (149, 242)[a] |
| Platelets < 200, n (%), $x10^9/L$ | 6,882 (41) | 1,324 (44)[a,b,c] | 2,329 (40)[a,b] | 1,992 (38) | 1,237 (45)[a] |
| < 100 | 1,008 (15) | 204 (15)[a] | 406 (17)[a,b] | 238 (12) | 160 (13) |
| 100 - 200 | 5,874 (85) | 1,120 (85)[a] | 1,923 (83)[a,b] | 1,754 (88) | 1,077 (87) |
| International normalized ratio, tested, n (%) | 7,070 (42) | 1,579 (52)[a,b,c] | 2,575 (44)[a] | 1,778 (34) | 1,138 (42)[a] |
| >= 2 | 698 (10) | 163 (10) | 276 (11) | 158 (9) | 101 (9) |
| **Neurologic** | | | | | |
| Glasgow Coma Scale score, n (%) | | | | | |
| Moderate neurologic dysfunction (9 - 12) | 730 (4) | 172 (6)[a,b,c] | 131 (2)[b] | 148 (3) | 279 (10)[a] |
| Severe neurologic dysfunction (<= 8) | 593 (4) | 161 (5)[a,b,c] | 69 (1)[b] | 77 (1) | 286 (10)[a] |
| **Liver and metabolic** | | | | | |
| Bilirubin tested, n (%), mg/dL | 7,931 (47) | 1,830 (60)[a,b,c] | 3,186 (54)[a,b] | 2,104 (40) | 811 (30)[a] |
| ≥ 2 | 496 (6) | 113 (6) | 196 (6) | 119 (6) | 68 (8) |
| Highest glucose, median (IQR), mg/dL | 128 (104, 172) | 131 (107, 181)[a,b,c] | 122 (102, 164)[b] | 123 (101, 165) | 143 (117, 185)[a] |
| Albumin, tested, n (%) | 8,023 (48) | 1,838 (61)[a,b,c] | 3,209 (55)[a,b] | 2,138 (41) | 838 (31)[a] |
| < 2.5 | 445 (6) | 117 (6)[a,b] | 160 (5)[b] | 85 (4) | 83 (10)[a] |
| 2.5 - 3.5 | 2,549 (32) | 639 (35)[a,c] | 979 (31)[b] | 599 (28) | 332 (40)[a] |

Abbreviation: ICU: intensive care unit; IMC: intermediate care unit; MAP: mean arterial pressure; SBP: systolic blood pressure; RDW: red cell distribution width; SD: standard deviation; IQR: interquartile range.
All p-values were adjusted for multiple comparisons using the Bonferroni method.
[a] $p < 0.05$ compared to Phenotype C .
[b] $p < 0.05$ compared to Phenotype D.
[c] $p < 0.05$ compared to Phenotype B.
[d] Cardiovascular disease was considered if there was a history of congestive heart failure, coronary artery disease, or peripheral vascular disease.
[e] Reference glomerular filtration rate and reference creatinine were derived without use of race correction (see eMethods for details).

## eTable 9. Phenotype illness severity, clinical outcomes, and resource use in the testing cohort

| Variables | Total | Acute Illness Phenotypes | | | |
|---|---|---|---|---|---|
| | | Phenotype A | Phenotype B | Phenotype C | Phenotype D |
| Number of encounters (%) | 16,845 | 3,036 (18) | 5,880 (35) | 5,201 (31) | 2,728 (16) |
| **Acuity scores within 24h of admission** | | | | | |
| SOFA score > 6, n (%) | 1,454 (9) | 286 (9)[a,b,c] | 244 (4)[a,b] | 336 (6) | 588 (22)[a] |
| Patients in ICU/IMC, SOFA score ≤ 6, n (%) | 2,800 (17) | 741 (24)[a,c] | 742 (13)[b] | 631 (12) | 686 (25)[a] |
| Patients in ICU/IMC, SOFA score > 6, n (%) | 1,038 (6) | 236 (8)[a,b,c] | 149 (3)[a,b] | 196 (4) | 457 (17)[a] |
| Patients on ward, SOFA score ≤ 6, n (%) | 12,591 (75) | 2,009 (66)[a,b,c] | 4,894 (83)[b] | 4,234 (81) | 1,454 (53)[a] |
| Patients on ward, SOFA score > 6, n (%) | 416 (2) | 50 (2)[a,b] | 95 (2)[a,b] | 140 (3) | 131 (5)[a] |
| MEWS score ≥ 5, n (%) | 1,090 (6) | 358 (12)[a,c] | 271 (5)[a,b] | 175 (3) | 286 (10)[a] |
| Patients in ICU/IMC, MEWS score ≤ 4, n (%) | 3,040 (18) | 722 (24)[a,b,c] | 728 (12)[b] | 702 (13) | 888 (33)[a] |
| Patients in ICU/IMC, MEWS score > 4, n (%) | 798 (5) | 255 (8)[a,c] | 163 (3)[b] | 125 (2) | 255 (9)[a] |
| Patients on ward, MEWS score ≤ 4, n (%) | 12,715 (75) | 1,956 (64)[a,b,c] | 4,881 (83)[b] | 4,324 (83) | 1,554 (57)[a] |
| Patients on ward, MEWS score > 4, n (%) | 292 (2) | 103 (3)[a,b,c] | 108 (2)[a] | 50 (1) | 31 (1) |
| **Resource use during hospitalization** | | | | | |
| Hospital days, median (IQR) | 4 (2, 7) | 4 (2, 7)[a,c] | 4 (2, 7)[a,b] | 3 (2, 6) | 4 (3, 7)[a] |
| Surgery at any time, n (%) | 4,718 (28) | 338 (11)[a,b,c] | 856 (15)[a,b] | 1,728 (33) | 1,796 (66)[a] |
| Admitted to ICU/IMC[b], n (%) | 4,616 (27) | 1,130 (37)[a,b,c] | 1,233 (21)[b] | 1,055 (20) | 1,198 (44)[a] |
| Days in ICU/IMC[c], median (IQR) | 4 (2, 7) | 4 (2, 7) | 4 (2, 7)[b] | 4 (2, 6) | 4 (3, 7)[a] |
| ICU/IMC stay greater than 48 hrs, n (%) | 3,401 (74) | 821 (73) | 906 (73) | 756 (72) | 918 (77)[a] |
| Mechanical ventilation, n (%) | 1,349 (8) | 307 (10)[a,b,c] | 258 (4)[b] | 268 (5) | 516 (19)[a] |
| Mechanical ventilation hours, median (IQR)[d] | 28 (10, 92) | 31 (14, 90) | 29 (9, 92) | 20 (7, 69) | 29 (12, 98) |
| Mechanical ventilation greater than 2 calendar days, n (%) | 633 (47) | 151 (49) | 128 (50) | 114 (43) | 240 (47) |
| Renal replacement therapy, n (%) | 530 (3) | 127 (4)[c] | 160 (3) | 161 (3) | 82 (3) |
| **Complications** | | | | | |
| Acute kidney injury, n (%) | 2,741 (16) | 636 (21)[a,b,c] | 970 (16)[a] | 670 (13) | 465 (17)[a] |
| Community-acquired AKI, n (%) | 1,565 (57) | 366 (58) | 521 (54)[b] | 395 (59) | 283 (61) |

| Variables | Total | Acute Illness Phenotypes | | | |
|---|---|---|---|---|---|
| | | Phenotype A | Phenotype B | Phenotype C | Phenotype D |
| Hospital-acquired AKI, n (%) | 1,176 (43) | 270 (42) | 449 (46)[b] | 275 (41) | 182 (39) |
| Worst AKI staging, n (%) | | | | | |
| Stage 1 | 1,798 (66) | 401 (63) | 654 (67) | 455 (68) | 288 (62) |
| Stage 2 | 512 (19) | 127 (20) | 169 (17) | 131 (20) | 85 (18) |
| Stage 3 | 300 (11) | 78 (12) | 109 (11) | 57 (9) | 56 (12) |
| Stage 3 with RRT | 131 (5) | 30 (5) | 38 (4)[b] | 27 (4) | 36 (8) |
| Venous thromboembolism, n (%) | 937 (6) | 203 (7)[a,b] | 364 (6)[a,b] | 256 (5) | 114 (4) |
| Sepsis, n (%) | 1,913 (11) | 550 (18)[a,b,c] | 661 (11)[a] | 400 (8) | 302 (11)[a] |
| Hospital disposition, n (%) | | | | | |
| Hospital mortality | 513 (3) | 130 (4)[a,c] | 149 (3)[a,b] | 75 (1) | 159 (6)[a] |
| Another hospital, LTAC, SNF, Hospice | 1,946 (12) | 464 (15)[a,b,c] | 730 (12)[a] | 435 (8) | 317 (12)[a] |
| Home or short-term rehabilitation | 14,386 (85) | 2,442 (80)[a,c] | 5,001 (85)[a,b] | 4,691 (90) | 2,252 (83)[a] |
| 30-day mortality, n (%) | 705 (4) | 181 (6)[a,c] | 221 (4)[a,b] | 123 (2) | 180 (7)[a] |
| Three-year mortality, n (%) | 3,324 (20) | 781 (26)[a,b,c] | 1,228 (21)[a,b] | 815 (16) | 500 (18)[a] |

Abbreviation: SOFA: sequential organ failure assessment; MEWS: modified early warning score; ICU: intensive care unit; IMC: intermediate care unit; IQR: interquartile range.

[a] The p-values represent difference < 0.05 compared to Phenotype C and were adjusted for multiple comparisons using Bonferroni method. Supplemental Tables list p values for all within-group comparisons.

[b] At any time during hospitalization.

[c] Values were calculated among patients admitted to ICU/IMC.

[d] Values were calculated among patients requiring MV.